\documentclass[11pt,onecolumn,letterpaper]{article}
\usepackage[margin=1in]{geometry}
\usepackage{times,xspace}
\usepackage[dvipsnames]{xcolor}
\usepackage{graphicx,amsmath,amssymb,booktabs,bm}
\usepackage[numbers,sort&compress]{natbib}
\usepackage{etoolbox,comment,placeins,float}
\usepackage[format=plain,labelsep=period,font=small]{caption}
\usepackage[font=small,skip=3pt]{subcaption}
\usepackage{titlesec}
\titlespacing*{\section}{0pt}{2.5ex plus 1ex minus .2ex}{1.5ex plus .2ex}
\titlespacing*{\subsection}{0pt}{2ex plus .8ex minus .2ex}{1ex plus .2ex}
\titlespacing*{\subsubsection}{0pt}{1.8ex plus .6ex minus .2ex}{.8ex plus .2ex}
\usepackage[normalem]{ulem}
\usepackage{tabularx,adjustbox}
\usepackage[shortlabels,inline]{enumitem}
\setlist{noitemsep,topsep=4pt,leftmargin=*}
\usepackage[hyphens]{url}
\usepackage[breaklinks,colorlinks]{hyperref}
\usepackage[capitalize]{cleveref}
\definecolor{preprintblue}{rgb}{0.21,0.49,0.74}
\hypersetup{allcolors=preprintblue,pdftitle={Interpreting Driving World Models with Spatial Block-Sparse Features}}
\def\1{\bm{1}}

\def\vmu{{\bm{\mu}}}

\def\va{{\bm{a}}}
\def\vb{{\bm{b}}}
\def\vc{{\bm{c}}}
\def\vd{{\bm{d}}}

\def\vh{{\bm{h}}}

\def\vs{{\bm{s}}}

\def\vu{{\bm{u}}}
\def\vv{{\bm{v}}}

\def\vx{{\bm{x}}}

\def\vz{{\bm{z}}}

\def\mA{{\mathbf{A}}}

\def\mC{{\mathbf{C}}}
\def\mD{{\mathbf{D}}}
\def\mE{{\mathbf{E}}}

\def\mG{{\mathbf{G}}}
\def\mH{{\mathbf{H}}}
\def\mI{{\mathbf{I}}}

\def\mL{{\mathbf{L}}}

\def\mO{{\mathbf{O}}}
\def\mP{{\mathbf{P}}}
\def\mQ{{\mathbf{Q}}}

\def\mS{{\mathbf{S}}}
\def\mT{{\mathbf{T}}}

\def\mV{{\mathbf{V}}}
\def\mW{{\mathbf{W}}}
\def\mX{{\mathbf{X}}}

\def\mZ{{\mathbf{Z}}}

\def\mOmega{{\bm{\Omega}}}

\DeclareMathAlphabet{\mathsfit}{\encodingdefault}{\sfdefault}{m}{sl}
\SetMathAlphabet{\mathsfit}{bold}{\encodingdefault}{\sfdefault}{bx}{n}

\newcommand{\vbeta}{\bm{\beta}}

\newcommand{\vdelta}{\bm{\delta}}

\newif\ifmarkrecentedits
\markrecenteditstrue
\newcommand{\recenteditcolor}{%
  \ifmarkrecentedits
    \color{red}%
    \hypersetup{allcolors=red}%
  \fi
}

\newenvironment{recentedits}{\begingroup\recenteditcolor}{\par\endgroup}
\newcommand{\recenteditcaption}{%
  \ifmarkrecentedits
    \captionsetup{font+={color=red},labelfont+={color=red}}%
  \fi
}

\newif\ifmarkterminologyedits
\markterminologyeditsfalse
\DeclareRobustCommand{\terminologyedit}[1]{{%
  \ifmarkterminologyedits
    \color{red}%
    \hypersetup{allcolors=red}%
  \fi
  #1%
}}
\newif\ifmarkresourceedits
\markresourceeditsfalse
\newcommand{\resourceeditcolor}{%
  \ifmarkresourceedits
    \color{red}%
    \hypersetup{allcolors=red}%
  \fi
}
\DeclareRobustCommand{\resourceedit}[1]{{\resourceeditcolor #1}}
\expandafter\def\csname recentreference@lecun2015deep\endcsname{}
\expandafter\def\csname recentreference@alain2016understanding\endcsname{}
\expandafter\def\csname recentreference@raghu2021vision\endcsname{}
\expandafter\def\csname recentreference@gagvani_mechanistic_2026\endcsname{}
\expandafter\def\csname recentreference@opendrivelab_openscene\endcsname{}
\expandafter\def\csname recentreference@karnchanachari2024nuplan\endcsname{}
\expandafter\def\csname recentreference@wang2023openlanev2\endcsname{}
\expandafter\def\csname recentreference@robinson2026rfdetr\endcsname{}
\expandafter\def\csname recentreference@wu2024topomlp\endcsname{}
\expandafter\def\csname recentreference@wang2021fcos3d\endcsname{}
\newcommand{\installrecentreferencecolors}{%
  \let\recentoriginalbibitem\bibitem
  \RenewDocumentCommand{\bibitem}{o m}{%
    \par
    \ifmarkrecentedits
      \ifcsname recentreference@##2\endcsname
        \recenteditcolor
      \else
        \color{black}\hypersetup{allcolors=preprintblue}%
      \fi
    \fi
    \IfNoValueTF{##1}{\recentoriginalbibitem{##2}}%
      {\recentoriginalbibitem[##1]{##2}}%
  }%
}
\AddToHook{env/thebibliography/begin}{\installrecentreferencecolors}

\newcommand{\thirdbest}[1]{\textit{#1}}
\newcolumntype{Y}{>{\raggedright\arraybackslash}X}

\ExplSyntaxOn
\NewDocumentCommand{\suppsectiondef}{mm}
  { \cs_gset:cpn { suppsection@#1 } {#2} }
\NewDocumentCommand{\supptabledef}{mm}
  { \cs_gset:cpn { supptable@#1 } {#2} }
\NewDocumentCommand{\supptableref}{m}
  {
    Supplementary~Tab.~
    \cs_if_exist:cTF { r@#1 }
      { \hyperref[#1]{\ref*{#1}} }
      {
        \cs_if_exist:cTF { supptable@#1 }
          { \use:c { supptable@#1 } }
          { \PackageError{supplement-references}{Unknown~supplement~table~#1}
              {Refresh~the~number~map~with~make~supplement-references.} }
      }
  }
\NewDocumentCommand{\suppsectionnumber}{m}
  {
    \cs_if_exist:cTF { suppsection@#1 }
      { \use:c { suppsection@#1 } }
      { \PackageError{supplement-references}{Unknown~supplement~section~#1}
          {Refresh~the~section~map~with~make~supplement-references.} }
  }
\cs_new_protected:Npn \supp_ref_number:n #1
  {
    \cs_if_exist:cTF { r@#1 }
      { \hyperref[#1]{\ref*{#1}} }
      { \suppsectionnumber{#1} }
  }
\clist_new:N \l_supp_ref_labels_clist
\int_new:N \l_supp_ref_index_int
\NewDocumentCommand{\suppref}{m}
  {
    \group_begin:
    \clist_set:Nn \l_supp_ref_labels_clist {#1}
    \int_zero:N \l_supp_ref_index_int
    Supplementary~Sec
    \int_compare:nNnT { \clist_count:N \l_supp_ref_labels_clist } > {1} {s}
    .~
    \clist_map_inline:Nn \l_supp_ref_labels_clist
      {
        \int_incr:N \l_supp_ref_index_int
        \int_compare:nNnT { \l_supp_ref_index_int } > {1}
          {
            \int_compare:nNnTF { \l_supp_ref_index_int }
              = { \clist_count:N \l_supp_ref_labels_clist }
              { ~and~ } { ,~ }
          }
        \supp_ref_number:n {##1}
      }
    \group_end:
  }
\ExplSyntaxOff
\suppsectiondef{sec:supp-implementation}{A}
\suppsectiondef{sec:supp-layer-selection}{A.1}
\suppsectiondef{sec:supp-captured-representations}{A.2}
\suppsectiondef{sec:supp-interface-implications}{A.3}
\suppsectiondef{sec:supp-reconstruction-evaluation}{B}
\suppsectiondef{sec:supp-r2-populations}{B.1}
\suppsectiondef{sec:supp-spatial-metric}{B.2}
\suppsectiondef{sec:supp-temporal-metric}{B.3}
\suppsectiondef{sec:supp-reconstruction-pooling}{B.4}
\suppsectiondef{sec:supp-ears-dataset}{C}
\suppsectiondef{sec:supp-crs-schema}{C.2}
\suppsectiondef{sec:supp-crs-attributes}{C.2}
\suppsectiondef{sec:supp-crs-generation}{C.3}
\suppsectiondef{sec:supp-crs-reference-scope}{C.3}
\suppsectiondef{sec:supp-ears-generation-pipeline}{C.3.1}
\suppsectiondef{sec:supp-crs-relations}{C.4}
\suppsectiondef{sec:supp-method-reference-details}{C.5}
\suppsectiondef{sec:supp-interpretability}{D}
\suppsectiondef{sec:supp-interpretability-population}{D.1}
\suppsectiondef{sec:supp-interpretability-localization}{D.2}
\suppsectiondef{sec:supp-interpretability-relations}{D.3}
\suppsectiondef{sec:supp-interpretability-states}{D.4}
\suppsectiondef{sec:supp-interpretability-aggregation}{D.5}
\suppsectiondef{sec:supp-method-details}{E}
\suppsectiondef{sec:supp-method-encoder}{E.1}
\suppsectiondef{sec:supp-method-allocation}{E.2}
\suppsectiondef{sec:supp-method-spatial}{E.3}
\suppsectiondef{sec:supp-method-contribution}{E.3}
\suppsectiondef{sec:supp-method-refinement}{E.4}
\suppsectiondef{sec:supp-method-training}{E.5}
\suppsectiondef{sec:supp-training-details}{F}
\suppsectiondef{sec:supp-training-protocol}{F.1}
\suppsectiondef{sec:supp-sae-baselines}{F.2}
\suppsectiondef{sec:supp-dictionary-size}{F.3}
\suppsectiondef{sec:supp-r2-results}{G}
\suppsectiondef{sec:supp-results}{G}
\suppsectiondef{sec:supp-r2-comparison}{G.1}
\suppsectiondef{sec:supp-sae-reconstruction-results}{G.3}
\suppsectiondef{sec:supp-maveric-reconstruction}{G.4}
\suppsectiondef{sec:supp-full-r2-results}{G.4}
\suppsectiondef{sec:supp-residual-correction-diagnostic}{G.5}
\suppsectiondef{sec:supp-residual-correction-control}{G.6}
\suppsectiondef{sec:supp-semantic-results}{H}
\suppsectiondef{sec:supp-interpretability-results}{H.1}
\suppsectiondef{sec:supp-full-interpretability}{H.2}
\suppsectiondef{sec:supp-full-common-units}{H.2.1}
\suppsectiondef{sec:supp-full-readouts}{H.2.2}
\suppsectiondef{sec:supp-full-current-index}{H.2.3}
\suppsectiondef{sec:supp-road-spatial-transport}{H.3}
\suppsectiondef{sec:supp-lasso-average-budget}{I}
\suppsectiondef{sec:supp-mean-activity-protocol}{I.1}
\suppsectiondef{sec:supp-mean-activity-results}{I.2}
\suppsectiondef{sec:supp-mean-activity-reconstruction}{I.2}
\suppsectiondef{sec:supp-mean-activity-semantics}{I.3}
\suppsectiondef{sec:supp-mean-activity-fixed-comparison}{I.4}
\suppsectiondef{sec:supp-bsf-ablation}{J}
\suppsectiondef{sec:supp-ablation-design}{J.1}
\suppsectiondef{sec:supp-ablation-evaluation}{J.2}
\suppsectiondef{sec:supp-ablation-absolute-gains}{J.2.1}
\suppsectiondef{sec:supp-ablation-interface-results}{J.3}
\suppsectiondef{sec:supp-ablation-drivewam-results}{J.4}
\suppsectiondef{sec:supp-ablation-simwam-results}{J.5}
\suppsectiondef{sec:supp-score-calibration}{J.6}
\suppsectiondef{sec:supp-fixed-mixed-rank}{J.7}
\suppsectiondef{sec:supp-coefficient-fragmentation}{K}
\suppsectiondef{sec:supp-historical-diagnostics}{K}
\suppsectiondef{sec:supp-coefficient-reconstruction}{K.1}
\suppsectiondef{sec:supp-block-fragmentation}{K.2}
\suppsectiondef{sec:supp-application-demo}{L}
\suppsectiondef{sec:supp-demo-selection}{L.1}
\suppsectiondef{sec:supp-demo-edit-scale}{L.2}
\suppsectiondef{sec:supp-demo-checks}{L.3}
\suppsectiondef{sec:supp-demo-lane-clearance}{L.4}
\suppsectiondef{sec:supp-demo-block-semantics}{L.5}
\suppsectiondef{sec:supp-demo-drivelaw}{L.6}
\suppsectiondef{sec:supp-demo-drivewam}{L.7}
\suppsectiondef{sec:supp-demo-activation-sensitivity}{L.8}
\supptabledef{tab:supp-r2-support}{6}
\supptabledef{tab:supp-r2-support-full}{6a}
\supptabledef{tab:supp-crs-nodes}{7}
\supptabledef{tab:supp-crs-relations}{8}
\supptabledef{tab:supp-semantic-task-cohorts}{8a}
\supptabledef{tab:supp-semantic-readout-fitting}{8b}
\supptabledef{tab:supp-semantic-entity-features}{8c}
\supptabledef{tab:supp-training-protocol}{9}
\supptabledef{tab:supp-training-hyperparameters}{9a}
\supptabledef{tab:supp-r2-drivelaw}{10}
\supptabledef{tab:supp-r2-drivewam}{11}
\supptabledef{tab:supp-r2-simwam}{12}
\supptabledef{tab:supp-sae-reconstruction}{13}
\supptabledef{tab:supp-full-r2-overall}{14}
\supptabledef{tab:supp-full-r2-drivelaw}{15}
\supptabledef{tab:supp-full-r2-drivewam}{16}
\supptabledef{tab:supp-full-r2-simwam}{17}
\supptabledef{tab:supp-refinement-summary}{18}
\supptabledef{tab:supp-refinement-drive-sensitivity}{19}
\supptabledef{tab:supp-refinement-error-geometry}{20}
\supptabledef{tab:supp-refinement-drivelaw}{21}
\supptabledef{tab:supp-refinement-drivewam}{22}
\supptabledef{tab:supp-refinement-simwam}{23}
\supptabledef{tab:supp-residual-control}{23a}
\supptabledef{tab:supp-residual-refinement}{23b}
\supptabledef{tab:supp-ears-ten-task-selected}{24}
\supptabledef{tab:supp-ears-eleven-task-selected-localization}{25}
\supptabledef{tab:supp-ears-ten-task-selected-models}{26}
\supptabledef{tab:supp-ears-ten-task-full}{27}
\supptabledef{tab:supp-ears-eleven-task-full-localization}{28}
\supptabledef{tab:supp-ears-ten-task-full-models}{29}
\supptabledef{tab:supp-ears-ten-task-full-sites}{30}
\supptabledef{tab:supp-full-layer-semantics}{31}
\supptabledef{tab:supp-family-layer-means}{32}
\supptabledef{tab:supp-road-placement-controls}{33}
\supptabledef{tab:supp-lasso-average-reconstruction}{34}
\supptabledef{tab:supp-ears-eleven-task-mean-activity}{35a}
\supptabledef{tab:supp-ablation-summary}{36}
\supptabledef{tab:supp-ablation-contrasts}{37}
\supptabledef{tab:supp-ablation-portfolio-tasks}{38}
\supptabledef{tab:supp-ablation-drivelaw}{39}
\supptabledef{tab:supp-ablation-components-drivelaw}{40}
\supptabledef{tab:supp-ablation-drivewam}{41}
\supptabledef{tab:supp-ablation-components-drivewam}{42}
\supptabledef{tab:supp-ablation-simwam}{43}
\supptabledef{tab:supp-ablation-components-simwam}{44}
\supptabledef{tab:supp-rank-compression-summary}{44c}
\supptabledef{tab:supp-coefficient-budget}{45a}
\supptabledef{tab:supp-demo-all-case-clearance}{46a}
\supptabledef{tab:supp-demo-lane-clearance}{46}
\supptabledef{tab:supp-demo-drivelaw-recipes}{47}

\markrecenteditsfalse
\markterminologyeditsfalse
\markresourceeditsfalse
\crefname{table}{Tab.}{Tabs.}
\Crefname{table}{Tab.}{Tabs.}
\crefname{figure}{Fig.}{Figs.}
\Crefname{figure}{Fig.}{Figs.}
\crefname{section}{Sec.}{Secs.}
\Crefname{section}{Section}{Sections}
\crefname{appendix}{Sec.}{Secs.}
\Crefname{appendix}{Section}{Sections}
\newcommand{\supplementaryresultbreak}{\par\medskip}
\newenvironment{supplementaryresults}{\FloatBarrier}{}

\makeatletter
\patchcmd{\@maketitle}{\vskip 2em}{\vskip 0em}{}{\errmessage{Title top-spacing patch failed}}
\patchcmd{\@maketitle}{\vskip 1.5em}{\vskip 1em}{}{\errmessage{Title author-spacing patch failed}}
\patchcmd{\@maketitle}{\vskip 1em{\large \@date}}{}{}{\errmessage{Title date-spacing patch failed}}
\patchcmd{\@maketitle}{\par\vskip 1.5em}{\par\vskip .5em}{}{\errmessage{Title bottom-spacing patch failed}}
\makeatother
\title{Interpreting Driving World Models with Spatial Block-Sparse Features}

\author{%
  Yide Tao\textsuperscript{1}\quad
  Yichen Liu\textsuperscript{2}\quad
  Weize Cai\textsuperscript{2}\quad
  Maple Shao\textsuperscript{2}\\
  Matthew Willaton\textsuperscript{1}\quad
  Mehrtash Harandi\textsuperscript{1}\quad
  Hai L. Vu\textsuperscript{1}\quad
  Yongqi Dong\textsuperscript{2}\\[0.25em]
  {\normalsize\textsuperscript{1}Monash University, Melbourne, Australia}\\
  {\normalsize\textsuperscript{2}RWTH Aachen University, Aachen, Germany}%
}
\hypersetup{pdfauthor={Yide Tao, Yichen Liu, Weize Cai, Maple Shao, Matthew Willaton, Mehrtash Harandi, Hai L. Vu, Yongqi Dong}}

\date{}
\begin{document}
\maketitle
% Source: main-paper/figures/driving-block-overview.tex
\noindent\begin{minipage}{\textwidth}
\centering
\includegraphics[width=0.935\textwidth]{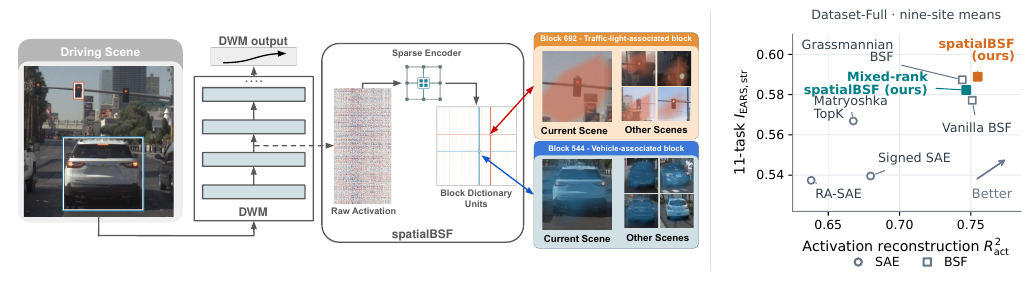}
\captionsetup{hypcap=false}
\captionof{figure}{\textbf{Interpreting driving world models with spatialBSF.}
Left: Interpretability in the ``wild''. Our analysis reveals features within driving world models (DriveLaW~\citep{xia_drivelaw_2026}) that are strongly associated with human-understandable concepts, such as traffic lights and cars.
Right: Activation reconstruction fidelity and the eleven-task EARS \terminologyedit{semantic agreement index (strength readout)}, averaged across nine interfaces for dictionaries trained on \textit{Dataset-Full}. Higher values are better on both axes.}
\label{fig:system-overview}
\end{minipage}
\par\vspace{6pt}
% Source: main-paper/sections/00-abstract.tex
\begin{abstract}
Driving world models (DWMs) unify scene prediction and action generation, but their internal representations remain opaque. Dictionary-based mechanistic interpretability (MI), including Sparse Autoencoders (SAEs), can disentangle language-model activations into interpretable concepts. However, visual driving scenes lack standardised object and relational descriptions that can be linked consistently to dictionary units. We introduce Ego-centric Attributed Road Scene Graph (\textbf{EARS-Graph}), a traceable scene graph representation that grounds dictionary units in spatial entities and quantifies relational fidelity. Extracting spatially coherent concepts remains challenging: Block-Sparse Featurisers (BSFs) group correlated directions into multi-dimensional blocks, but reconstruct tokens independently without coupling object representations across neighbouring tokens. We therefore propose \textbf{Spatial Block-Sparse Featurisers (spatialBSF)}, which pairs spatial decoding across neighbouring tokens with residual refinement, updating block selection and coefficients from reconstruction error. We evaluate spatialBSF on NAVSIM across three frozen DWMs: DriveWAM, SimWAM, and DriveLaW. We also contribute multi-layer activation datasets and trained SAE and BSF checkpoints, paired with EARS-Graph semantic references and evaluation code, to support future research on mechanistic interpretability of DWMs. Our eleven-task EARS benchmark evaluates dictionary-unit localisation and the accessibility of relations and object states. spatialBSF achieves the highest mean activation reconstruction fidelity, while EARS rankings vary by dataset and readout. Finally, interventions on spatialBSF blocks shortlisted using EARS-Graph prevent simulated at-fault collisions in two selected SimWAM fitting scenes.
\end{abstract}

% Source: main-paper/sections/01-introduction.tex
\FloatBarrier
\section{Introduction}
\label{sec:introduction}
We present a Mechanistic Interpretability (MI) framework for scene semantics in ``wild''~\citep{wang_interpretability_2022} driving world models (DWMs)~\citep{shi_drivewam_2026,zhao_simwam_2026,xia_drivelaw_2026}. Despite growing interest~\citep{guan_world_2024,feng_survey_2025}, DWM perception remains poorly understood, with few tailored dictionary tools and benchmarks linking units to standardised object and relational annotations.

Inspired partly by the Combined Road Substrate (CRS)~\citep{wild_bridging_2026}, our \textbf{Ego-centric Attributed Road Scene Graph (EARS-Graph)} maps camera images to fixed-vocabulary graphs of road surfaces, vehicles, and traffic elements. Across 300 annotated scenes, eleven tasks benchmark dictionary units across MI tools, DWMs, and layers: cars, large vehicles (truck$\cup$bus), roads, traffic lights, and selected relations and states.

Sparse Autoencoders (SAEs)~\citep{cunningham_sparse_2023,stevens_sparse_2025} and Block-Sparse Featurisers (BSFs)~\citep{felStructuringSparsityBlockSparse2026} decompose activations into sparse \emph{dictionary units}: scalar features or blocks. SAEs isolate directions but can fragment multidimensional visual concepts; BSFs group directions into subspaces preserving local geometry~\citep{felStructuringSparsityBlockSparse2026}. Prior work applies sparse dictionary methods to general vision backbones~\cite{felStructuringSparsityBlockSparse2026} and driving representations linked to trajectory prediction~\citep{motzkus_driving_2026,gagvani_mechanistic_2026}. Here, we compare scalar and block dictionaries using a shared
scene-semantic benchmark at three sampled layers in each of three DWMs~\citep{shi_drivewam_2026,zhao_simwam_2026,xia_drivelaw_2026}.
Subsequent studies can reuse our collected activations, trained dictionaries, and semantic references.

Token-independent BSFs lack explicit spatial relationships across structures such as roads~\citep{felStructuringSparsityBlockSparse2026}. Single-pass SAE encoders can leave an amortisation gap relative to input-optimised codes~\citep{oneill_amortisation_2025}. We introduce \textbf{Spatial Block-Sparse Featurisers (spatialBSF)}, combining decoding across neighbouring tokens with one residual-driven refinement step and block reselection, following iterative sparse inference~\citep{gregor_learning_2010}. Motivated by varying block dimensions~\citep{jerpelea_deeper_2026}, we compare fixed-rank and mixed-rank blocks.

Our contributions are:
\begin{enumerate}
    \item \textbf{Reusable DWM activation datasets and MI checkpoints.} Collected activations and trained SAE and BSF checkpoints cover nine interfaces across three frozen DWMs and two training-data regimes.
    \item \textbf{Scene-graph formalism and EARS-Graph.} We define a graph schema for driving scenes and build a scalable pre-labelling pipeline to release EARS-Graph, standardising semantic evaluation.
    \item \textbf{Spatial Block-Sparse Featurisers (spatialBSF).} We propose a spatially aware dictionary learning method that decomposes multi-token visual features via cross-token decoding and residual refinement.
\end{enumerate}
% Source: main-paper/sections/02-related-work.tex
\FloatBarrier
\section{Related Works}
\label{sec:related-work}

\textbf{Interpretability in Autonomous Driving.}
Earlier work achieved interpretability by modifying model architectures. Structured planners expose goals and manoeuvres~\citep{albrecht_interpretable_2021,guan_integrated_2023}, while neural models add visual attention~\citep{kim_interpretable_2017}, concept bottlenecks~\citep{echterhoff_driving_2023}, or localised feature maps~\citep{mirzaie_interpretable_2025}. Architectural modifications require retraining and cannot probe arbitrary pretrained networks. Post-hoc methods bypass this constraint by analysing trained models directly, such as using SAEs to link representations to trajectory predictions in decoupled vision-encoder and motion-decoder architectures~\citep{gagvani_mechanistic_2026,motzkus_driving_2026}. We extend this post-hoc approach to large DWMs beyond trajectory prediction, examining how spatial features encode scene elements and relations in the ``wild'' across layers.

\Cref{tab:resource-release-comparison} compares publicly verified driving MI resources with our planned release.
% Source: main-paper/tables/resource-release-comparison.tex
\begin{table}[!htbp]
\centering
\begin{minipage}{0.94\textwidth}

\centering\footnotesize
\setlength{\tabcolsep}{2pt}
\captionsetup{skip=4pt}
\caption{\textbf{Research-resource availability in driving MI.}
Model checkpoints and data will be released subject to applicable upstream licences (including Apache 2.0 and MIT) and non-commercial research-use restrictions.}
\label{tab:resource-release-comparison}
\begin{tabular*}{\columnwidth}{@{\extracolsep{\fill}}lcccc@{}}
\toprule
Study & Code & \shortstack{MI\\ckpts.} & \shortstack{Cached\\activ.} & \shortstack{Other\\data}\\
\midrule
Motzkus \& Bernhard~\citep{motzkus_driving_2026} & $\times$ & $\times$ & $\times$ & $\times$\\
Gagvani et al.~\citep{gagvani_mechanistic_2026} & $\checkmark$ & $\checkmark$ (4) & $\times$ & $\times$\\
Panda et al.~\citep{panda_what_2026} & $\times$ & $\times$ & $\times$ & $\times$\\
\midrule
\textbf{Ours (available soon)} & $\checkmark^{\dagger}$ & $\checkmark^{\dagger}$ (126) & $\checkmark^{\dagger}$ & $\checkmark^{\dagger}$\\
\bottomrule
\end{tabular*}
\par\smallskip
\begin{minipage}{\columnwidth}
\scriptsize
$\checkmark$: public; $\times$: public access unverified (17 Nov. 2026);
$\dagger$: planned, subject to upstream terms.
Data excludes reused corpora.
\citet{gagvani_mechanistic_2026}: four public SAEs; linked data inaccessible.
\end{minipage}

\end{minipage}
\end{table}

\textbf{SAEs and BSFs for Vision.}
Existing SAE variants adapted from language MI tasks to vision MI tasks include TopK~\citep{gao_scaling_nodate}, BatchTopK~\citep{pach_monosemantic_2025}, and Matryoshka~\citep{bussmann2025matryoshka,pach_monosemantic_2025}. Specialised vision variants include Archetypal (A-SAE) and Relaxed Archetypal SAEs (RA-SAE)~\citep{fel2025archetypal}. Vision-oriented BSF methods comprise vanilla, Grassmannian, and group-lasso variants~\citep{felStructuringSparsityBlockSparse2026}. These MI tools have also been applied to interpret vision-based robot foundation models~\citep{swann_multi-dimensional_2026}. We evaluate signed TopK SAE~\citep{gao_scaling_nodate}, Matryoshka TopK~\citep{bussmann2025matryoshka}, RA-SAE~\citep{fel2025archetypal}, and vanilla and Grassmannian BSFs~\citep{felStructuringSparsityBlockSparse2026} alongside spatialBSF and its mixed-rank variant (\cref{sec:spatialbsf}) on DWMs.

\textbf{Structured Scene Representations.}
OpenLane-V2~\citep{wang2023openlanev2} benchmarks lane geometry, lane connectivity, and lane--traffic-element relations. TopoNet~\citep{li2023toponet} and TopoMLP~\citep{wu2024topomlp} predict these scene structures and relationships from images. CRS~\citep{wild_bridging_2026} incorporates dynamic objects into road scene graphs for vision-language reasoning. We adapt and extend the CRS formulation to construct EARS-Graph for evaluating internal DWM activations. To our knowledge, the code and annotations for CRS have not been released, so a direct comparison of CRS with EARS-Graph is not available.
% Source: main-paper/sections/03-method.tex

\FloatBarrier
\section{Mechanistic Interpretability of Driving World Models}
\label{sec:method}
\noindent\textbf{Visual activations.}
We train dictionary-based MI tools on visual activations extracted during action prediction from three frozen DWMs: \textit{DriveLaW}~\citep{xia_drivelaw_2026}, \textit{DriveWAM}~\citep{shi_drivewam_2026}, and \textit{SimWAM}~\citep{zhao_simwam_2026}.
For each model, we collect activations from three layers on NAVSIM OpenScene~\citep{opendrivelab_openscene}, using the \textit{Dataset-Selected} and \textit{Dataset-Full} splits described in \cref{sec:evaluation-setup}.
We follow the BSF extraction procedure~\citep{felStructuringSparsityBlockSparse2026}, retaining each token's scene context and native spatial position.

Following the centring and global scaling used in BSF~\citep{felStructuringSparsityBlockSparse2026}, we normalise activations separately for each DWM and layer. We represent each normalised spatial field as a matrix $\mX\in\mathbb R^{N\times d}$, whose row $\vx_p=(\mX)_{p,:}\in\mathbb R^{1\times d}$ contains the $d$ activation channels at position $p$. Here $N$ is the number of retained spatial tokens; retained subsets preserve their native spatial coordinates. Details of layer selection, model-specific interfaces, extraction, and normalisation are provided in \suppref{sec:supp-layer-selection,sec:supp-captured-representations,sec:supp-interface-implications,sec:supp-r2-populations}.

\noindent\textbf{Dictionary units and block representation.}
A \emph{dictionary unit} is one SAE feature, associated with a single decoder atom, or one whole BSF block (containing one or more decoder atoms). At position $p$, SAE feature $j$ contributes $z_{pj}\vd_j^{\mathsf T}$ (scalar coefficient $z_{pj}$, decoder atom $\vd_j$). BSF block $b$ contributes $\vz_{pb}\mD_b^{\mathsf T}$ (coefficient row $\vz_{pb}$, block dictionary $\mD_b$). An SAE dictionary unit varies only in magnitude along a fixed axis. A block dictionary unit varies in both magnitude and orientation within its learned subspace, yet still counts as a single unit when active.

\noindent\textbf{Strength and direction readouts.}
Following BSF~\citep{felStructuringSparsityBlockSparse2026}, we evaluate strength and within-block direction separately.
A \emph{strength readout} is the norm of a dictionary unit's decoded contribution, testing whether the target concept is associated with a large response regardless of sign or orientation. The \emph{direction readout} is that contribution's signed projection onto a direction fitted to the target concept, divided by a fitting-set RMS. Unlike cosine similarity, the direction readout preserves the contribution's magnitude along the fitted direction. Equally strong block contributions can have different orientations within the learned subspace and may therefore encode different concept attributes. Comparing the two readouts tests whether a target concept is better identified by response magnitude or by a signed response along a fitted direction. For our spatialBSF, both readouts use its decoder's block contributions (\cref{sec:spatialbsf}).

\begin{figure}[!htbp]
\centering
\begin{minipage}{0.90\textwidth}

 \recenteditcolor\recenteditcaption
 \centering
 \includegraphics[width=0.87\columnwidth,height=0.50\textheight,keepaspectratio]{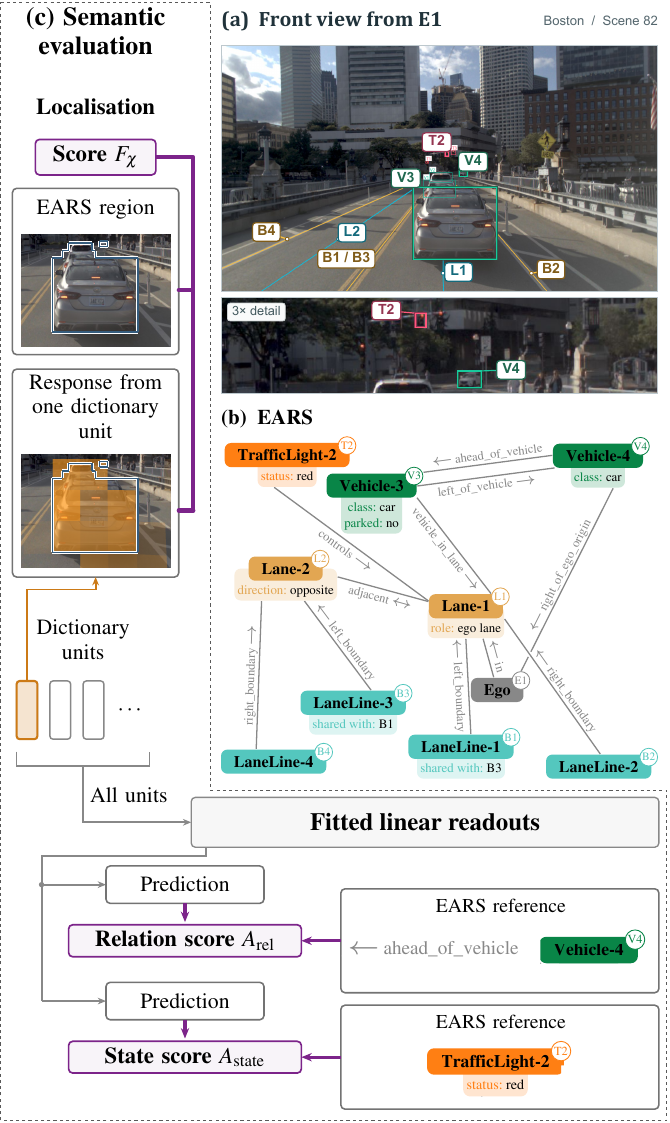}
 \caption{\textbf{EARS-Graph and eleven-task semantic evaluation.}
 (a) Scene; (b) Schematic EARS-Graph example: ahead, left, vehicle--lane and light--lane edges are illustrative, not this frame's evaluation labels; (c) Evaluation pathways: measured localisation response, illustrative relation example.}
 \label{fig:ears-reference}

\end{minipage}
\end{figure}

\subsection{Evaluation Criteria}
\label{sec:interpretability-problem}
We evaluate MI tools along two complementary dimensions:
\emph{activation reconstruction} and \emph{\terminologyedit{semantic agreement}}. High reconstruction fidelity does not guarantee reliable semantic features~\citep{gao_scaling_nodate}, while visually plausible features lacking strong activation alignment may reflect input correlations rather than the DWM's internal representations~\citep{adebayo_sanity_2018,heap_automated_2026}. Neither criterion alone suffices for a faithful account of model computation.

\noindent\textbf{Activation reconstruction.}
We use the standard centred $R^2_{\mathrm{act}}$ metric~\citep{felStructuringSparsityBlockSparse2026} to measure the variance in $\mX$ retained by each reconstruction $\widehat{\mX}$. Inspired by spatial gradient reconstruction~\citep{perez_poisson_2003} and temporal-difference modelling~\citep{isobe_look_2022}, we also report $R^2_{\mathrm{sp}}$ and $R^2_{\mathrm{temp}}$ for differences between neighbouring spatial positions and consecutive observations, respectively. These metrics provide a partial assessment of how well each MI tool preserves spatial and temporal variation that DWMs may rely on for their output. Detailed formulations appear in
\suppref{sec:supp-reconstruction-evaluation}.

\noindent\textbf{\terminologyedit{Semantic agreement.}}
Since learned dictionary units lack ground-truth targets, we evaluate alignment with selected EARS-Graph regions, attributes, and relations (\cref{sec:ears-reference-metrics}). For localisation ($F_\chi$, with $\chi\in\{\mathrm{str},\mathrm{dir}\}$ denoting strength or direction), we adapt BSF's single-concept F1~\citep{felStructuringSparsityBlockSparse2026} from image-level detection to spatial localisation of four selected targets. Following CRS~\citep{wild_bridging_2026}, we test whether unit combinations support relation and conjunction readouts ($A_{\mathrm{rel}}$, via balanced accuracy and F1) alongside states ($A_{\mathrm{state}}$, via balanced accuracy). Accounting for layer-wise variation~\citep{bau_network_dissection_2017}, we select the best-matching localisation unit per concept and readout per layer via fitting-drive cross-validation, then average validation scores uniformly across shared fixed layers (\suppref{sec:supp-interpretability}).

\begin{figure}[!htbp]
\centering
\begin{minipage}{0.90\textwidth}

 \recenteditcolor
 \recenteditcaption
 \centering
 \includegraphics[width=0.6435\textwidth,height=0.50\textheight,keepaspectratio]{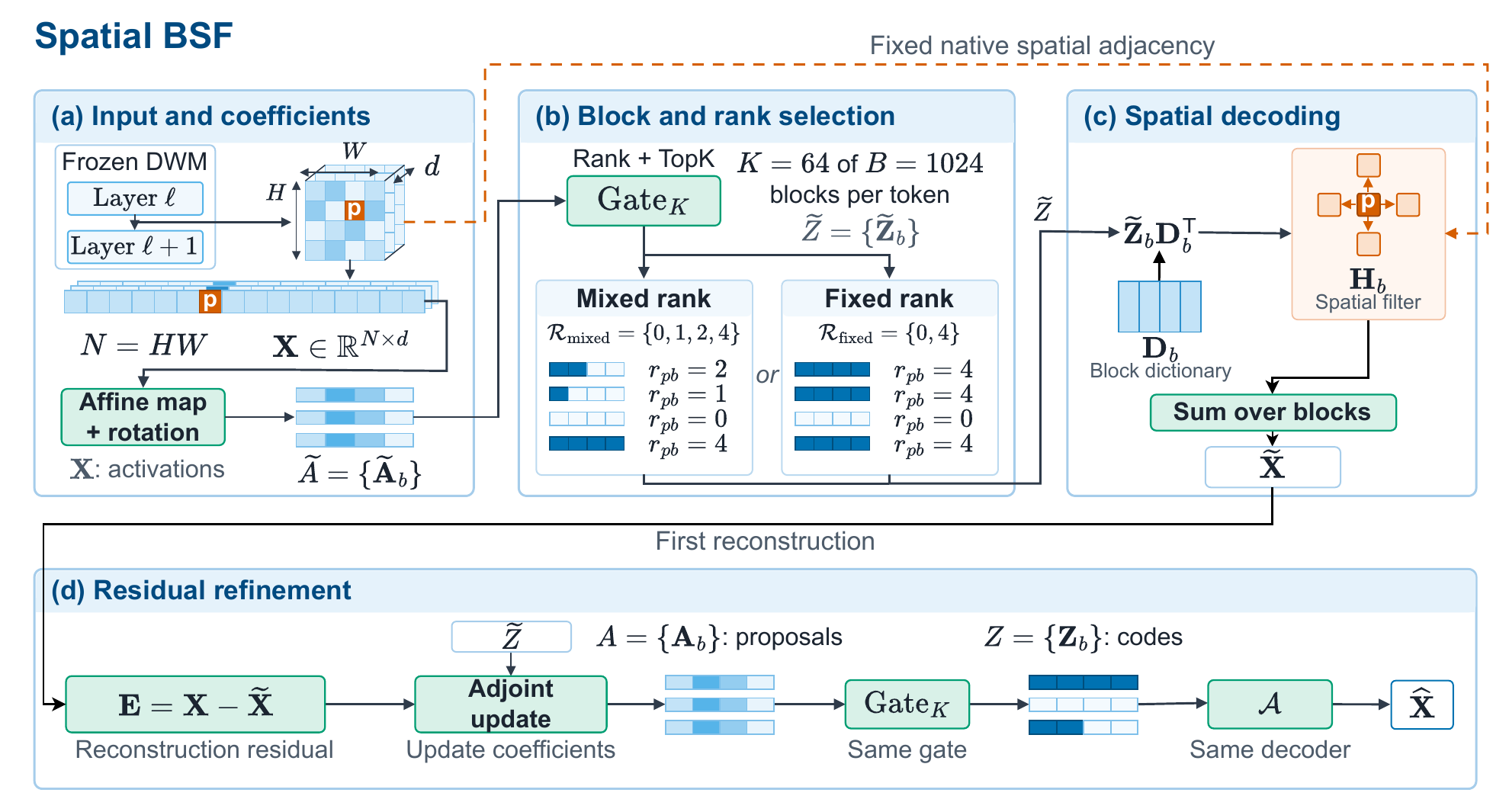}
 \caption{\textbf{spatialBSF architecture.}
 (a) A token-wise encoder produces coefficient proposals from visual activations $\mX$ (complete grid shown).
 (b) The gate retains $K$ blocks per token with fixed or adaptive ranks.
 (c) Dictionaries $\mD_b$ and native-neighbour operators $\mH_b$ reconstruct $\widetilde{\mX}$.
 (d) The residual drives an adjoint coefficient update, followed by reselection and decoding of the final reconstruction $\widehat{\mX}$.}
 \label{fig:spatialbsf-architecture}

\end{minipage}
\end{figure}

\subsection{EARS-Graph: semantic references and metrics}
\label{sec:ears-reference-metrics}
We introduce EARS-Graph, an image-aligned driving scene graph of typed entities (with geometry and attributes) and directed relations that preserves provenance and distinguishes unknown labels from negatives. Using this representation, we annotate 300 front-camera images from 50 drives across four cities. Semantic evaluation uses 232 fitting and 66 development-exposed validation images, excluding two original-test images (\suppref{sec:supp-interpretability-population}). We populate entities using map geometry and frozen perception models~\citep{robinson2026rfdetr,wu2024topomlp,wang2021fcos3d}, while inferring activity from image history. We derive relations via geometric rules and heads trained on OpenLane-V2~\citep{wang2023openlanev2}. We used the automatically generated graphs for evaluation and conducted a pre-release human review of selected AI-revised graphs (\suppref{sec:supp-ears-generation-pipeline,sec:supp-method-reference-details}). \Cref{fig:ears-reference}b shows a schematic EARS-Graph example; the graph schema, generation rules and relation definitions are detailed in \suppref{sec:supp-crs-schema,sec:supp-crs-generation,sec:supp-crs-relations}.

\noindent\textbf{Semantic evaluation.}
\Cref{fig:ears-reference}c illustrates the two evaluation paths: the orange response from one dictionary unit is compared with the outlined EARS-Graph car region for localisation ($F_\chi$). For relations ($A_{\mathrm{rel}}$) and states ($A_{\mathrm{state}}$), we pool the \terminologyedit{complete reconstruction} over EARS-Graph entity regions and fit linear readouts to these pooled features using ridge regression~\citep{hoerl_ridge_1970}. EARS-Graph relation and state labels provide the targets for these readouts; dictionary training uses only reconstruction. \terminologyedit{The conjunction is evaluated by applying a fixed AND to the ahead and shared-lane predictions.} All selected units, thresholds and readouts are frozen before evaluation on separate validation drives.
Localisation covers cars, large vehicles (the pixelwise union of truck and bus boxes), roads and traffic lights; large-vehicle grouping is exploratory.

\noindent\textbf{Summarising semantic agreement.}
We combine the task-group means into one descriptive summary of semantic agreement:
\begin{equation}
I_{\mathrm{EARS},\chi}=\frac{4F_\chi+5A_{\mathrm{rel}}+2A_{\mathrm{state}}}{11}.
\label{eq:ears-index}
\end{equation}
The eleven equally weighted tasks comprise the four localisation targets ($F_\chi$); ahead, left, lane membership, shared lane and the ahead-and-shared-lane conjunction ($A_{\mathrm{rel}}$); and red/green traffic lights and stationary/moving cars ($A_{\mathrm{state}}$). Relation and car-state targets retain their original car-only scope.
Only $F_\chi$ depends on the strength or direction readout $\chi$; $A_{\mathrm{rel}}$ and $A_{\mathrm{state}}$ are shared.
We average all nine fixed DWM--layer interfaces equally.
The index describes semantic agreement, not causal relevance.
Task definitions, fitting, scoring and interpretation limits appear in \suppref{sec:supp-interpretability}.

\subsection{Spatial Block-Sparse Featurisers (spatialBSF)}
\label{sec:spatialbsf}
DWM inputs contain large, spatially continuous structures, such as roads and other extended scene elements, that span multiple visual tokens. To capture spatial structure, spatialBSF extends BSFs~\citep{felStructuringSparsityBlockSparse2026} with learned decoder filters that couple block contributions across neighbouring positions without enforcing code similarity. A single residual correction then refines the initial coefficients and block selection based on the joint reconstruction error (\cref{fig:spatialbsf-architecture}). To help explain our method, we fix $B=1024$ dictionary blocks with maximum rank $r_{\max}=4$ and select $K=64$ blocks per token.

\noindent\textbf{Block coefficients.}
We follow vanilla BSF's encoder implementation~\citep{felStructuringSparsityBlockSparse2026}, using an untied affine encoder to map $\mX$ to initial coefficient proposals $\widetilde{\mA}_b\in\mathbb R^{N\times r_{\max}}$ for each block $b$ (\cref{fig:spatialbsf-architecture}a).
We denote the collections of our initial proposal and selected-coefficient matrices by $A=\{\mA_b\}_{b=1}^{B}$ and $Z=\{\mZ_b\}_{b=1}^{B}$. Each block has an orthonormal decoder basis $\mD_b\in\mathbb R^{d\times r_{\max}}$, satisfying $\mD_b^{\mathsf T}\mD_b=\mI_{r_{\max}}$. A learned orthogonal rotation, shared by the block's encoder and decoder, defines its ordered basis directions. Encoder and orthogonal-parameterisation details appear in \suppref{sec:supp-method-encoder}.

\noindent\textbf{Block selection.}
At each token, let $\va_b$ denote the coefficient proposal for block $b$. We score each block using its calibrated coefficient energy:
\begin{equation}
  s_b=\gamma_b\frac{\|\va_b\|_2^2}{\bar e}+v_b,
  \label{eq:main-block-score}
\end{equation}
where $\bar e$ is the mean full-block proposal energy across blocks, floored away from zero. We adjust the normalised energies using a learned affine transformation, analogous to vector scaling~\citep{guo2017calibration}, with a bounded positive scale $\gamma_b$ and offset $v_b$.
By default, spatialBSF retains all $r_{\max}=4$ coefficients of each selected block. Mixed-rank spatialBSF applies the same scoring rule—with rank-specific calibration—to candidate coefficient counts of $\{1, 2, 4\}$, selecting the highest-scoring option for each block. The gate then retains exactly the $K$ highest-scoring blocks and their chosen coefficients, setting all other coefficients to zero (\cref{fig:spatialbsf-architecture}b). We denote this operation by $Z=\operatorname{Gate}_K(A)$, with $\widetilde Z=\operatorname{Gate}_K(\widetilde A)$ for the initial pass. Full allocation equations, calibration parameterisation and tie rules appear in \suppref{sec:supp-method-allocation}.

\noindent\textbf{Spatial decoding.}
To let blocks contribute beyond their source positions, our decoder combines local and neighbouring token contributions (\cref{fig:spatialbsf-architecture}c):
\begin{equation}
 \begin{aligned}
 \mathcal A(Z)&=\sum_b\mH_b\mZ_b\mD_b^{\mathsf T},\\
 \mH_b&=\theta_{b0}\mI_N+\theta_{b1}\mS,
 \end{aligned}
 \label{eq:main-spatial-decoder}
\end{equation}
mapping $Z$ to an $N\times d$ activation matrix, where each block learns local and neighbour gains $\theta_{b0},\theta_{b1}$ and each summand defines the block response used for interpretation. The fixed spatial operator $\mS=\mL_{\mathrm{sym}}-\mI_N\in\mathbb R^{N\times N}$ uses the symmetric normalised graph Laplacian over native token coordinates: horizontal/vertical neighbours for DriveWAM and SimWAM, and symmetrised four-nearest neighbours for DriveLaW's retained positions. Placing this coupling in the decoder lets reconstruction loss learn where each block contributes—recovering token-wise decoding when $\mH_b=\mI_N$—while the $K$-block budget applies at sources and contributions can overlap at destinations (\suppref{sec:supp-method-spatial} details operator conventions and gain constraints).

\noindent\textbf{Residual refinement.}
Building on iterative sparse-inference methods~\citep{gregor_learning_2010}, we use one residual correction with block reselection.
Initial selection precedes evaluation of the blocks' joint reconstruction error.
The residual provides feedback for revising that selection (\cref{fig:spatialbsf-architecture}d):
\begin{equation}
 \begin{aligned}
 \mE&=\mX-\mathcal A(\widetilde Z),\\
 \mG_b&=\mH_b^{\mathsf T}\mE\mD_b,\\
 \mA_b&=\widetilde{\mZ}_b+\alpha\kappa_b\mG_b,\\
 Z&=\operatorname{Gate}_K(A),\quad
 \widehat{\mX}=\mathcal A(Z).
 \end{aligned}
 \label{eq:main-residual-refinement}
\end{equation}
The adjoint block $\mG_b$ maps the residual back through spatial transport and channel decoding, giving the negative coefficient gradient of $\tfrac12\|\mE\|_F^2$.
A field-specific step $\alpha$ and learned positive gains $\kappa_b$ scale the correction.
Reapplying the same gate can replace selected blocks and change mixed ranks. This single correction per forward pass shares gate and decoder parameters (see \suppref{sec:supp-method-refinement} for the step-size derivation); both stages use exact block-and-rank selection in the forward pass. During training, gradients for the selection scores are approximated by perturbing the scores and repeating the selection (\suppref{sec:supp-method-training}).
% Source: main-paper/sections/04-results.tex
\FloatBarrier
\section{Experiments and Results}
\label{sec:results}
\subsection{\resourceedit{Research Resources and Evaluation Protocol}}
\label{sec:evaluation-setup}
\resourceedit{\noindent\textbf{Research resources.}
Our collected visual-conditioning activations, trained MI checkpoints, and EARS-Graph semantic references form a reusable resource for comparing reconstruction fidelity and semantic agreement across driving world models (\cref{tab:research-resources}).
Activation capture, data splits, and reference provenance are detailed in \suppref{sec:supp-captured-representations,sec:supp-r2-populations,sec:supp-ears-dataset}.}

% Source: main-paper/tables/reconstruction-summary.tex
\begin{table}[!htbp]
\centering
\begin{minipage}{0.94\textwidth}

\color{black}
\hypersetup{allcolors=black}
\captionsetup{font+={color=black},labelfont+={color=black}}
\centering\footnotesize\setlength{\tabcolsep}{2pt}
\renewcommand{\arraystretch}{0.92}
\caption{\textbf{Three-layer reconstruction means ($\uparrow$) at a matched dictionary-unit budget.} Each tool selects 64 scalar features or blocks per token. \textit{Dataset-Selected} and \textit{Dataset-Full} evaluations have not been verified to share populations and $R^2$ denominators; panel differences therefore do not isolate the effect of training-set size. \textbf{Bold}, \underline{underline} and \thirdbest{italics} mark the top three per column and panel before rounding; exact ties share ranks.}
\label{tab:reconstruction}
\scalebox{0.95}{%
\begin{adjustbox}{max width=\linewidth}\begin{tabular}{@{}lrrrrrrrrrrrr@{}}
\toprule
MI tool & \multicolumn{3}{c}{DriveLaW} & \multicolumn{3}{c}{DriveWAM} & \multicolumn{3}{c}{SimWAM} & \multicolumn{3}{c}{Across-DWM mean}\\
\cmidrule(lr){2-4}\cmidrule(lr){5-7}\cmidrule(lr){8-10}\cmidrule(l){11-13}
& $R^2_{\mathrm{act}}$ & $R^2_{\mathrm{sp}}$ & $R^2_{\mathrm{temp}}$ & $R^2_{\mathrm{act}}$ & $R^2_{\mathrm{sp}}$ & $R^2_{\mathrm{temp}}$ & $R^2_{\mathrm{act}}$ & $R^2_{\mathrm{sp}}$ & $R^2_{\mathrm{temp}}$ & $R^2_{\mathrm{act}}$ & $R^2_{\mathrm{sp}}$ & $R^2_{\mathrm{temp}}$\\
\midrule
\multicolumn{13}{l}{\textbf{(a) Dataset-Selected}}\\
\addlinespace[1pt]
SAE & 0.7560 & 0.6336 & 0.4272 & 0.5520 & 0.4017 & 0.2717 & 0.4700 & 0.3602 & 0.2905 & 0.5927 & 0.4652 & 0.3298\\
Matryoshka TopK & 0.7501 & 0.6321 & 0.4141 & 0.5249 & 0.3781 & 0.2632 & 0.4466 & 0.3401 & 0.2777 & 0.5739 & 0.4501 & 0.3183\\
RA-SAE & 0.6467 & 0.5018 & 0.2322 & 0.3927 & 0.2313 & 0.1246 & 0.3234 & 0.2203 & 0.1598 & 0.4543 & 0.3178 & 0.1722\\
Vanilla BSF & \underline{0.7946} & \underline{0.6876} & \textbf{0.5308} & \thirdbest{0.6334} & \thirdbest{0.5064} & \thirdbest{0.3995} & \thirdbest{0.5702} & \thirdbest{0.4785} & \thirdbest{0.4157} & \underline{0.6661} & \underline{0.5575} & \underline{0.4486}\\
Grassmannian BSF & \textbf{0.7984} & \textbf{0.6890} & \underline{0.5253} & 0.6257 & 0.4941 & 0.3836 & 0.5634 & 0.4676 & 0.4037 & \thirdbest{0.6625} & \thirdbest{0.5503} & 0.4375\\
\midrule
spatialBSF (ours) & \thirdbest{0.7811} & \thirdbest{0.6693} & \thirdbest{0.5098} & \textbf{0.6492} & \textbf{0.5300} & \textbf{0.4346} & \textbf{0.5784} & \textbf{0.4871} & \textbf{0.4263} & \textbf{0.6696} & \textbf{0.5621} & \textbf{0.4569}\\
Mixed-rank spatialBSF (ours) & 0.7593 & 0.6350 & 0.4923 & \underline{0.6440} & \underline{0.5236} & \underline{0.4272} & \underline{0.5759} & \underline{0.4851} & \underline{0.4240} & 0.6597 & 0.5479 & \thirdbest{0.4479}\\
\midrule
\multicolumn{13}{l}{\textbf{(b) Dataset-Full}}\\
\addlinespace[1pt]
SAE & 0.8028 & 0.7090 & 0.5120 & 0.6576 & 0.5371 & 0.4030 & 0.5787 & 0.4857 & 0.4130 & 0.6797 & 0.5773 & 0.4427\\
Matryoshka TopK & 0.7986 & 0.7077 & 0.5052 & 0.6407 & 0.5191 & 0.3934 & 0.5635 & 0.4698 & 0.4006 & 0.6676 & 0.5655 & 0.4331\\
RA-SAE & 0.7852 & 0.6943 & 0.4874 & 0.6049 & 0.4781 & 0.3496 & 0.5239 & 0.4292 & 0.3625 & 0.6380 & 0.5338 & 0.3998\\
Vanilla BSF & \thirdbest{0.8498} & \underline{0.7775} & \textbf{0.6334} & \underline{0.7340} & \textbf{0.6391} & \textbf{0.5339} & \underline{0.6692} & \textbf{0.5939} & \textbf{0.5308} & \underline{0.7510} & \textbf{0.6702} & \textbf{0.5660}\\
Grassmannian BSF & 0.8486 & \thirdbest{0.7756} & \underline{0.6317} & 0.7238 & 0.6270 & \thirdbest{0.5252} & 0.6596 & \thirdbest{0.5837} & \thirdbest{0.5217} & 0.7440 & \thirdbest{0.6621} & \thirdbest{0.5595}\\
\midrule
spatialBSF (ours) & \textbf{0.8555} & \textbf{0.7793} & \thirdbest{0.6284} & \textbf{0.7375} & \underline{0.6387} & \underline{0.5268} & \textbf{0.6713} & \underline{0.5912} & \underline{0.5265} & \textbf{0.7548} & \underline{0.6698} & \underline{0.5606}\\
Mixed-rank spatialBSF (ours) & \underline{0.8518} & 0.7742 & 0.6186 & \thirdbest{0.7270} & \thirdbest{0.6273} & 0.5184 & \thirdbest{0.6615} & 0.5824 & 0.5181 & \thirdbest{0.7468} & 0.6613 & 0.5517\\
\bottomrule
\end{tabular}\end{adjustbox}
}

\end{minipage}
\end{table}

\begin{table}[!htbp]
\centering
\begin{minipage}{0.94\textwidth}

% Source: main-paper/tables/interpretability-summary.tex
\recenteditcolor\recenteditcaption
\centering\footnotesize\setlength{\tabcolsep}{1.7pt}
\caption{\textbf{Eleven-task EARS-Graph \terminologyedit{semantic agreement} ($\uparrow$),} averaged over nine interfaces at a matched dictionary-unit budget: 64 scalar features or blocks per token (64/256 coefficient slots for SAEs/rank-four BSFs; variable for mixed rank). $F_\chi$ averages car, large vehicle (truck$\cup$bus), road and traffic-light localisation. Ranking notation follows \cref{tab:reconstruction}.}
\label{tab:interpretability-summary}
\scalebox{0.9}{%
\begin{adjustbox}{max width=\linewidth}\begin{tabular}{@{}lrrrrrr@{}}
\toprule
MI tool & $A_{\mathrm{rel}}$ & $F_{\mathrm{str}}$ & $F_{\mathrm{dir}}$ & $A_{\mathrm{state}}$ & $I_{\mathrm{str}}$ & $I_{\mathrm{dir}}$\\
\midrule
\multicolumn{7}{l}{\textbf{(a) Dataset-Selected training}}\\
\addlinespace[2pt]
SAE & 0.7057 & 0.2519 & 0.2542 & 0.6962 & 0.5390 & 0.5398\\
Matryoshka TopK & 0.7277 & 0.2889 & 0.2889 & \textbf{0.7437} & \thirdbest{0.5711} & 0.5711\\
RA-SAE & 0.6996 & 0.1730 & 0.1730 & 0.6788 & 0.5043 & 0.5043\\
Vanilla BSF & \underline{0.7323} & 0.2803 & 0.2841 & 0.7014 & 0.5623 & 0.5637\\
Grassmannian BSF & 0.7202 & \thirdbest{0.2966} & \underline{0.3192} & \thirdbest{0.7292} & 0.5678 & \thirdbest{0.5760}\\
\midrule
spatialBSF & \thirdbest{0.7291} & \underline{0.3128} & \thirdbest{0.3175} & \underline{0.7335} & \underline{0.5785} & \underline{0.5802}\\
Mixed-rank spatialBSF & \textbf{0.7359} & \textbf{0.3160} & \textbf{0.3226} & 0.7184 & \textbf{0.5800} & \textbf{0.5824}\\
\midrule
\multicolumn{7}{l}{\textbf{(b) Dataset-Full training}}\\
\addlinespace[2pt]
SAE & 0.7136 & 0.2381 & 0.2416 & 0.7076 & 0.5396 & 0.5408\\
Matryoshka TopK & 0.7014 & \underline{0.3373} & 0.3373 & 0.6897 & 0.5669 & 0.5669\\
RA-SAE & 0.7139 & 0.2126 & 0.2126 & \underline{0.7459} & 0.5374 & 0.5374\\
Vanilla BSF & 0.7115 & 0.3288 & \thirdbest{0.3390} & \thirdbest{0.7379} & 0.5772 & 0.5809\\
Grassmannian BSF & \thirdbest{0.7235} & 0.3305 & \underline{0.3460} & \textbf{0.7607} & \underline{0.5874} & \textbf{0.5930}\\
\midrule
spatialBSF & \underline{0.7236} & \textbf{0.3476} & \textbf{0.3492} & 0.7348 & \textbf{0.5889} & \underline{0.5895}\\
Mixed-rank spatialBSF & \textbf{0.7252} & \thirdbest{0.3372} & 0.3362 & 0.7152 & \thirdbest{0.5823} & \thirdbest{0.5819}\\
\bottomrule
\end{tabular}\end{adjustbox}%
}

\end{minipage}
\end{table}

\resourceedit{\noindent\textbf{Comparison protocol.}
We compare 4,096-feature SAEs—signed SAE, Matryoshka TopK~\citep{bussmann2025matryoshka}, and relaxed archetypal SAE (RA-SAE)~\citep{fel2025archetypal}—with 1,024-block Vanilla, Grassmannian, spatialBSF, and mixed-rank spatialBSF dictionaries.
Following BSF's precedent in comparing dictionary units directly~\citep{felStructuringSparsityBlockSparse2026}, the matched dictionary-unit budget selects 64 scalar features or whole blocks per source token: 64/256 coefficient slots for SAEs/rank-four BSFs, variable for mixed rank.
Spatial transport and refinement costs appear in \suppref{sec:supp-dictionary-size,sec:supp-bsf-ablation}.}

\resourceedit{Within each dataset/interface, tools share fitting activations, fitting-only channel-mean/RMS normalisation, Adam~\citep{kingma2015adam}, and seed 42.
\textit{Dataset-Selected} fixes 2,400 updates with a shared learning-rate schedule; imported \textit{Dataset-Full} dictionaries use best reconstruction-validation checkpoints under a 30-epoch cap.
We evaluate delivered inference without post-encoding optimisation.
Training and mean-activity variants are detailed in \suppref{sec:supp-training-protocol,sec:supp-sae-baselines,sec:supp-lasso-average-budget}.}

\subsection{Results and Discussion}
\noindent\textbf{Fidelity and \terminologyedit{Semantic Agreement}}
\label{sec:reconstruction-results}
\label{sec:semantic-results}
High reconstruction fidelity does not establish that dictionary units capture meaningful scene concepts~\citep{gao_scaling_nodate}. EARS tests whether individual units localise the four selected targets and whether their combined reconstruction supports the specified relation, composition, and state readouts. Fidelity and semantic agreement can diverge: across both datasets, Matryoshka TopK trails the signed SAE on all three reconstruction metrics yet achieves higher eleven-task EARS scores (\cref{tab:reconstruction,tab:interpretability-summary}). Mixed-rank spatialBSF leads \textit{Dataset-Selected} (strength/direction: 0.5800/0.5824); \textit{Dataset-Full} leaders are spatialBSF for strength (0.5889) and Grassmannian BSF for direction (0.5930).

\begin{figure}[!htbp]
\centering
\begin{minipage}{0.90\textwidth}

\centering
% Source: main-paper/figures/qualitative-localization-layout.tex
\begingroup
\setlength{\unitlength}{1bp}
\resizebox{0.9\textwidth}{!}{%
\begin{picture}(504.000000,172.777500)
\put(0,0){\includegraphics[width=504bp]{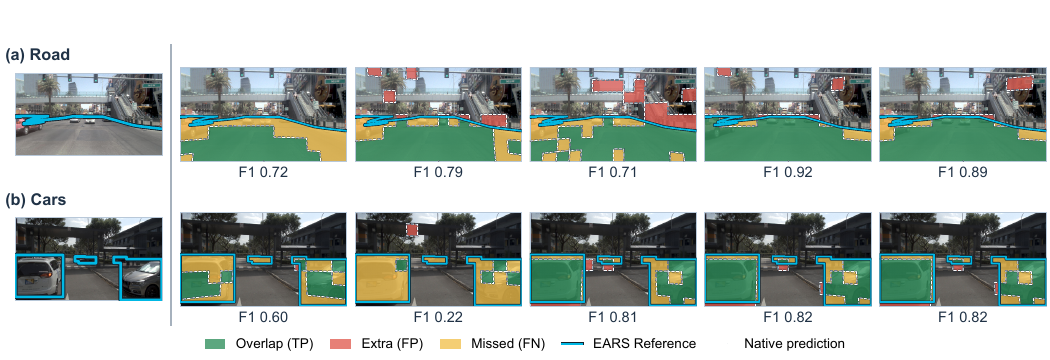}}
\put(42.510000,169.537500){\makebox(0,0)[t]{\fontsize{8}{8.5}\selectfont\shortstack{Raw image\\+ EARS Reference}}}
\put(126.450000,169.537500){\makebox(0,0)[t]{\fontsize{8}{8.5}\selectfont\shortstack{SAE~\citep{gao_scaling_nodate}}}}
\put(210.390000,169.537500){\makebox(0,0)[t]{\fontsize{8}{8.5}\selectfont\shortstack{Vanilla BSF\\\citep{felStructuringSparsityBlockSparse2026}}}}
\put(294.330000,169.537500){\makebox(0,0)[t]{\fontsize{8}{8.5}\selectfont\shortstack{Grassmannian BSF\\\citep{felStructuringSparsityBlockSparse2026}}}}
\put(378.270000,169.537500){\makebox(0,0)[t]{\fontsize{8}{8.5}\selectfont\shortstack{\textbf{spatialBSF}\\\textbf{(ours)}}}}
\put(462.210000,169.537500){\makebox(0,0)[t]{\fontsize{8}{8.5}\selectfont\shortstack{\textbf{mixed-rank}\\\textbf{spatialBSF (ours)}}}}
\end{picture}%
}
\endgroup

\recenteditcaption
\caption{\textbf{Localisation with one dictionary unit.} Selected exposed-validation road/car examples from two of the four localisation targets, using DriveWAM layer~15 strength readouts. Smaller raw images precede the divider. Cyan outlines: EARS Reference boxes/polygons; dashed: native predictions. Green/coral/yellow show TP/FP/FN area, coloured using the EARS Reference after prediction. F1 scores the complete frame.}
\label{fig:qualitative-localization}

\end{minipage}
\end{figure}

\noindent\textbf{Spatial Transport Improves Road Localisation.}
In the \textit{Dataset-Selected} ablation, spatialBSF's spatial decoder improves mean road strength F1 from 0.5821 with a local decoder to 0.6493, averaged across all nine interfaces (\supptableref{tab:supp-ablation-portfolio-tasks}). Qualitative examples also show more complete road coverage (\Cref{fig:qualitative-localization}). Additional experiments test the readout's sensitivity to neighbour placement while keeping dictionary-unit activations and readouts fixed. Norm-preserving neighbour permutations reduce mean road strength F1 from 0.6493 to 0.5087 for \textit{Dataset-Selected} and from 0.6748 to 0.5060 for \textit{Dataset-Full}. All eight permutations reduce F1 at every \textit{Dataset-Selected} interface. These results support the importance of spatial placement for localising projected-road regions; further controls and their interpretation appear in \suppref{sec:supp-road-spatial-transport}.

\noindent\textbf{Duality of Residual Correction}
spatialBSF leads activation fidelity but not spatial or temporal fidelity on \textit{Dataset-Full} (\cref{tab:reconstruction}). On 100 shared \textit{Dataset-Selected} validation scenes, correction removes 74.5\% of initial activation error for \textit{Dataset-Full}-trained models versus 6.6\% for \textit{Dataset-Selected} (\suppref{sec:supp-residual-correction-diagnostic}). Yet correction improves activation $R^2$ by 7.3\% and EARS by 4.1–4.7\% in the component ablation (\cref{tab:bsf-ablation}). High final fidelity can thus conceal weak initial reconstruction. Encoder fine-tuning improves both EARS indices at modest fidelity cost; joint supervision improves initial reconstruction more but lowers EARS (\suppref{sec:supp-residual-correction-control}). Balancing the stages therefore matters alongside retaining correction. We use the checkpoints before additional fine-tuning in the main comparison, as other methods do not receive this optimisation.

% Source: main-paper/figures/qualitative-localization.tex

% Source: main-paper/figures/simwam-application-demo.tex
\begin{figure}[!htbp]
\centering
\begin{minipage}{0.90\textwidth}

\centering
\includegraphics[width=0.85\textwidth,height=0.50\textheight,keepaspectratio]{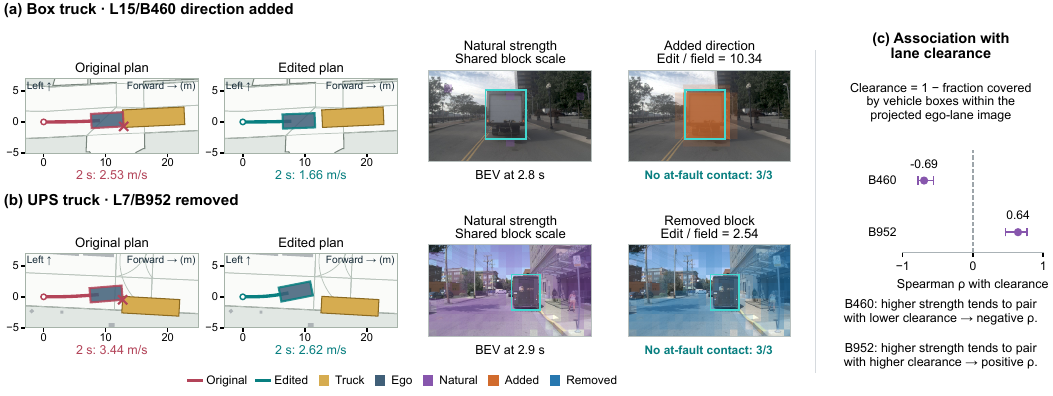}
\recenteditcaption
\captionsetup{hypcap=false}
\caption{\textbf{Intervening on blocks shortlisted with EARS-Graph.}
EARS-Graph shortlists the blocks: (a) adds a direction within a car-associated block; (b) removes a road-associated block contribution.
Purple natural-source norms share scales with \cref{fig:simwam-natural-references}; orange/blue show added/removed decoded contributions, separately scaled.
Edit/field: consumed-change/original-field Frobenius-norm ratio.
Both recipes remove at-fault contact on seeds 42/45/46.
(c) Associations with lane clearance: higher strength tends to accompany lower clearance for B460 ($\rho<0$), higher clearance for B952 ($\rho>0$).
Equal-drive Spearman estimates and 95\% drive-bootstrap intervals (shown in the figure) use 122 fitting scenes/35 drives (\suppref{sec:supp-demo-lane-clearance}).}
\label{fig:simwam-application}

\end{minipage}
\end{figure}

% Source: main-paper/tables/research-resources.tex
\begin{table}[!htbp]
\centering
\begin{minipage}{0.94\textwidth}

\centering\footnotesize
\setlength{\tabcolsep}{3pt}
\renewcommand{\arraystretch}{0.95}
\captionsetup{skip=4pt}
\resourceeditcolor
\ifmarkresourceedits
  \captionsetup{font+={color=red},labelfont+={color=red}}
\fi
\caption{\resourceedit{\textbf{Research resources.}
Scenes: fitting/validation; storage: decimal GB; layers: zero-indexed.}}
\label{tab:research-resources}
\begin{adjustbox}{max width=\linewidth}\begin{tabular}{@{}p{0.29\columnwidth}p{\dimexpr0.71\columnwidth-6pt\relax}@{}}
\toprule
\raggedright\textbf{Resource} & \raggedright\textbf{Coverage and availability}\tabularnewline
\midrule
\raggedright DWM interfaces & \raggedright DriveLaW~\citep{xia_drivelaw_2026}: 14/18/24; DriveWAM~\citep{shi_drivewam_2026}, SimWAM~\citep{zhao_simwam_2026}: 7/15/23 (nine total).\tabularnewline
\raggedright Dataset-Selected & \raggedright 800 / 100 scenes; 49.77\,GB of cached FP32 activations.\tabularnewline
\raggedright Dataset-Full & \raggedright 68,538 / 6,653 scenes; 3.8Tb of cached FP32 activations.\tabularnewline
\raggedright Primary MI checkpoints & \raggedright 126: seven methods $\times$ nine interfaces $\times$ two regimes (63 per regime).\tabularnewline
\raggedright EARS-Graph & \raggedright 300 scenes/50 drives; eleven tasks; derived references with documented provenance.\tabularnewline
\raggedright Code and access & \raggedright Evaluation code, configurations, metadata; resources available soon, subject to upstream terms.\tabularnewline
\bottomrule
\end{tabular}\end{adjustbox}

\end{minipage}
\end{table}

\noindent\textbf{Layer-Dependent Semantic Accessibility and MI Tool Performance.}
\label{sec:results-layers-models}
For \textit{Dataset-Full}-trained MI tools evaluated on the shared EARS cohort, early road localisation (L7) and middle-layer relation/motion-state accessibility (L15) recur across SAE and BSF family means in DriveWAM and SimWAM (\supptableref{tab:supp-family-layer-means}). This pattern indicates task-dependent semantic accessibility across the sampled layers~\citep{lecun2015deep}. BSFs achieve higher mean car-localisation scores (the main object in most driving scenes) than SAEs at every sampled layer in both models. This is consistent with visual concepts spanning multiple coordinated directions~\citep{felStructuringSparsityBlockSparse2026}; grouping these directions within a block may make object features easier to recover from a single unit.

% Source: main-paper/sections/05-ablation.tex
\FloatBarrier
\section{Ablation Study}
\label{sec:ablation-study}
The \textit{Dataset-Selected} component ablation fixes 1,024 rank-four blocks and 64 selected blocks (256 coefficient slots) across configurations, measuring component contributions to \terminologyedit{semantic agreement} and reconstruction fidelity (\cref{tab:bsf-ablation}).
\begin{table}[!htbp]
\centering
\begin{minipage}{0.94\textwidth}

% Source: main-paper/tables/component-ablation.tex
\recenteditcolor\recenteditcaption
\centering\scriptsize\setlength{\tabcolsep}{1.5pt}
\captionsetup{font+=footnotesize}
\caption{\textbf{spatialBSF component ablation on \textit{Dataset-Selected},} averaged over nine interfaces. Enc./Spat./Res.: independent encoding, neighbour transport and residual refinement/reselection. $I_{\mathrm{str}},I_{\mathrm{dir}}$ use eleven equally weighted EARS tasks. Bold marks the best score per column.}
\label{tab:bsf-ablation}
\resizebox{\linewidth}{!}{%
\begin{adjustbox}{max width=\linewidth}\begin{tabular}{@{}lcccrrrrr@{}}
\toprule
 & \multicolumn{3}{c}{Components} & \multicolumn{3}{c}{Reconstruction ($\uparrow$)} & \multicolumn{2}{c}{\terminologyedit{Semantic agreement} ($\uparrow$)}\\
\cmidrule(lr){2-4}\cmidrule(lr){5-7}\cmidrule(l){8-9}
Configuration & Enc. & Spat. & Res. & $R^2_{\mathrm{act}}$ & $R^2_{\mathrm{sp}}$ & $R^2_{\mathrm{temp}}$ & $I_{\mathrm{str}}$ & $I_{\mathrm{dir}}$\\
\midrule
spatialBSF & $\checkmark$ & $\checkmark$ & $\checkmark$ & 0.6696 & 0.5621 & 0.4569 & \textbf{0.5785} & \textbf{0.5802}\\
Tied encoder & \textemdash & $\checkmark$ & $\checkmark$ & \textbf{0.6841} & \textbf{0.5838} & \textbf{0.4822} & 0.5628 & 0.5717\\
Local decoder & $\checkmark$ & \textemdash & $\checkmark$ & 0.6676 & 0.5621 & 0.4573 & 0.5724 & 0.5757\\
One stage & $\checkmark$ & $\checkmark$ & \textemdash & 0.5965 & 0.4638 & 0.3183 & 0.5319 & 0.5392\\
Local + one stage & $\checkmark$ & \textemdash & \textemdash & 0.5877 & 0.4475 & 0.3088 & 0.5277 & 0.5303\\
\bottomrule
\end{tabular}\end{adjustbox}
}

\end{minipage}
\end{table}

\noindent\textbf{Ablation Summary.}
Spatial decoding and residual refinement play complementary roles: spatial decoding couples neighbouring tokens, while residual refinement accounts for the larger reconstruction improvement in this ablation. Tied encoding, as in Grassmannian BSF~\citep{felStructuringSparsityBlockSparse2026}, improves reconstruction but lowers both EARS indices (\cref{tab:bsf-ablation}). Detailed component scores and costs appear in \suppref{sec:supp-bsf-ablation}.

\noindent\textbf{Score Calibration at Fixed Rank.}
Motivated by the difficulties of selecting among different block dimensions identified by \citet{jerpelea_deeper_2026}, we introduce rank-specific score calibration for mixed-rank spatialBSF. We retain the same scoring formulation in fixed-rank spatialBSF for consistency across variants. The fixed-rank comparison does not establish consistent superiority over raw-energy TopK across interfaces (\suppref{sec:supp-score-calibration}).

\noindent\textbf{Fixed and Mixed Rank.}
spatialBSF retains rank four, while mixed-rank spatialBSF selects rank-1/2/4 prefixes at the same 64-block budget. Mixed rank lowers all three reconstruction means on both datasets (\cref{tab:reconstruction}). EARS strength/direction rise slightly on \textit{Dataset-Selected} (0.5785/0.5802 to 0.5800/0.5824) but fall on \textit{Dataset-Full} (0.5889/0.5895 to 0.5823/0.5819; \cref{tab:interpretability-summary}). These results are inconclusive as to whether our mixed-rank implementation allows features of different ranks to coexist. Further experiments are reported in \suppref{sec:supp-fixed-mixed-rank}.
% Source: main-paper/sections/06-application.tex
\FloatBarrier
\section{Intervention case studies in SimWAM}
\label{sec:simwam-application}

\begin{figure}[!htbp]
\centering
\begin{minipage}{0.90\textwidth}

\centering
\includegraphics[width=0.75\columnwidth,height=0.50\textheight,keepaspectratio]{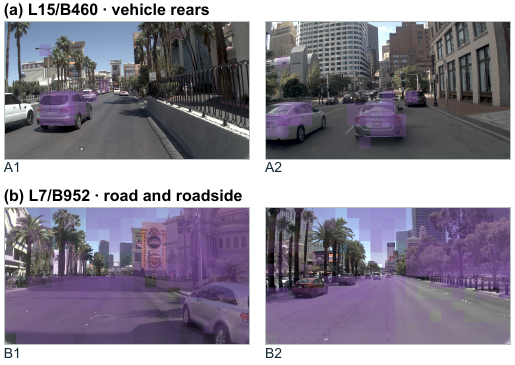}
\recenteditcaption
\caption{\textbf{Visual concepts associated with the selected blocks.}
Examples from two distinct fitting drives per block show responses to car rears (A1--A2) and road/roadside structures (B1--B2).}
\label{fig:simwam-natural-references}

\end{minipage}
\end{figure}

We start from SimWAM predictions that lead to at-fault collisions in simulation, selecting two cases to demonstrate complementary spatialBSF interventions: adding a within-block direction and removing a block contribution (\cref{fig:simwam-application}). EARS-Graph car and road localisation scores shortlist candidate blocks from fitting drives; natural examples support retrospective interpretation (\cref{fig:simwam-natural-references}).

% Source: main-paper/figures/simwam-natural-references.tex

\noindent\textbf{Case 1: addition within a car-associated block.}
At layer~15, block~460 ranks first for car localisation and highlights car rears.
Adding a direction within this block slows the approach and avoids the box-truck collision.

\noindent\textbf{Case 2: removal of a road-associated block.}
At layer~7, block~952 ranks third for road localisation; its strength correlates positively with visual ego-lane clearance ($\rho=0.64$).
Removing its contribution ($\mZ_{952}\leftarrow\mathbf 0$) slows and redirects the approach, avoiding the UPS-truck collision.

Both recipes remove at-fault collisions and pass road, driving-direction, time-to-collision and comfort checks across three action seeds per scene, with progress 1.00 and approximately 0.75, respectively.
Recipes were frozen before two fresh-noise seeds on the same scenes; each condition evaluates one four-second plan without re-planning.
The edits are substantial: consumed-change/native-field Frobenius-norm ratios are 10.34 and 2.54, respectively (\cref{fig:simwam-application}).
We use EARS-Graph car and road localisation scores to identify candidate spatialBSF blocks for intervention in SimWAM. These demonstrations use recipes selected through exploratory searches on fitting scenes; the search procedure, case-selection details and limitations are described in \suppref{sec:supp-application-demo}.

% Source: main-paper/sections/07-conclusion.tex
\FloatBarrier
\section{Conclusion}
\label{sec:conclusion}

We introduce spatialBSF and EARS-Graph as complementary tools for investigating the content and behavioural relevance in DWMs' internal representations.
Across three frozen DWMs, spatialBSF achieves the highest mean activation reconstruction fidelity, while EARS-Graph enables structured evaluation of localisation, relations, and object states.
Semantic rankings depend on the training-data regime and readout.
EARS scores summarise agreement with semantic references (\suppref{sec:supp-interpretability-aggregation}).
Preliminary interventions on blocks shortlisted using EARS-Graph prevent simulated at-fault collisions in two selected SimWAM fitting scenes, preparing for broader applications to improve models and driving safety.
Future work can build on our provided spatialBSF checkpoints, semantic references, and evaluation code to study spatiotemporal representations, incorporate richer scene semantics, and examine all DWM layers, while assessing the causal specificity of interventions and their generalisation to unseen driving scenarios.

% Source: main-paper/sections/08-llm-usage.tex
\section*{LLM Usage Statement}
We used OpenAI Codex to polish the writing, assist with parts of the code implementation, and help prepare preliminary figures from the results. All AI-assisted content was reviewed by the authors, who take full responsibility.

\FloatBarrier
\bigskip
\section*{Supplementary material}
\setlength{\parskip}{0pt}
% Source: supplementary/content.tex
\markrecenteditsfalse
\markterminologyeditsfalse
\markresourceeditsfalse
\color{black}
\appendix

\raggedbottom
We detail the representations, evaluation and spatialBSF method, then examine reconstruction, semantic agreement, component effects and interventions.\par
Bold, underline and italics mark the top 3 distinct displayed values; ties share a rank. Captions specify the comparison and metric direction. Rankings are descriptive.
Settings, counts and diagnostic quantities are unranked unless stated otherwise.

% Source: supplementary/sections/01-representations.tex
\FloatBarrier
\section{Driving world model representations}
\label{sec:supp-implementation}

We compare dictionary-based mechanistic interpretability (MI) tools at 9 interfaces across 3 driving world models (DWMs).
We explain layer selection, capture points and the contexts represented.
A dictionary unit is 1 scalar SAE feature or 1 whole BSF block.
Selected units count allocated slots; active units have nonzero coefficients.
% Source: supplementary/sections/representations/01-layer-selection.tex
\subsection{Layer selection}
\label{sec:supp-layer-selection}
\ifdefined\includesupplementary
The selected interfaces are specified in \cref{sec:evaluation-setup}.
\fi
We sample three layers per DWM to compare reconstruction and semantic agreement across network depth. Visual features vary with depth in vision transformers~\citep{ghiasi_vision_2022}. DriveWAM and SimWAM use layers 7, 15 and 23, spaced equally within their 30-layer stacks. For DriveLaW, earlier explorations decoded task-relevant information at layer 18, including vehicle counts within 40 m ahead. We bracket this reference with layers 14 and 24. Indices are 0-based and count transformer layers, excluding tokenisers, input embeddings and output heads. ``Native'' denotes the DWM's original activations, coordinates or computation.

% Source: supplementary/sections/representations/02-capture-boundaries.tex
\subsection{Capture points in the layer}
\label{sec:supp-captured-representations}
We collect visual tokens $\mC$ before their key and value projections for action computation. \paragraph{DriveLaW: Video-to-planner cross-attention.}
DriveLaW receives 4 camera frames spanning approximately 2 s, the current vehicle state and navigation commands.
Its video generator and diffusion action planner contain 28 paired visual and action layers~\citep{xia_drivelaw_2026}.
At layer $\ell$, the 2,048-channel visual output $\mC$ supplies the corresponding action layer's cross-attention keys and values.

\paragraph{DriveWAM: Shared transformer layers.}
Our DriveWAM configuration uses 1 current camera frame and 4 historical ego poses for 1 prediction horizon.
The architecture also supports longer autoregressive rollouts with cached attention memory.
Video and action share 30 transformer layers, with modality-specific input, output and time-conditioning paths~\citep{shi_drivewam_2026}.
We capture the 3,072-channel visual state $\mC$ after normalisation and time modulation, immediately before key and value projection.

\paragraph{SimWAM: Separate experts with mixed attention.}
SimWAM builds its visual key/value cache from 1 current camera frame.
Actions use this cache, the current vehicle state and command; this path generates no future video.
Separate video and action experts each contain 30 transformer layers~\citep{zhao_simwam_2026}.
We capture the 3,072-channel video state $\mC$ after normalisation and modulation, before key and value projection.
Action queries attend jointly to cached visual keys and values and current action keys and values.

% Source: supplementary/sections/representations/03-interpretation-limits.tex
\subsection{Activation collection during inference and limitations}
\label{sec:supp-interface-implications}
During the first action-denoising step, we collect the conditioning visual tokens and log their source context and original spatial positions. DriveLaW contributes 6 latent frames: 4 initialised from observed-frame VAE encodings and 2 from Gaussian noise. These labels describe initialisation, not physical timestamps. DriveWAM contributes observed-frame and generated-rollout tokens; SimWAM contributes only observed-frame tokens. We distinguish DriveLaW's unconditioned slots from DriveWAM's generated contexts.

Architecture and capture differences affect cross-model comparisons. Similar layer indices need not represent equivalent processing stages; DriveLaW also samples later depths. DriveLaW is captured at layer output, whereas DriveWAM and SimWAM are captured within a layer. Channel width, context and capture point can affect reconstruction difficulty. We therefore compare MI tools within each interface using common normalisation, scenes and token positions. Reconstruction measures representation recovery, not driving performance.

\FloatBarrier
% Source: supplementary/sections/02-reconstruction-evaluation.tex
\FloatBarrier
\section{Reconstruction evaluation}
\label{sec:supp-reconstruction-evaluation}

We measure the reconstruction of both raw activation values and their spatial and temporal differences to provide a holistic view of reconstruction quality. In this section, we define the evaluation populations, metrics, and averaging rules.

\begingroup
\color{black}\hypersetup{allcolors=black}
% Source: supplementary/sections/reconstruction-evaluation/01-population-and-normalization.tex
\subsection{Evaluation data, normalisation and activation fidelity}
\label{sec:supp-r2-populations}
Activation fidelity measures recovery of feature vectors.
Spatial and temporal fidelity measure recovery of signed differences~\citep{perez_poisson_2003}~\citep{isobe_look_2022}.
A reconstruction can preserve activation magnitude while smoothing spatial contrasts or temporal changes.
Difference metrics detect this loss.

\paragraph{Dataset-Selected.}
Dataset-Selected uses 800 fitting scenes and 100 validation scenes per interface. The validation scenes span 8 drives, disjoint from fitting drives. Another 100 test scenes are excluded. Validation scenes were used during method development and are shared across layers and MI tools within each DWM. Activation and spatial scores use all validation scenes; temporal scores use the pairs defined in \cref{sec:supp-temporal-metric}.

\paragraph{Dataset-Full.}
Dataset-Full uses 68,538 fitting scenes from 109 drives and 6,653 validation scenes from 8 other drives. All 9 interfaces share these populations. Temporal evaluation uses 5,542 adjacent-decision pairs. 

\begin{recentedits}
\paragraph{Notation and normalisation.}
At each interface, $i\in\mathcal I$ indexes validation scenes, $c\in\mathcal C$ retained contexts and $p=1,\ldots,N$ retained spatial tokens.
A context identifies a visual field, such as an observed or generated image slot.
$\mX_{ic}\in\mathbb R^{N\times d}$ contains normalised activation rows $\vx_{icp}=(\mX_{ic})_{p,:}$.
The corresponding final reconstruction row is $\widehat{\vx}_{icp}=(\widehat{\mX}_{ic})_{p,:}$.
We use the same channel centring and global scaling as BSF~\citep{felStructuringSparsityBlockSparse2026}.
For a raw activation row $\vc_{icp}$, normalisation is
\begin{equation}
 \vx_{icp}=\frac{\vc_{icp}-\vmu_{\mathrm{fit}}}{\sigma_{\mathrm{fit}}},
 \label{eq:supp-activation-normalization}
\end{equation}
Here $\vmu_{\mathrm{fit}}$ is the channel mean and $\sigma_{\mathrm{fit}}$ is a scalar root-mean-square (RMS) scale. Both use fitting tokens from the corresponding dataset and interface. Targets and reconstructions share these coordinates. The support tables give context and vector counts (\cref{tab:supp-r2-support,tab:supp-r2-support-full}).

\paragraph{Activation fidelity.}
We pool squared errors and centred target energies over all validation rows using 1 channel-wise validation mean. \begin{equation}
\begin{aligned}
 \vmu^{\mathrm{act}}&=\frac{1}{|\mathcal I|\,|\mathcal C|\,N}\sum_{i,c,p}\vx_{icp},\\
 R^2_{\mathrm{act}}&=1-
 \frac{\sum_{i,c,p}\|\vx_{icp}-\widehat{\vx}_{icp}\|_2^2}
 {\sum_{i,c,p}\|\vx_{icp}-\vmu^{\mathrm{act}}\|_2^2}.
\end{aligned}
\label{eq:supp-activation-r2}
\end{equation}
Unless stated otherwise, sums cover all validation scenes, retained contexts and token positions. A score of 1 indicates exact recovery. A score of 0 matches the validation-mean predictor; negative scores are worse.

\begin{table}[!htbp]
\centering
\begin{minipage}{0.94\textwidth}

\color{black}\hypersetup{allcolors=black}
\captionsetup{font+={color=black},labelfont+={color=black}}
\centering\small\setlength{\tabcolsep}{5pt}
\caption{\textbf{Dataset-Selected reconstruction evaluation counts per layer.} Counts are shared across each DWM's 3 layers and all dictionary methods. Each vector contains $d$ channels. Temporal counts use 56 adjacent-decision pairs and observed contexts only.}
\label{tab:supp-r2-support}
\begin{adjustbox}{max width=\linewidth}\begin{tabular}{@{}lrrrr@{}}
\toprule
DWM & Retained contexts & Activation vectors & Spatial-difference vectors & Temporal-difference vectors\\
\midrule
DriveLaW & 6 & 153,600 & 328,800 & 57,344\\
DriveWAM & 2 & 22,400 & 40,400 & 6,272\\
SimWAM & 1 & 25,200 & 47,100 & 14,112\\
\bottomrule
\end{tabular}\end{adjustbox}

\end{minipage}
\end{table}

\begin{table}[!htbp]
\centering
\begin{minipage}{0.94\textwidth}

\color{black}\hypersetup{allcolors=black}
\captionsetup{font+={color=black},labelfont+={color=black}}
\centering\small\setlength{\tabcolsep}{5pt}
\caption{\textbf{Dataset-Full reconstruction evaluation counts per layer.} Counts cover 6,653 validation scenes. Counts are shared across each DWM's 3 layers and all dictionary methods. Each vector contains $d$ channels. Temporal counts use 5,542 adjacent-decision pairs and observed contexts only.}
\label{tab:supp-r2-support-full}
\begin{adjustbox}{max width=\linewidth}\begin{tabular}{@{}lrrrr@{}}
\toprule
DWM & Retained contexts & Activation vectors & Spatial-difference vectors & Temporal-difference vectors\\
\midrule
DriveLaW & 6 & 10,219,008 & 21,875,064 & 5,675,008\\
DriveWAM & 2 & 1,490,272 & 2,687,812 & 620,704\\
SimWAM & 1 & 1,676,556 & 3,133,563 & 1,396,584\\
\bottomrule
\end{tabular}\end{adjustbox}

\end{minipage}
\end{table}

% Source: supplementary/sections/reconstruction-evaluation/02-spatial-differences.tex
\subsection{Spatial activation-difference fidelity}
\label{sec:supp-spatial-metric}

The spatial metric tests whether reconstruction preserves differences between activation vectors at neighbouring token positions in the same visual field.
Let $\mathcal E_g$ list the ordered neighbouring token pairs of type $g$, such as horizontal or vertical neighbours.
For each pair, we compare the difference between the original activation vectors with the corresponding difference after reconstruction:
\begin{equation}
\begin{aligned}
 \vdelta^{\mathrm{sp}}_{ic,pq}&=\vx_{icq}-\vx_{icp},\\
 \widehat{\vdelta}^{\mathrm{sp}}_{ic,pq}&=\widehat{\vx}_{icq}-\widehat{\vx}_{icp},
 \qquad (p,q)\in \mathcal E_g.
\end{aligned}
\label{eq:supp-spatial-increments}
\end{equation}
For each context and edge type, we average target differences across validation scenes and edges.
\begin{equation}
 \vmu^{\mathrm{sp}}_{cg}=\frac{\sum_{i\in\mathcal I}\sum_{(p,q)\in \mathcal E_g}\vdelta^{\mathrm{sp}}_{ic,pq}}
 {|\mathcal I|\,|\mathcal E_g|}.
 \label{eq:supp-spatial-mean}
\end{equation}
We pool squared errors and centred target energies across scenes, contexts and edge types:
\begin{equation}
 R^2_{\mathrm{sp}}=1-
 \frac{\sum_{i,c,g}\sum_{(p,q)\in\mathcal E_g}\|\vdelta^{\mathrm{sp}}_{ic,pq}-\widehat{\vdelta}^{\mathrm{sp}}_{ic,pq}\|_2^2}
 {\sum_{i,c,g}\sum_{(p,q)\in\mathcal E_g}\|\vdelta^{\mathrm{sp}}_{ic,pq}-\vmu^{\mathrm{sp}}_{cg}\|_2^2}.
 \label{eq:supp-spatial-r2}
\end{equation}
Centring uses 1 mean per context and edge type. DriveWAM's 8 $\times 14$ grid has 104 rightward and 98 downward edges. SimWAM's $12\times21$ grid has 240 rightward and 231 downward edges. DriveLaW uses a symmetrised 4-nearest-neighbour graph on 256 positions retained from its native 24 $\times 42$ grid. Squared native-coordinate distance selects neighbours; flattened grid index breaks ties. Its 548 unique edges point in increasing row-first, then column order and form 1 edge type. These orientations define centring. Differences use retained positions without interpolation or division by edge length.

\paragraph{Interpreting spatial fidelity.}
Higher spatial $R^2$ means better recovery of local contrasts, including their magnitude and sign. Distortion can suppress or exaggerate these contrasts. Exact recovery scores 1; the context-and-edge-type mean predictor scores 0; negative scores are worse. A spatially constant offset changes activation values but cancels in spatial differences.

% Source: supplementary/sections/reconstruction-evaluation/03-temporal-differences.tex
\subsection{Temporal activation-difference fidelity}
\label{sec:supp-temporal-metric}

Temporal $R^2$ measures recovery of channel-wise differences between consecutive decisions within a drive. Squared norms pool channel errors. The pair set $\mathcal T$ contains 56 pairs across 64 Dataset-Selected validation scenes from 8 drives, or 5,542 Dataset-Full pairs. Decision timestamps and image correspondence are verified across DWMs.
For each observed context $c$ and retained token position $p$, we compute
\begin{equation}
\begin{aligned}
 \vdelta^{\mathrm{temp}}_{ii'cp}&=\vx_{i'cp}-\vx_{icp},\\
 \widehat{\vdelta}^{\mathrm{temp}}_{ii'cp}&=\widehat{\vx}_{i'cp}-\widehat{\vx}_{icp}.
\end{aligned}
\label{eq:supp-temporal-increments}
\end{equation}
For each observed context, we compute a channel-wise target mean across all pairs and positions:
\begin{equation}
 \vmu^{\mathrm{temp}}_c=\frac{1}{|\mathcal T|\,N}
 \sum_{(i,i')\in\mathcal T}\sum_p\vdelta^{\mathrm{temp}}_{ii'cp}.
 \label{eq:supp-temporal-mean}
\end{equation}
The temporal score pools differences over all pairs and token positions, using observed contexts only:
\begin{equation}
 R^2_{\mathrm{temp}}=1-
 \frac{\sum_{(i,i')\in\mathcal T}\sum_{c,p}\|\vdelta^{\mathrm{temp}}_{ii'cp}-\widehat{\vdelta}^{\mathrm{temp}}_{ii'cp}\|_2^2}
 {\sum_{(i,i')\in\mathcal T}\sum_{c,p}\|\vdelta^{\mathrm{temp}}_{ii'cp}-\vmu^{\mathrm{temp}}_c\|_2^2}.
 \label{eq:supp-temporal-r2}
\end{equation}
Adjacent decisions are approximately 0.5\,s apart; differences are not divided by elapsed time.
We use 4 observed contexts for DriveLaW and 1 each for DriveWAM and SimWAM.
Generated DriveWAM contexts and DriveLaW slots without observed-image conditioning are excluded.

\paragraph{Interpreting temporal fidelity.}
Higher temporal $R^2$ means better recovery of the magnitude and sign of temporal changes. Exact recovery scores 1; the observed-context mean-difference predictor scores 0; negative scores are worse. An offset shared by paired observations cancels in their difference.

% Source: supplementary/sections/reconstruction-evaluation/04-pooling-and-support.tex
\subsection{Averaging scores across layers and DWMs}
\label{sec:supp-reconstruction-pooling}

For each metric and dataset, we average the three layer scores within each DWM. The across-DWM mean weights all three DWMs equally, which is equivalent to averaging all nine interfaces with equal weight.

\paragraph{Within-layer calculation.}
Within each layer, we pool squared errors and centred target energies before taking their ratio. Groups with greater target variation receive more weight. Accumulation uses 64-bit floating point; denominators must be finite and positive. Negative scores are not clipped.

\paragraph{Interpreting reconstruction fidelity.}
Spatial edges connect computational positions whose physical receptive fields differ across DWMs.
Temporal differences combine actor motion, ego motion, viewpoint changes and other representation changes (\cref{eq:supp-activation-r2,eq:supp-spatial-r2,eq:supp-temporal-r2}).
Positions are matched without tracking or motion compensation, so paired vectors may refer to different actors or world points.
These reconstruction metrics do not measure semantic localisation accuracy, future-prediction accuracy or driving performance.

\end{recentedits}

\par\endgroup

\FloatBarrier
% Source: supplementary/sections/03-ears-dataset.tex
\FloatBarrier
\section{EARS-Graph dataset and references}
\label{sec:supp-ears-dataset}

EARS-Graph provides image-aligned references for localisation, relations and entity states. We describe the scene sample, graph construction and review.

\begin{recentedits}
\clubpenalty=10000
\widowpenalty=10000
% Source: supplementary/sections/ears-dataset/01-population.tex
\subsection{Source population and experimental role}
The Ego-centric Attributed Road Scene Graph (EARS-Graph) provides image-aligned semantic references for DWM representations.
Inspired by the Combined Road Substrate (CRS), it describes vehicles, road structure and traffic controls through attributes and relations~\citep{wild_bridging_2026}.
Pre-labelling combines nuPlan map geometry with current-image detections; preceding frames help classify vehicle motion~\citep{karnchanachari2024nuplan}.
The dataset contains 300 front-camera frames from OpenScene v1.1 mini, spanning 50 drives in 4 cities~\citep{opendrivelab_openscene}.
Evaluations use the automatically generated graphs.
Human review checked selected revised graphs prepared for release.%
\footnote{Each scene record (version \texttt{selected300-v2}) includes timestamp, recording and drive identifiers, and camera calibration.
It also includes ego pose, annotation provenance and a 5-frame image sequence spanning about 2~s.
Image coordinates and boxes refer to the rectified $1920\times1080$ PNG; the original unrectified JPEG is also retained.}

We selected 300 of 676 candidates using generated annotations and known graph attributes. Selection targeted at least 40 frames per city and 40 predicted control-to-lane links across the 4 cities. We limited each drive to 18 frames, spaced at least 5~s apart.
\par\smallskip
\begingroup
\centering
\small
\setlength{\tabcolsep}{4pt}
\renewcommand{\arraystretch}{1.12}
\begin{tabularx}{0.7\columnwidth}{@{}Yr@{}}
\toprule City & Frames\\\midrule
Singapore & 54\\Boston & 64\\Las Vegas & 110\\Pittsburgh & 72\\\bottomrule
\end{tabularx}
\par\endgroup
\smallskip

Frames from each drive share a split. We preserve dictionary-study assignments and hash the remaining drive identifiers, independently of annotation-model training partitions. The 232 fitting frames from 40 drives select localisation units and thresholds and fit relation and state readouts. Validation contains 66 frames from 9 drives and was used during development. We exclude 2 frames from an original-test drive. Graph statistics cover all 300 frames.

\paragraph{Evaluation targets and image regions.}
The graph vocabulary extends beyond the scored tasks. \cref{sec:supp-interpretability-population} defines the evaluation targets, and \cref{sec:supp-interpretability-localization} explains pixel mapping and localisation references. Missing detections remain unlabelled rather than defining class absence. Regulatory elements have only 1 positive validation frame and drive before native mapping.
Parking attributes are not scored; pedestrian and physical-sign references are unavailable.

% Source: supplementary/sections/ears-dataset/02-schema.tex
\subsection{Graph nodes, attributes, and relations}
\label{sec:supp-crs-schema}
Each graph represents 1 front-camera frame using typed nodes and directed relations. The dataset contains 12,596 nodes and 15,378 edges. Lane nodes carry centrelines and surface polygons; some retain ego-lane context outside the image. The ego node marks the vehicle-frame origin (\cref{tab:supp-crs-nodes}).

\begin{table}[!htbp]
\centering
\begin{minipage}{0.94\textwidth}

 \recenteditcolor
 \recenteditcaption
 \centering
 \small
 \setlength{\tabcolsep}{4pt}
 \renewcommand{\arraystretch}{1.12}
 \caption{\textbf{Node counts in the 300 automatically generated graphs.}
 Vehicle counts refer to detections in individual frames. Cars, buses, and trucks are subclasses of the vehicle node type.}
 \label{tab:supp-crs-nodes}
 \begin{tabularx}{0.99\linewidth}{@{}>{\raggedright\arraybackslash}p{0.34\columnwidth}rY@{}}
 \toprule
 Node type & Count & Labels or role\\
 \midrule
 Lane boundary & 6,993 & Geometric lane boundary\\
 Lane & 3,589 & 3,508 lanes; 81 ego-lane context nodes\\
 Ego & 300 & Known ego origin\\
 Vehicle & 1,413 & 1,222 car; 167 truck; 24 bus\\
 Traffic light & 294 & Detected traffic light\\
 Regulatory element & 7 & Generic regulatory element\\
 \bottomrule
 \end{tabularx}

\end{minipage}
\end{table}

\paragraph{Geometry and evidence.}
Nodes record identity, type, label, granularity, observation field and geometry. Geometry specifies coordinates, points and any camera or role. Detection identifiers are frame-local; linking frames requires additional evidence. Each property records its value, evidence or unknown status. Evidence identifies its source, producer, time and available scores or review details. Producer-specific scores require calibration before they can represent correctness probabilities.
Evidence distinguishes annotations, model predictions and rules. Optional fields include source identifiers, camera, raw score, review status and method-specific details.

Each edge records its source and target nodes, relation, predicate, reference frame, and evidence.
\Cref{tab:supp-crs-relations} lists the 11 relation types with their counts, directions, and meanings.

\begin{table}[!htbp]
\centering
\begin{minipage}{0.94\textwidth}

 \recenteditcolor
 \recenteditcaption
 \centering
 \small
 \setlength{\tabcolsep}{4pt}
 \renewcommand{\arraystretch}{1.12}
 \caption{\textbf{Directed relations in the 300 automatically generated graphs.}
 Arrows point from source to target. Each edge records whether it comes from a map annotation, a geometric rule, or a model prediction.}
 \label{tab:supp-crs-relations}
 \begin{tabularx}{\textwidth}{@{}>{\raggedright\arraybackslash}p{0.19\textwidth}r>{\raggedright\arraybackslash}p{0.21\textwidth}Y@{}}
 \toprule
 Relation & Count & Direction & Meaning\\
 \midrule
 Left boundary & 3,361 & Boundary $\to$ lane & Boundary is on the lane's left in its travel direction.\\
 Right boundary & 3,366 & Boundary $\to$ lane & Boundary is on the lane's right in its travel direction.\\
 Successor & 2,103 & Lane $\to$ lane & Target continues the source lane in its travel direction.\\
 Adjacent & 4,064 & Lane $\to$ lane & The lane boundaries coincide within a spatial tolerance over a minimum length.\\
 In & 234 & Ego $\to$ lane & Supplied full lane surface uniquely contains the ego origin.\\
 Vehicle in lane & 318 & Vehicle $\to$ lane & Lane surface uniquely contains the estimated vehicle position.\\
 Controls & 165 & Control element $\to$ lane & The model predicts that this light or regulatory element applies to the target lane.\\
 Left of ego origin & 785 & Vehicle $\to$ ego & Vehicle lies to the left of the ego origin.\\
 Right of ego origin & 376 & Vehicle $\to$ ego & Vehicle lies to the right of the ego origin.\\
 Ahead of vehicle & 332 & Vehicle $\to$ vehicle & Source has greater forward $X$, beyond a position margin.\\
 Left of vehicle & 274 & Vehicle $\to$ vehicle & Source has greater leftward $Y$, beyond a position margin.\\
 \bottomrule
 \end{tabularx}

\end{minipage}
\end{table}

% Source: supplementary/sections/ears-dataset/03-generation.tex
\subsection{Graph construction and review}
\label{sec:supp-crs-generation}
We combine map geometry, detections and temporal activity estimates, then review nodes and relations. \paragraph{Road geometry and image alignment.}
The image-timestamp ego pose transforms nuPlan geometry into vehicle coordinates. Lane centrelines and boundaries lie on the local $z=0$ plane; slopes can therefore introduce projection offsets. We retain geometry within the map window and calibrated front-camera frustum, including occluded segments. Camera extrinsics and undistortion-adjusted intrinsics project it onto the rectified image. Coordinates use metres: $X$ forward, $Y$ left and $Z$ up. Boundaries are geometric limits and need not be painted markings. Outside-view ego-lane context is excluded from learned relation candidates.

\paragraph{Image-based entity detection.}
RF-DETR Large detects car, bus and truck boxes in the rectified image~\citep{robinson2026rfdetr}.
TopoMLP R50 detects lights and regulatory elements, with categories specifying signal state or regulatory meaning~\citep{wu2024topomlp}.
FCOS3D estimates vehicle positions~\citep{wang2021fcos3d}.
We project its 3D boxes into the image and match class-compatible RF-DETR detections one-to-one by maximum total 2D intersection-over-union (IoU).
Matches must pass overlap, detection-score and position checks.
The matched 3D box's bottom centre estimates ground position in ego coordinates.
Other vehicles retain only their 2D boxes.
Ground-position estimates cover 600 of 1,413 vehicles; detector vocabularies limit coverage.

\paragraph{Temporal modelling of vehicle activity.}
\label{sec:supp-crs-reference-scope}
Vehicle activity uses the current frame and up to 4 preceding frames. A histogram gradient boosting classifier combines point tracks, detection associations and FCOS3D estimates to classify stationary versus moving vehicles. It is trained offline on vehicle-speed references, including some recordings in this cohort. Calibration and ego poses account for ego motion. Parking requires stationarity, consistent multiframe detection associations and repeated FCOS3D parking support. Motion and parking remain separate attributes and are unknown when evidence is insufficient. Car-activity evaluation uses the motion labels.

\paragraph{Relation construction.}

Map associations define boundary-to-lane links; geometric rules define spatial relations.
Adjacent boundaries must overlap within 0.15~m. Overlap must cover at least 2~m and 20\% of the shorter boundary.
We store both directions.
Vehicle membership requires its estimated ground position to fall inside exactly 1 candidate lane.
Separate histogram gradient boosting classifiers predict control and successor relations from OpenLane-V2 training annotations~\citep{wang2023openlanev2}.
Control features compare detected traffic elements with projected lane geometry.
Successor features describe endpoint proximity and direction for ordered lane pairs.
Threshold-passing pairs become edges; unresolved relations remain unknown.
Training, threshold calibration and validation use recording-separated partitions.
Auxiliary validation uses known nodes and geometry, so it does not measure end-to-end OpenScene graph accuracy.

\subsubsection{Review and export checks}
\label{sec:supp-ears-generation-pipeline}
The dataset's graph statistics and all semantic scores are based on the
automatically generated graphs before review corrections. We performed
structural checks on all 300 graphs.

The AI-assisted review screened all 300 frames and inspected 14 at higher
magnification. It first corrected node annotations and then assessed
relations while holding the revised nodes, their geometry and their
attributes fixed. New traffic-control edges required evidence from both
the images and lane directions. Vehicle relations were assessed using
calibrated position estimates. The review added 196 edges across 84 frames.
Transitive reduction then replaced 29 direct edges with paths, yielding
a net increase of 167 direct edges. The archived records retain review
decisions and unresolved pairs.

Before release, we manually checked selected graphs against their source
images. We recorded decisions by frame and used ego-frame positions to
check vehicle relations. Estimating
dataset-wide annotation accuracy would require independent human annotations.

% Source: supplementary/sections/ears-dataset/04-relations.tex
\subsection{Relation vocabulary and direction}
\label{sec:supp-crs-relations}
A lane's direction of travel defines its left and right boundaries. Vehicle positions are compared along the ego vehicle's forward and leftward axes.
Adjacent lanes share a boundary. Their travel directions and any permission to change lanes must be established separately.
An \emph{ahead-of} edge records longitudinal order. Shared lane membership and following behaviour require separate evidence.
Vehicle-order counts use transitive reduction. If A is ahead of B and B ahead of C, the path establishes A ahead of C. A direct edge is unnecessary. Unresolved pairs remain unknown; negatives require an explicit reference rule.

% Source: supplementary/sections/ears-dataset/07-figure-provenance.tex
\supplementaryresultbreak

\subsection{\texorpdfstring{\raggedright}{}An EARS-Graph example for \mbox{semantic evaluation}}
\label{sec:supp-method-reference-details}
\begin{figure}[!htbp]
\centering
\begin{minipage}{0.90\textwidth}

 \centering
 \recenteditcaption
 \includegraphics[width=0.92\columnwidth,height=0.40\textheight,keepaspectratio]{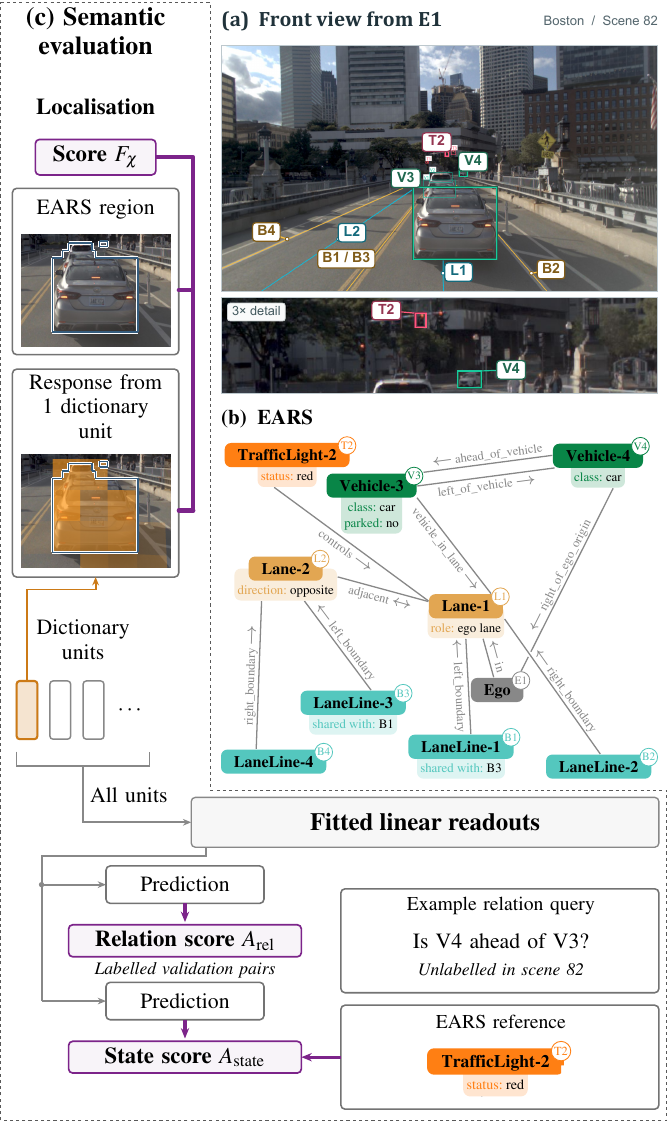}
 \caption{\textbf{Scene~82 and semantic evaluation.} Adapted from Figure~2 of the main paper. The relation query is illustrative and is unlabelled in this scene.}
 \label{fig:supp-ears-evaluation-example}

\end{minipage}
\end{figure}
The figure uses fitting scene 82 from Boston. Panel (a) shows detections; lane boundaries are aligned to visible markings for display. Panel (b) shares entity identifiers with the image. Its arrows illustrate queries: scene 82 labels T2 red but leaves V4's order relative to V3 unknown. Only known orders enter scoring.

Panel (c) connects annotations to evaluation. The V3 outline shows the union of car boxes; orange shows the fitting-selected unit's response strength. This is mixed-rank spatialBSF block 284 at DriveWAM layer 15. Relation and state readouts pool all decoded unit contributions over entity regions and project them to 128 dimensions. Linear readouts predict pairwise relations or individual states; scores measure agreement across eligible validation examples (\cref{fig:supp-ears-evaluation-example}).

\end{recentedits}

\FloatBarrier
% Source: supplementary/sections/04-semantic-evaluation.tex
\FloatBarrier
\section{Semantic evaluation}
\label{sec:supp-interpretability}

EARS evaluates whether individual dictionary units localise scene entities and whether the dictionary’s complete reconstruction preserves information about entity relations and states. In this section, we define the readout fitting, scoring metrics, and combined index.

\begin{recentedits}
% Source: supplementary/sections/semantic-evaluation/01-population.tex
\subsection{Data and tasks}
\label{sec:supp-interpretability-population}
We test whether 1 dictionary unit locates a target and whether collective readouts predict entity relations or states. The task table lists targets, metrics and eligible cohorts. Localisation and conjunction use F1; relation and state readouts use balanced accuracy, the mean recall across both classes. EARS labels fit readouts, not dictionaries. Fitting-drive cross-validation selects units, thresholds and readouts; separate validation drives evaluate them (\cref{tab:supp-semantic-task-cohorts}).

We use the C.1 split: 232 fitting and 66 development-exposed validation frames. The 2 original-test frames remain excluded. New drives are sorted by identity hash; the first quarter, rounded up, enters validation. Existing assignments and the original dictionary split remain unchanged.

\cref{sec:supp-training-protocol}--\cref{sec:supp-sae-baselines} define checkpoint and capacity settings. The 7 fixed-unit methods select 64 units per token; scalar capacity differs between SAE features and BSF blocks. Mixed-rank spatialBSF uses 167--256 coefficient slots here. Mean-activity methods are compared separately in \cref{sec:supp-lasso-average-budget}.

\begin{table}[!htbp]
\centering
\begin{minipage}{0.94\textwidth}

\centering\small
\setlength{\tabcolsep}{4pt}
\renewcommand{\arraystretch}{1.12}
\caption{\textbf{Validation cohorts for the 11 EARS tasks.} Each task uses a common cohort across dictionary methods and interfaces. Localisation uses positive-reference frames; relations and the combined relation query use pairs with known references; states use measurable entities. For relation balanced accuracy (BA), parentheses give the number of drives containing both classes.}
\label{tab:supp-semantic-task-cohorts}
\begin{tabularx}{\textwidth}{@{}>{\raggedright\arraybackslash}p{0.94in}>{\raggedright\arraybackslash}p{2.04in}p{0.40in}Y@{}}
\toprule
Task & Reference or positive label & Metric & Validation support\\
\midrule
Car & Union of car boxes; excludes truck/bus & F1 & 57 frames, 9 drives\\
Large vehicle & Pixelwise union of truck/bus boxes; excludes car & F1 & 36 frames, 7 drives\\
Road & Union of noncontext lane-surface polygons & F1 & 65 frames, 9 drives\\
Traffic light & Union of traffic-light boxes & F1 & 10 frames, 5 drives\\
\midrule
Ahead & Annotated strict order or transitive reachability & BA & 130 pairs, 19 frames, 8 drives (8)\\
Left & Annotated strict order or transitive reachability & BA & 120 pairs, 16 frames, 9 drives (9)\\
Lane membership & Certified unique membership & BA & 582 pairs, 31 frames, 9 drives (9)\\
Shared lane & 2 known memberships in the same lane & BA & 54 pairs, 9 frames, 6 drives (2)\\
Conjunction & Ahead AND shared lane; both references known & F1 & 46 pairs (5 positive, 41 negative), 9 frames, 6 drives; positives in 3 frames, 2 drives\\
\midrule
Light state & Red positive, green negative & BA & 17 lights (5 red, 12 green), 5 drives\\
Car activity & Stationary positive, moving negative & BA & 202 cars (159 stationary, 43 moving), 9 drives\\
\bottomrule
\end{tabularx}
\par\smallskip
\noindent\begin{minipage}{\textwidth}\footnotesize
\noindent\textbf{Fitting support within the 232-frame set.}
\par\smallskip
\noindent\begin{adjustbox}{max width=\linewidth}\begin{tabular}{@{}p{0.21\columnwidth}p{0.55\columnwidth}r@{}}
\toprule Task & Entities or positive frames & Drives\\\midrule
Large vehicle & 104 positive frames & 29\\
Light state & 122 lights: 64 red, 58 green & 22\\
Car activity & 504 cars: 324 stationary, 180 moving & 37\\\bottomrule
\end{tabular}\end{adjustbox}
\par\smallskip
State BA includes all 5 light-validation drives; only 3 contain both states.
\end{minipage}

\end{minipage}
\end{table}

% Source: supplementary/sections/semantic-evaluation/02-localization.tex
\subsection{Localisation with 1 dictionary unit}
\label{sec:supp-interpretability-localization}
Pixel-area F1 measures agreement between the unit's predicted region and the EARS-Graph reference (\cref{fig:supp-pixel-area-f1-example}).

\paragraph{Reference regions.}
To evaluate target localisation, we map the EARS-Graph reference regions onto the model's native spatial grid. Road references follow the annotated road polygons, including occluded portions, while box-based references cover the full annotated box area, including enclosed background. We merge overlapping regions of each target before mapping so each pixel counts once, and combine truck and bus boxes into the large-vehicle category. Calibration maps rectified pixel centres to raw native cells, keeping only valid rays and retained cells. Boxes use half-open pixel-centre bounds without padding. Polygons use 8-bit fractional fixed-point rasterisation, with a 0.1 m near plane for road geometry.

\paragraph{Strength and direction readouts.}
Strength measures contribution magnitude. Direction measures projection onto a fitted axis that separates target regions from background, retaining sign and amplitude. This axis lies in activation space. 
We use reduced QR factorisation $\mD_b=\bar{\mQ}_b\mT_b$.
Here $\bar{\mQ}_b\in\mathbb R^{d\times k_b}$ has orthonormal columns and $\mT_b\in\mathbb R^{k_b\times k_b}$ is upper triangular.
For SAE, $k_b=1$; for BSF, $k_b=r_{\max}$.
For scene $i$ and cell $p$, $\overline{\vz}_{ipb}\in\mathbb R^{1\times k_b}$ is the coefficient row after spatialBSF transport.
Set $\vu_{ipb}=\mT_b\overline{\vz}_{ipb}^{\mathsf T}$, giving decoded contribution $\bar{\mQ}_b\vu_{ipb}=\mD_b\overline{\vz}_{ipb}^{\mathsf T}$.
These coordinates express the contribution in an orthonormal basis.
For readout $\chi\in\{\mathrm{str},\mathrm{dir}\}$,
\begin{equation}
 \begin{aligned}
 \psi^{(\mathrm{str})}_{ipb}&=\|\vu_{ipb}\|_2,\\
 \psi^{(\mathrm{dir})}_{ipb}&=\frac{\vv_b^{\mathsf T}\vu_{ipb}}{\sigma_b^{\mathrm{rd}}}.
 \end{aligned}
\end{equation}
Here $\sigma_b^{\mathrm{rd}}$ is a fitting-only RMS scale and $\vv_b\in\mathbb R^{k_b\times1}$ a fitted unit vector.
SAE strength is the absolute coefficient times the decoder-atom norm.
We select a detector separately for each readout.

\paragraph{Detector selection.}
To accommodate the differing activation modelling assumptions across MI methods, we grant each method the flexibility to select its own best-performing dictionary unit.
\begin{enumerate}
\item Hash fitting-drive identities and assign whole drives round-robin to 6 folds.
\item Fit each candidate direction on 5 folds using L2 logistic regression, $C=1$.
\item Select the threshold maximising equal-drive mean scene F1 from 257 active-score quantiles and $\pm\infty$.
\item Score the remaining fold and repeat across folds.
\item Select the unit with highest out-of-fold mean F1, then refit on all eligible fitting scenes.
\item Fix the unit, direction and threshold before validation.
\end{enumerate}
The strength readout measures magnitude and requires no direction fitting. Searches are separate for each dataset, method, interface and readout (\cref{tab:supp-semantic-readout-fitting}).
We predict foreground when $\|\vu_{ipb}\|_2>10^{-12}$ and $\psi^{(\chi)}_{ipb}>\tau$, where $\tau$ is the fitted threshold.
Otherwise, we predict background.
Direction readouts also require a valid fitted projection.

\begin{figure}[!htbp]
\centering
\begin{minipage}{0.90\textwidth}

\centering
\includegraphics[width=\textwidth,height=0.40\textheight,keepaspectratio]{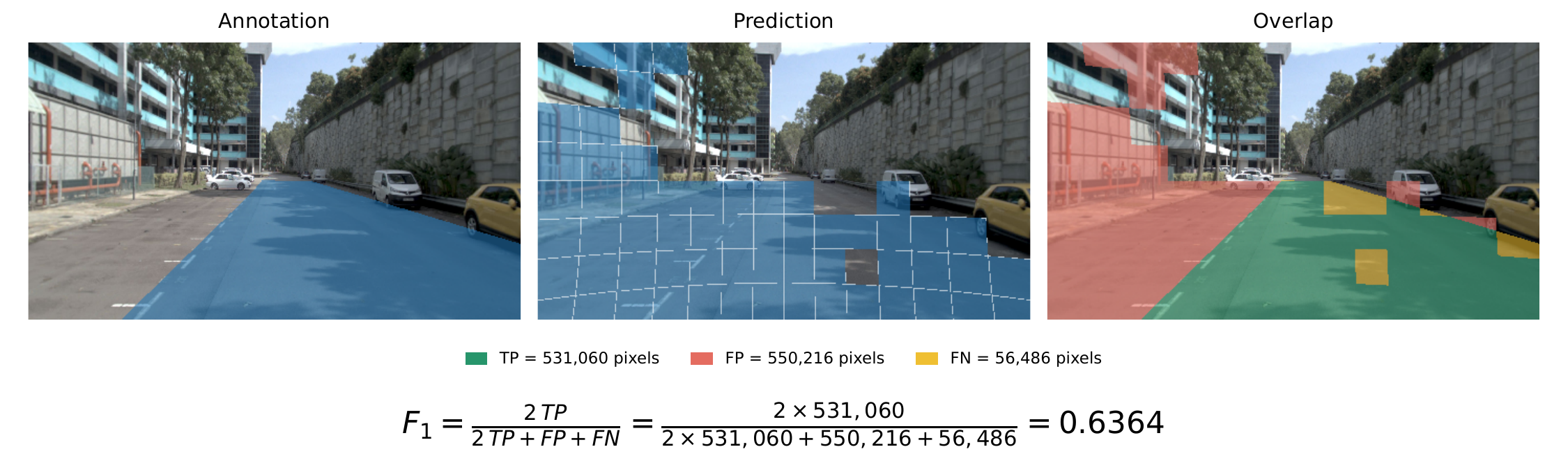}
\caption{\textbf{Pixel-area F1 for 1 road example.} DriveWAM L15, spatialBSF trained on Dataset-Full, block 378, strength readout. Green: true positives; red: false positives; yellow: false negatives.}
\label{fig:supp-pixel-area-f1-example}

\end{minipage}
\end{figure}

\begin{table}[!htbp]
\centering
\begin{minipage}{0.94\textwidth}

\centering\small
\setlength{\tabcolsep}{4pt}
\renewcommand{\arraystretch}{1.15}
\caption{\textbf{Features and prediction rules for EARS readouts.} Localisation evaluates one dictionary unit at a time. Relation and state readouts use entity features pooled from the complete dictionary reconstruction; the combined query joins existing relation predictions without fitting another readout.}
\label{tab:supp-semantic-readout-fitting}
\begin{tabularx}{\textwidth}{@{}>{\raggedright\arraybackslash}p{0.92in}>{\raggedright\arraybackslash}p{1.46in}YY@{}}
\toprule
Task & Input & What we fit or select & Prediction\\
\midrule
Localisation & 1 dictionary unit's response at each cell & Unit and threshold; the direction readout also learns a target-sensitive projection & Foreground or background\\
\addlinespace
Relations & Features from 2 entities & A linear predictor using ridge regression~\citep{hoerl_ridge_1970} & Whether a relation holds\\
\addlinespace
States & Features from 1 entity & A linear predictor using ridge regression & Red/green light; stationary/moving car\\
\addlinespace
Conjunction & Ahead and shared-lane predictions & No additional fitting: combine with AND & Both relations hold\\
\bottomrule
\end{tabularx}

\end{minipage}
\end{table}

\paragraph{Pixel-area F1.}
TP counts measurable pixels shared by prediction and annotation. FP counts predicted foreground outside the annotation; FN counts missed annotated foreground. A predicted cell contributes its entire measurable footprint, so it can contain both TP and FP. \begin{equation}
 \mathrm{F1}_i=\frac{2\mathrm{TP}_i}{2\mathrm{TP}_i+\mathrm{FP}_i+\mathrm{FN}_i}.
\end{equation}
Partial coverage contributes pixel area. We average scene F1 within drives, then equally across drives. Only target-positive scenes enter; absent-target false alarms are not tested. All compared methods use the same eligible scenes for each target and readout.
Diagnostics include all-foreground predictions, static spatial templates, source-local strength, pixel coverage and a label-informed native-cell ceiling.

% Source: supplementary/sections/semantic-evaluation/03-relations-and-composition.tex
\subsection{Entity features across the dictionary}
\label{sec:supp-interpretability-relations}
We use linear readouts~\citep{alain2016understanding} to test whether a dictionary's complete reconstruction supports predictions of entity relations and states. These global readouts combine contributions across dictionary units, complementing the single-unit localisation evaluation in \cref{sec:supp-interpretability-localization}. All methods use the same annotated entity regions and types, pooling features separately within each entity's region.

We construct each entity's features by combining coefficients from all dictionary units through the fixed decoder mapping, including any spatial transport. We average the reconstructed activations within the entity's region and project them to a 128-dimensional feature vector:
\begin{equation}
 \vh_e=\left(\sum_p w_{ep}\widehat{\vx}_p\mP_{m,\ell}\right)^{\mathsf T}.
 \label{eq:supp-entity-feature}
\end{equation}
Here $\widehat{\vx}_p$ combines the units' decoded contributions at cell $p$, including any decoder bias.
The weight $w_{ep}$ is entity $e$'s measurable pixel fraction in cell $p$; weights sum to 1.
The label-free Gaussian projection $\mP_{m,\ell}$ is fixed per DWM--layer interface and shared across methods.
Regions with no measurable pixels are excluded; partial regions remain.
Reference geometry and coverage determine the common entities and pairs before fitting.

Relation readouts combine 2 entities; state readouts use 1 (\cref{tab:supp-semantic-entity-features}).
Ridge penalties are selected through 6-fold fitting-drive cross-validation~\citep{hoerl_ridge_1970}.
Target labels are used to fit and score the readouts; dictionary training uses reconstruction.

\begin{table}[!htbp]
\centering
\begin{minipage}{0.94\textwidth}

\centering\small
\setlength{\tabcolsep}{3pt}
\renewcommand{\arraystretch}{1.15}
\caption{\textbf{Inputs to relation and state readouts.} The vectors $\va$ and $\vb$ are entity features pooled from reconstructed activations. Brackets concatenate entries; ``Yes'' denotes a fitted intercept.}
\label{tab:supp-semantic-entity-features}
\begin{tabularx}{0.99\linewidth}{@{}>{\raggedright\arraybackslash}p{0.74in}Yrc@{}}
\toprule
Task & Feature & Dim. & Intercept\\
\midrule
Ahead/left & $\va-\vb$ & 128 & No\\
Membership & $[\va;\vb;\va^{\mathsf T}\vb;$\newline$\|\va-\vb\|_2^2]$ & 258 & Yes\\
Shared lane & $[\va+\vb;\va^{\mathsf T}\vb;$\newline$\|\va-\vb\|_2^2]$ & 130 & Yes\\
State & $\vh_e$ & 128 & Yes\\
\bottomrule
\end{tabularx}

\end{minipage}
\end{table}

% Source: supplementary/sections/semantic-evaluation/04-states.tex
\subsection{Relations and entity states}
\label{sec:supp-interpretability-states}
Using the entity features and fitting procedure in \cref{sec:supp-interpretability-relations}, we evaluate 4 relation tasks---ahead, left, lane membership, and shared lane---and 2 state tasks: red versus green traffic lights and stationary versus moving cars. Directional relations (ahead and left) use $\va-\vb$ with a zero intercept, so reversing a pair reverses its score; their reference orders include chains of relations, while known reverse orders supply negative examples. By contrast, lane membership and shared lane require known, unique memberships, leaving missing relations unknown and excluded from scoring.

For relation balanced accuracy, we average the 2 class recalls within each drive, then average equally across drives containing both classes (shared-lane scoring uses 2 such drives). For state balanced accuracy, each drive receives equal weight, shared equally among its eligible entities; we then compute class recalls over all weighted entities and average both recalls, including drives containing only 1 class.

Following CRS's use of combined relation queries~\citep{wild_bridging_2026}, we also test ``A is ahead of B and both share a lane'' by applying a logical AND to existing predictions for pairs where both references are known. Each drive receives equal weight, divided equally among eligible pairs, and we pool weighted TP, FP, and FN across drives before calculating conjunction F1 (including drives without positive pairs). This conjunction reuses the existing readouts without additional fitting, as withholding this combination left too few fitting examples to test unseen-combination generalisation.

% Source: supplementary/sections/semantic-evaluation/05-aggregation.tex
\subsection{Index aggregation and interpretation}
\label{sec:supp-interpretability-aggregation}
\paragraph{Equal task and interface weights.}
For each interface and readout $\chi$, the EARS semantic agreement index is
\begin{equation}
 I_{\mathrm{EARS},\chi}=\frac{4F_\chi+5A_{\mathrm{rel}}+2A_{\mathrm{state}}}{11}.
 \label{eq:supp-ears-index}
\end{equation}
$F_\chi$ averages 4 localisation F1 scores. $A_{\mathrm{rel}}$ averages 4 relation balanced accuracies and conjunction F1. $A_{\mathrm{state}}$ averages 2 state balanced accuracies. Only localisation differs between strength and direction indices. We average 3 interfaces within each DWM and all 9 overall. Tables abbreviate $I_{\mathrm{EARS},\chi}$ to $I_\chi$.

\paragraph{What semantic agreement establishes.}
EARS measures agreement with selected annotations and readouts, where relation and state readouts receive full entity regions so that both geometry and surrounding context can inform predictions (though small traffic lights may still share grid cells with background or adjacent lights). The index combines task-specific cohorts and scoring rules, and conjunction queries reuse two relation predictions. Validation feedback informed choices such as the truck/bus grouping, so small ranking differences require confirmation with repeated training and new scenes. Establishing behavioural relevance therefore requires targeted interventions, whose effects we examine in DWMs in \cref{sec:supp-application-demo}~\citep{swann_multi-dimensional_2026}.

\end{recentedits}

\FloatBarrier
% Source: supplementary/sections/05-spatialbsf.tex
\FloatBarrier
\section{spatialBSF computational details}
\label{sec:supp-method-details}

spatialBSF encodes, selects and spatially decodes blocks, then corrects and reselects them once. We describe these steps and their training objective.
Fixed-rank spatialBSF and mixed-rank spatialBSF reconstruct spatial activation fields using a fixed number of selected blocks at each source position. The original mixed-rank design permits shorter prefixes; the fixed-rank variant keeps every selected block at full rank.
\begin{recentedits}
% Source: supplementary/sections/spatialbsf/01-encoder-and-frames.tex
\subsection{Activation fields and block encoding}
\label{sec:supp-method-encoder}
A complete $H\times W$ token grid contains $N=HW$ positions; a retained subset keeps its original coordinates.
Let $d$ be the interface's channel count and $\mX\in\mathbb R^{N\times d}$ its normalised activation matrix.
Its row at position $p$ is $\vx_p\in\mathbb R^{1\times d}$.
We use maximum rank $r_{\max}=4$, with $d\geq r_{\max}$, and $B$ dictionary blocks.
$\mA_b,\mZ_b\in\mathbb R^{N\times r_{\max}}$ hold proposed and selected coefficients.
Their rows are $\va_{pb}=(\mA_b)_{p,:}$ and $\vz_{pb}=(\mZ_b)_{p,:}$, with entries $a_{pb,j}$ and $z_{pb,j}$.
Collections $A=\{\mA_b\}_{b=1}^{B}$ and $Z=\{\mZ_b\}_{b=1}^{B}$ follow the guide's notation.

\ifdefined\includesupplementary
The affine encoder in Fig.~\ref{fig:spatialbsf-architecture}(a) of the main paper gives the initial coefficient proposal
\else
The affine encoder in Fig.~3(a) of the main paper gives the initial coefficient proposal
\fi
\begin{equation}
 \widetilde{\va}_{pb}=(\vx_p\mW_b+\vbeta_b)\mO_b,
 \label{eq:supp-affine-encoder}
\end{equation}
The learned encoder parameters are $\mW_b\in\mathbb R^{d\times r_{\max}}$ and $\vbeta_b\in\mathbb R^{1\times r_{\max}}$.
They are shared across positions and untied from the decoder.
Encoding is position-wise; spatial interactions enter through decoding and correction.
The unrotated affine proposal is $\vc_{pb}=\vx_p\mW_b+\vbeta_b$.

\paragraph{Why add a rotation?}
Rotation determines which information shorter rank prefixes retain.
Vanilla BSF uses the affine proposal $\vc_{pb}$ without a separate rotation~\citep{felStructuringSparsityBlockSparse2026}.
Many orthonormal bases span the same 4-dimensional subspace.
Keeping only the first 1 or 2 directions makes their orientation matter.
We learn $\mO_b$ to make these prefixes useful; orthogonality preserves lengths, angles and the full span.
We obtain $\mQ_b$ by reduced QR of an independent learned $d\times r_{\max}$ decoder matrix.
Column signs follow the triangular factor's diagonal.
Its columns are orthonormal: $\mQ_b^{\mathsf T}\mQ_b=\mI_{r_{\max}}$.
We parameterise the rotation as
\begin{equation}
 \mO_b=\mO_{b,0}\exp(\mOmega_b-\mOmega_b^{\mathsf T}),
 \qquad \mO_b^{\mathsf T}\mO_b=\mI_{r_{\max}},
 \label{eq:supp-block-rotation}
\end{equation}
where $\mOmega_b\in\mathbb R^{r_{\max}\times r_{\max}}$ is unconstrained and trainable, and $\mO_{b,0}$ is a fixed orthogonal initialisation.
The matrix exponential preserves orthogonality during training~\citep{lezcano_casado_orthogonal_2019}.
The block dictionary is
\begin{equation}
 \mD_b=\mQ_b\mO_b,
 \qquad \mD_b^{\mathsf T}\mD_b=\mI_{r_{\max}}.
 \label{eq:supp-block-dictionary}
\end{equation}
This preserves the within-block orthonormal geometry of Grassmannian BSF~\citep{felStructuringSparsityBlockSparse2026}.
A rank-$r$ prefix uses the first $r$ dictionary columns and their coefficients.

\paragraph{The rotation preserves the full-rank output.}
With the full set of directions, the encoder and decoder share $\mO_b$, and orthogonality gives
\[
 \begin{aligned}
 (\vc_{pb}\mO_b)(\mQ_b\mO_b)^{\mathsf T}
 &=\vc_{pb}\mO_b\mO_b^{\mathsf T}\mQ_b^{\mathsf T}\\
 &=\vc_{pb}\mQ_b^{\mathsf T}.
 \end{aligned}
\]
The decoded output is unchanged.
Full-block energy is also unchanged: $\|\vc_{pb}\mO_b\|_2^2=\|\vc_{pb}\|_2^2$.
The energy-based gate therefore selects the same blocks.

% Source: supplementary/sections/spatialbsf/02-block-and-rank-selection.tex
\subsection{Calibrated energy scores and selection}
\label{sec:supp-method-allocation}
We extend BSF's block-norm selection~\citep{felStructuringSparsityBlockSparse2026} with a calibrated energy score.
We multiply the normalised prefix energy by a learned scale and add a learned offset.

\paragraph{Calibrated scores.}
A prefix's energy is the sum of its squared coefficients.
We normalise it by the mean full-block energy at the same position:
\begin{align}
 e_{pb}(r)&=\sum_{j=1}^{r}a_{pb,j}^2,\notag\\
 \bar e_p&=\max\!\left(\frac{1}{B}\sum_{b=1}^{B}e_{pb}(r_{\max}),10^{-12}\right).
 \label{eq:supp-block-energy}
\end{align}
Because the dictionary columns are orthonormal, $e_{pb}(r)$ measures the squared norm of the prefix's contribution before spatial transport.
The calibrated score is
\begin{align}
 s_{pb}(r)&=\gamma_{br}\frac{e_{pb}(r)}{\bar e_p}+v_{br},\notag\\
 \gamma_{br}&=\exp(\tanh u_{br}),\qquad e^{-1}<\gamma_{br}<e.
 \label{eq:supp-prefix-score}
\end{align}
Each block/rank pair learns $u_{br}$ and $v_{br}$, both initialised to 0. Scores therefore initially equal normalised energies. The exponential keeps scales positive; tanh limits amplification. Raw energy cannot decrease with rank, so calibration lets shorter prefixes compete.

\paragraph{Using the scores.}
\ifdefined\includesupplementary
The gate in Fig.~\ref{fig:spatialbsf-architecture}(b) considers the ranks
\else
The gate in Fig.~3(b) considers the ranks
\fi
\begin{equation}
 \mathcal R_{\mathrm{fixed}}=\{0,4\},\qquad
 \mathcal R_{\mathrm{mixed}}=\{0,1,2,4\},
 \label{eq:supp-rank-options}
\end{equation}
Rank 0 means unselected, with score 0. Powers of 2 keep the rank grid small; rank 3 is omitted. spatialBSF scores rank-4 blocks; mixed-rank spatialBSF chooses each block's highest-scoring positive rank. The gate retains exactly $K=64$ blocks, even when some scores are nonpositive. Ties favour smaller positive rank, then lower block index. This maximises selection score; overlapping contributions mean it need not minimise reconstruction error.

\paragraph{Selected coefficients and budget.}
Let $r_{pb}$ be the chosen rank for a selected block and 0 otherwise.
The gate keeps the first $r_{pb}$ coefficients in the learned basis order:
\begin{equation}
 z_{pb,j}=\begin{cases}
 a_{pb,j},&j\leq r_{pb},\\
 0,&j>r_{pb},
 \end{cases}
 \qquad j=1,\ldots,r_{\max},
 \label{eq:supp-selected-coefficients}
\end{equation}
Coordinates are not sorted by magnitude, and selected coefficients may be 0.
We write this operation as $Z=\operatorname{Gate}_K(A)$ and apply it at both stages.

Each selected block counts as 1 dictionary unit. spatialBSF allocates $4K=256$ coefficient slots; mixed rank allocates $\sum_b r_{pb}$, between $K$ and $4K$. A scalar SAE allocates 1 slot per selected unit.

% Source: supplementary/sections/spatialbsf/03-spatial-decoding.tex
\subsection{Spatial decoding}
\label{sec:supp-method-spatial}
\label{sec:supp-method-contribution}
All blocks share a fixed graph connecting positions within 1 spatial field.
DriveWAM and SimWAM connect horizontal and vertical neighbours on the native token grid, without wrapping at image boundaries.
DriveLaW uses a symmetrised 4-nearest-neighbour graph on 256 retained positions, with 548 undirected edges.

Each block learns a local-and-neighbour filter on the fixed graph.
Let $\mL_{\mathrm{sym}}$ be its symmetric normalised Laplacian and $\mS=\mL_{\mathrm{sym}}-\mI_N$.
\begin{equation}
 \mH_b=\theta_{b0}\mI_N+\theta_{b1}\mS,
 \qquad |\theta_{b0}|+|\theta_{b1}|\leq1.
 \label{eq:supp-spatial-filter}
\end{equation}
Non-isolated Laplacian diagonals are 1; isolated diagonals are 0.
Edge entries are $-1/\sqrt{\deg(p)\deg(q)}$; other entries are 0.
Thus $S_{pp}=-1$ at isolated positions.
For learned raw gains $\vartheta_{b0},\vartheta_{b1}$, we set $\theta_{bk}=\vartheta_{bk}/\max(1,|\vartheta_{b0}|+|\vartheta_{b1}|)$, $k\in\{0,1\}$.
Gains start at $(\theta_{b0},\theta_{b1})=(1,0)$ and are renormalised after each update.
Positive $\theta_{b1}$ gives negative neighbour weights because $\mS$ has negative edge entries.

The filter $\mH_b\in\mathbb R^{N\times N}$ mixes positions, and $\mD_b\in\mathbb R^{d\times r_{\max}}$ maps coefficients to channels.
\ifdefined\includesupplementary
For fixed parameters, the linear decoder $\mathcal A$ sums these block contributions (Fig.~\ref{fig:spatialbsf-architecture}(c)):
\else
For fixed parameters, the linear decoder $\mathcal A$ sums these block contributions (Fig.~3(c)):
\fi
\begin{equation}
 \widetilde{\mX}=\mathcal A(\widetilde Z)
 =\sum_{b=1}^{B}\mH_b\widetilde{\mZ}_b\mD_b^{\mathsf T}.
 \label{eq:supp-spatial-decoder}
\end{equation}
Selected source blocks can contribute at neighbouring destinations through the same channel dictionary. Reconstruction loss learns these contributions jointly with the dictionary. The K-block budget applies at sources; destinations have no second gate, so transported contributions can overlap. Block $b$ has destination coefficients $\overline{\mZ}_b=\mH_b\mZ_b$ and contribution $\overline{\mZ}_b\mD_b^{\mathsf T}$.
A joint orthogonal coordinate change preserves each decoded contribution but can alter rank-1 and rank-2 prefixes. We therefore interpret decoded contributions rather than individual coordinates.

% Source: supplementary/sections/spatialbsf/04-residual-refinement.tex
\subsection{Residual correction and reselection}
\label{sec:supp-method-refinement}
\ifdefined\includesupplementary
Inspired by learned sparse inference~\citep{gregor_learning_2010}, we use the residual $\mE=\mX-\widetilde{\mX}$ to correct the initial coefficients (Fig.~\ref{fig:spatialbsf-architecture}(d)).
\else
Inspired by learned sparse inference~\citep{gregor_learning_2010}, we use the residual $\mE=\mX-\widetilde{\mX}$ to correct the initial coefficients (Fig.~3(d)).
\fi
The decoder adjoint maps this error back to coefficient space:
\begin{equation}
 \mG_b=[\mathcal A^*(\mE)]_b
 =\mH_b^{\mathsf T}\mE\mD_b.
 \label{eq:supp-decoder-adjoint}
\end{equation}
Each $\mG_b\in\mathbb R^{N\times r_{\max}}$ is the negative coefficient gradient of $\tfrac12\|\mE\|_F^2$.
A learned positive scale $\kappa_b=\exp(\tanh\rho_b)$ weights each block's correction, where $\rho_b$ is an unconstrained scalar.
Let $\mV_b=\kappa_b\mG_b$ and $V=\{\mV_b\}_{b=1}^{B}$.
The step size is
\begin{equation}
 \alpha=\frac{2\operatorname{sigmoid}(\eta)\sum_b\kappa_b\|\mG_b\|_F^2}
 {\max(\|\mathcal A(V)\|_F^2,10^{-12})},
 \label{eq:supp-refinement-step}
\end{equation}
where $\operatorname{sigmoid}(t)=(1+\exp(-t))^{-1}$ and $\eta$ is learned.
Sums and norms cover the complete field.
Above the denominator floor, the ratio without $2\operatorname{sigmoid}(\eta)$ is the least-squares step along $V$.
Minimising $\frac12\|\mE-t\mathcal A(V)\|_F^2$ over $t$ gives
\[
 t^*=\frac{\langle\mE,\mathcal A(V)\rangle_F}{\|\mathcal A(V)\|_F^2}
 =\frac{\sum_b\kappa_b\|\mG_b\|_F^2}{\|\mathcal A(V)\|_F^2}.
\]
The second equality follows from the decoder adjoint and $\mV_b=\kappa_b\mG_b$.
The learned multiplier scales this step within $(0,2)$.
We update the coefficients, reapply the gate and decode:
\begin{align}
 \mA_b&=\widetilde{\mZ}_b+\alpha\kappa_b\mG_b,\notag\\
 Z&=\operatorname{Gate}_K(A),\qquad
 \widehat{\mX}=\mathcal A(Z).
 \label{eq:supp-residual-refinement}
\end{align}
Both stages share scoring and decoder parameters.
Reselection recomputes scores from the corrected proposals, allowing new blocks to enter and mixed-rank choices to change.
This discrete step can increase reconstruction error, despite the preceding residual correction.

% Source: supplementary/sections/spatialbsf/05-fitting.tex
\subsection{Training and evaluation}
\label{sec:supp-method-training}
We train all parameters jointly through both stages, including the residual correction, to minimise the final reconstruction loss
\begin{equation}
 \mathcal L_{\mathrm{rec}}
 =\frac{\|\mX-\widehat{\mX}\|_F^2}{Nd}.
 \label{eq:supp-reconstruction-loss}
\end{equation}
For discrete selection, we compare assignments before and after perturbing their scores to estimate a surrogate gradient~\citep{vlastelica_differentiation_2020}.
$\operatorname{Alloc}_K(\vs)$ gives one-hot rank assignments at 1 position.
Let $\vh$ be the loss gradient with respect to these assignments.
The score-gradient estimate is
\begin{align}
 \widehat{\nabla}_{\vs}\mathcal L
 &=\frac{\operatorname{Alloc}_K(\vs)-\operatorname{Alloc}_K(\vs-\lambda\vh)}{\lambda},\notag\\
 \lambda&=\frac{\lambda_0}{\operatorname{RMS}(\vh)+10^{-12}},
 \label{eq:supp-selection-surrogate}
\end{align}
Here $\lambda_0>0$ controls the perturbation size, and RMS averages over blocks and rank options.
The selection calls using the original and perturbed scores each retain exactly $K$ blocks by comparing positive-rank scores against each block's rank-0 score. During inference, hard selection is applied in both the initial reconstruction stage and the residual-correction stage, with no further coefficient optimisation.

\end{recentedits}

\FloatBarrier
% Source: supplementary/sections/06-training.tex
\FloatBarrier
\section{Training protocols and baseline implementations}
\label{sec:supp-training-details}

We compare 7 MI tools using shared data and dictionary-unit budgets. We describe their training, baseline implementations and the capacity trade-off.

% Source: supplementary/sections/training/01-comparison-protocol.tex
\subsection{Training protocols and comparison controls}
\label{sec:supp-training-protocol}
Within each dataset, all seven MI tools share the fitting corpus at each interface. Under the Dataset-Selected training protocol, all seven tools also share scene batches, sample order, and the learning-rate schedule. Under the Dataset-Full protocol, training budgets are matched by token count, while spatialBSF processes complete spatial fields; consequently, spatialBSF's minibatch order and realised update counts can differ from those of token-wise methods. The accompanying tables specify the dataset populations and optimisation settings (\cref{tab:supp-training-protocol,tab:supp-training-hyperparameters}).

\begin{table}[!htbp]
\centering
\begin{minipage}{0.94\textwidth}

\centering\small\setlength{\tabcolsep}{5pt}
\caption{\textbf{Dictionary training setup.} Validation counts refer to reconstruction evaluation.}
\label{tab:supp-training-protocol}
\begin{adjustbox}{max width=\linewidth}\begin{tabular}{@{}>{\raggedright\arraybackslash}p{0.28\textwidth}>{\raggedright\arraybackslash}p{0.27\textwidth}>{\raggedright\arraybackslash}p{0.38\textwidth}@{}}
\toprule
Setting & Dataset-Selected & Dataset-Full\\
\midrule
Fitting scenes & 800 & 68,538\\
Validation scenes & 100 & 6,653\\
GPU & NVIDIA GB10 & NVIDIA B200\\
Seed & 42 & 42\\
\bottomrule
\end{tabular}\end{adjustbox}

\end{minipage}
\end{table}

\begin{table}[!htbp]
\centering
\begin{minipage}{0.94\textwidth}

\centering\small\setlength{\tabcolsep}{5pt}
\caption{\textbf{Dictionary training hyperparameters.} LR denotes learning rate.}
\label{tab:supp-training-hyperparameters}
\begin{adjustbox}{max width=\linewidth}\begin{tabular}{@{}>{\raggedright\arraybackslash}p{0.28\textwidth}>{\raggedright\arraybackslash}p{0.27\textwidth}>{\raggedright\arraybackslash}p{0.38\textwidth}@{}}
\toprule
Hyperparameter & Dataset-Selected & Dataset-Full\\
\midrule
Optimiser & Adam~\citep{kingma2015adam} & Adam\\
Adam $(\beta_1,\beta_2)$ & $(0.9,0.999)$ & $(0.9,0.999)$\\
Adam $\epsilon$ & $10^{-8}$ & $10^{-8}$\\
Weight decay & 0 & 0\\
Selected dictionary units/token & 64 & 64\\
Batch budget & 2 scenes/update & 65,536 tokens/update\\
Peak LR & $3\times10^{-4}$ & $4\times10^{-4}$\\
Minimum LR & $3\times10^{-5}$ & $10^{-5}$\\
Warmup (updates) & 100 & $\min(1000,\lfloor U_{30}/10\rfloor)$\\
LR schedule & Cosine decay & Plateau: $\times0.5$, patience 2\\
Training limit & 2,400 updates & 30 epochs; plateau stopping\\
\bottomrule
\end{tabular}\end{adjustbox}

\end{minipage}
\end{table}

\paragraph{Dataset-Full checkpoint selection.}
$U_{30}$ denotes the 30-epoch update budget. Warmup lasts 1,000 updates for DriveLaW, 705 for DriveWAM and 792 for SimWAM. Token-wise methods shuffle tokens; spatialBSF accumulates complete fields with a 64-field microbatch limit. After each epoch, checkpoint selection minimises scene-averaged validation NMSE.

The token-wise controller requires mean nonzero activity within 10\% of 64. Its learning-rate score adds the absolute relative activity deviation to reconstruction error. All saved SAE checks satisfy this condition, and each selected epoch minimises reconstruction NMSE. spatialBSF's exact-64 controller uses reconstruction NMSE directly.
Early stopping requires at least 5 epochs, 6 checks without $10^{-4}$ relative improvement, and 2 learning-rate reductions. Reconstruction must also improve over initialisation and have NMSE below 1. All 27 SAE and 18 Vanilla/Grassmannian fits reach 30 epochs. Of 18 spatialBSF fits, 5 stop at 14--26 epochs. Selected epochs span 8--30 for BSFs and 27--30 for SAEs.

\cref{sec:supp-reconstruction-evaluation} and \cref{sec:supp-interpretability} define the shared reconstruction and semantic evaluation. Methods retain their native objectives and parameter constraints, detailed in E and F.2. \cref{sec:supp-dictionary-size} compares dictionary size, block rank and selection budget.
Matryoshka uses the clipping rule in F.2. Other methods check finite gradients without clipping and retain native parameter projections or decoder normalisation.

% Source: supplementary/sections/training/02-sae-baselines.tex
\subsection{SAE and BSF baselines}
\label{sec:supp-sae-baselines}
We compare 3 SAE and 4 BSF variants across 9 interfaces on both datasets. SAEs provide 4,096 scalar features ($1024 \times 4$), whereas BSFs provide 1,024 rank-4 subspace blocks, matching total dictionary capacity. Each method selects 64 units per token: scalar features for SAEs and whole blocks for BSFs. This matches the number of selected units, but not the number of scalar coefficients. Further comparisons of dictionary size, block rank and selection budget are presented in \cref{sec:supp-dictionary-size}, which explains our choice of 1,024 rank-4 blocks and 64 block selections per token.
\paragraph{Modified TopK SAE.}
We modify TopK SAE to use signed coefficients, allowing either direction along each decoder vector~\citep{gao_scaling_nodate}.
An untied affine encoder selects the 64 coefficients of largest absolute magnitude, preserving their signs.
Decoder vectors have unit norm, and training minimises reconstruction MSE.

\paragraph{Matryoshka TopK SAE.}
Matryoshka~\citep{bussmann2025matryoshka} trains nested dictionaries with cumulative widths 128, 384, 896, 1,920 and 4,096.
Our per-token variant selects 64 positions after ReLU; selected values can be 0.
Smaller dictionaries mask this full-width code without reselection, and reported results use the final 4,096-feature dictionary.
We retain the reference initialisation, nested reconstruction losses, loss scaling and auxiliary loss~\citep{marksDictionaryLearningRepository}.
Decoder gradients are projected, global gradient norms clipped at 1.0, and decoder vectors renormalised to unit norm.

\paragraph{Relaxed archetypal SAE.}
RA-SAE uses affine--ReLU--TopK encoding with identity normalisation and 64 selected positions, including possible zeros~\citep{fel2025archetypal}.
Its decoder forms convex combinations of 4,096 fitting-token anchors sampled uniformly without replacement.
It adds relaxation vectors with row norm at most $\delta=1$ and a learned positive multiplier, without further renormalisation.
Training minimises reconstruction MSE~\citep{felOvercompleteRepository}.
Some fitting probes still improve near update 2,400, so this endpoint reflects the fixed training budget.

\paragraph{Vanilla and Grassmannian BSFs.}
Both baselines select the 64 rank-4 blocks with largest coefficient norms and minimise reconstruction MSE~\citep{felStructuringSparsityBlockSparse2026}.
Vanilla BSF uses an untied affine encoder and unit-norm decoder vectors.
Grassmannian BSF uses orthonormal block bases and a tied encoder.
Its coefficients project onto each basis, scaled by a learned block gain.
Both decode each token independently.

\paragraph{spatialBSF and mixed-rank spatialBSF.}
spatialBSF uses the encoder, spatial decoder and residual correction defined in \cref{sec:supp-method-details}. The fixed-rank variant keeps 4 coefficients per block; mixed rank selects prefixes of 1, 2 or 4. Both stages select 64 source blocks and may change block identities during correction. \cref{sec:supp-lasso-average-budget} compares Lasso-BSF, mean-activity spatialBSF and Matryoshka BatchTopK, which target an average activity of 64 units while allowing activity to vary between tokens.

% Source: supplementary/sections/training/03-dictionary-size.tex

\subsection{Dictionary capacity and comparison budgets}
\label{sec:supp-dictionary-size}

\paragraph{Comparison methods.}
We compare how dictionary size, selection budget and block rank affect spatialBSF reconstruction, then assess the coding cost of different ranks. We train fixed-rank dictionaries on Dataset-Selected activations from DriveLaW L18, DriveWAM L15 and SimWAM L15, using seed 42 and 2,400 updates per configuration. This middle-layer focus is motivated by findings that intermediate language-model representations can outperform final-layer representations on downstream tasks~\citep{skean2025layer}.

\begin{figure}[!htbp]
\centering
\begin{minipage}{0.90\textwidth}

\centering
\includegraphics[width=\textwidth,height=0.40\textheight,keepaspectratio]{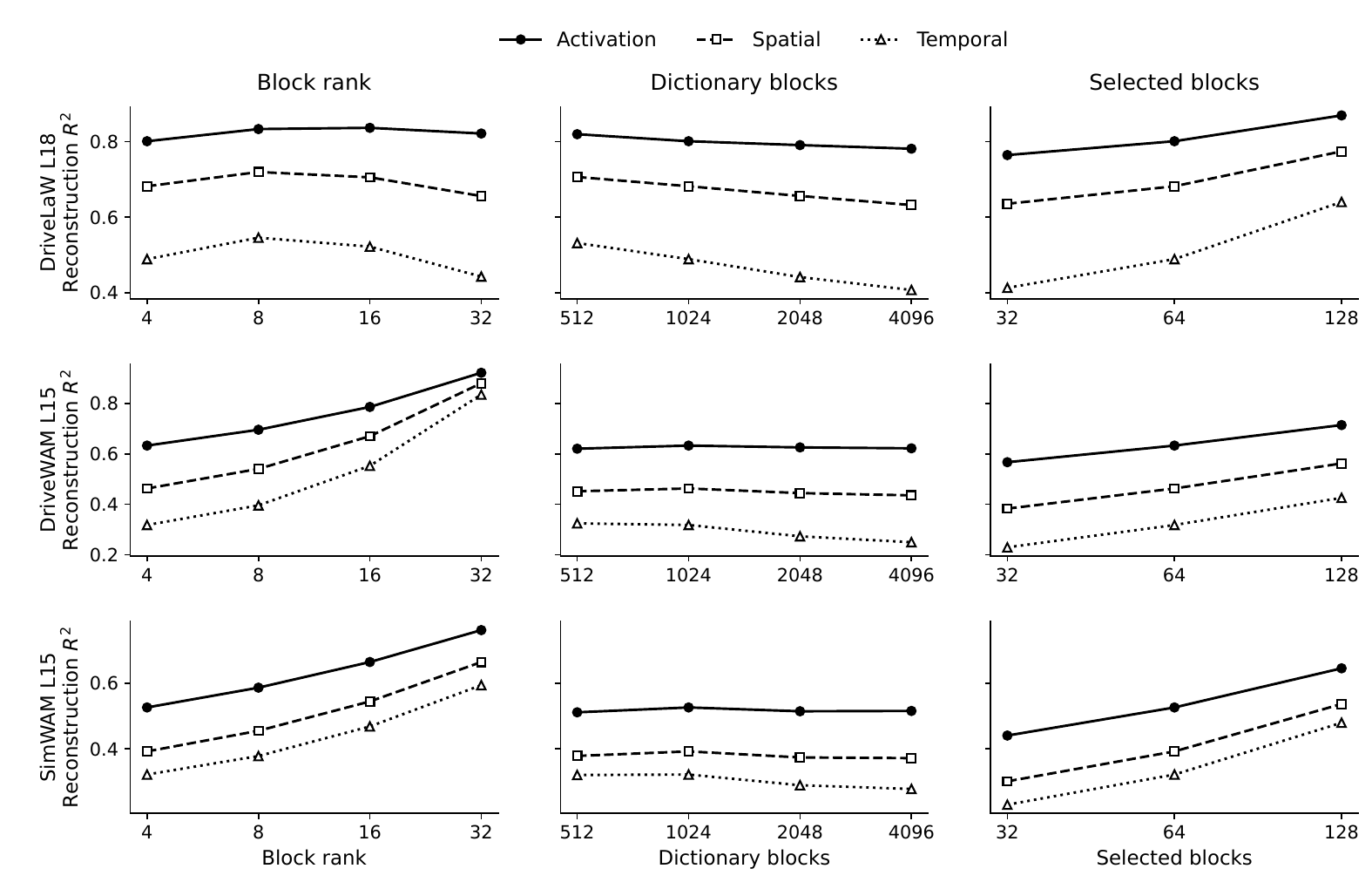}
\caption{\textbf{Reconstruction sensitivity at 3 middle interfaces.} Columns vary rank, dictionary size or selected blocks around $(r,B,K)=(4,1{,}024,64)$. Rows show DWMs. Circles, squares and triangles denote activation, spatial and temporal $R^2$; higher is better. Shared references give 27 trained conditions. Selected scalar slots equal $Kr$; nonzero activity and transported contributions can differ.}
\label{fig:supp-capacity-sweeps}

\end{minipage}
\end{figure}

\begin{figure}[!htbp]
\centering
\begin{minipage}{0.90\textwidth}

\centering
\includegraphics[width=\textwidth,height=0.40\textheight,keepaspectratio]{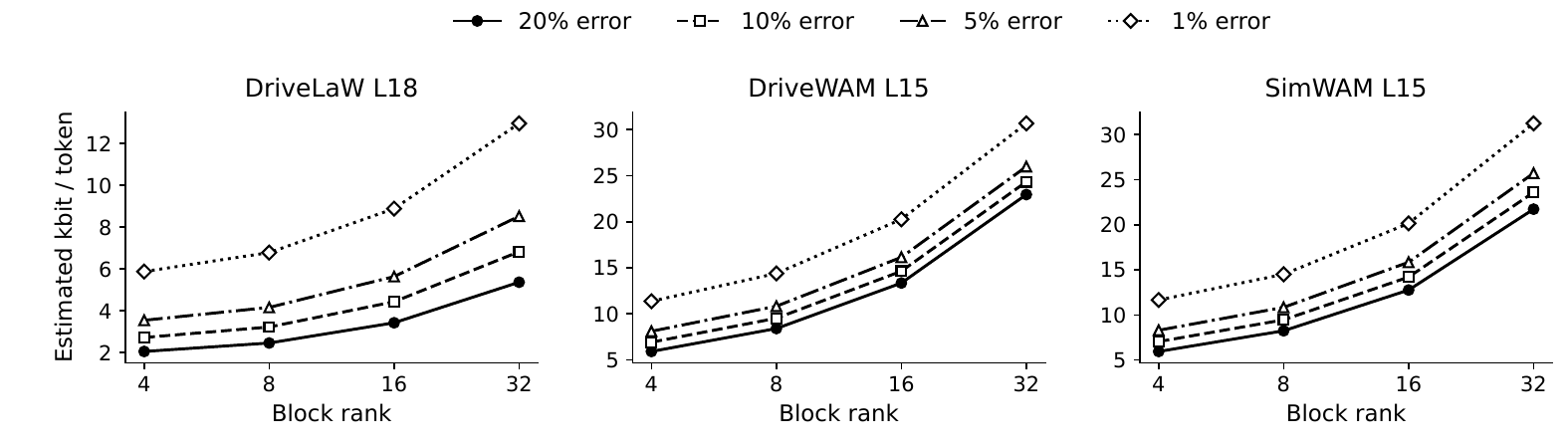}
\caption{\textbf{Estimated description length by rank.} Panels show middle interfaces with 1,024 blocks and 64 selections. Line styles and markers distinguish 4 coding-error fractions relative to input variance. Rates include decoder and coding-model storage at 32 bits per parameter; lower is better. The error budget covers coefficient and residual coding.}
\label{fig:supp-rank-mdl}

\end{minipage}
\end{figure}

Starting from a dictionary of $B=1{,}024$ blocks, block rank $r=4$ and $K=64$ selections per token, we vary one setting at a time:
\begin{itemize}
\item \textbf{Dictionary size:} 512, 1,024, 2,048 or 4,096 blocks, keeping rank 4 and 64 selections.
\item \textbf{Selection budget:} 32, 64 or 128 blocks per token, keeping 1,024 rank-4 blocks.
\item \textbf{Block rank:} 4, 8, 16 or 32, keeping 1,024 blocks and 64 selections.
\end{itemize}
Each dictionary selects its specified number of source blocks at both inference stages. Sharing the reference configuration across sweeps gives 27 trained conditions. \Cref{fig:supp-capacity-sweeps} shows their activation, spatial and temporal reconstruction scores. We also compare estimated description lengths for the 12 rank-sweep dictionaries at equal coding-error budgets (\cref{fig:supp-rank-mdl}).

\paragraph{Dictionary Dimension Selection.}
At rank 4, 1,024 blocks provide 4,096 decoder directions, matching the SAE dictionary width. The middle column of \cref{fig:supp-capacity-sweeps} also shows that 1,024 blocks gives the highest activation and spatial reconstruction scores for DriveWAM L15 and the highest scores on all three metrics for SimWAM L15. Increasing the dictionary to 2,048 or 4,096 blocks improves none of the metrics at any interface. A 512-block dictionary performs better for DriveLaW L18 and slightly better on DriveWAM L15's temporal metric; we retain 1,024 as a common dictionary size across models.

\paragraph{Number of Active Blocks Selection.}
We choose 64 selections to balance reconstruction and sparsity~\citep[Sec.~3.2]{felStructuringSparsityBlockSparse2026}. In the right column of \cref{fig:supp-capacity-sweeps}, increasing the budget from 32 to 64 improves every reconstruction metric at each interface. Selecting 128 blocks improves reconstruction further, but doubles the selected blocks and coefficient slots. With 64 selections, each token uses 6.25\% of the dictionary and 256 scalar coefficient slots at rank 4. This sets a common 64-unit selection budget for BSFs and SAEs, while preserving their different scalar capacities.

\paragraph{Rank Selection.}
We select rank~4 because it achieves the lowest estimated description length among all tested ranks across every evaluated interface (\cref{fig:supp-rank-mdl}). Although higher ranks improve reconstruction across all three metrics for DriveWAM~L15 and SimWAM~L15—and peak at rank~16 for activation reconstruction and rank~8 for spatial and temporal reconstruction on DriveLaW~L18 (\cref{fig:supp-capacity-sweeps}, left column)—these gains do not offset the additional bits required to encode larger blocks. Following the BSF description-length formulation~\citep[Sec.~3.3 and App.~C]{felStructuringSparsityBlockSparse2026}, our estimate accounts for block identities, coefficients, residuals, and model storage under a shared expected coding-error budget. Under this model, rank~4 yields the lowest description length at every tested error fraction (20\%, 10\%, 5\%, and 1\%) and across all 216 sensitivity comparisons varying storage penalties (including zero and BIC-style costs), token counts for storage amortisation, and the BSF per-coordinate noise convention, establishing it as the most compact tested configuration.

\FloatBarrier

\begin{supplementaryresults}
% Source: supplementary/sections/07-reconstruction-results.tex
\FloatBarrier
\section{Reconstruction results}
\label{sec:supp-r2-results}
\label{sec:supp-results}

Reconstruction gains differ across activation, spatial and temporal metrics.

\supplementaryresultbreak
% Source: supplementary/sections/reconstruction-results/04-residual-correction.tex
\subsection{spatialBSF Residual Analysis}
\label{sec:supp-residual-correction-diagnostic}

\subsubsection{Experiment and comparison}
Using the Dataset-Selected evaluation set, we examine how spatialBSF's initial reconstruction and residual correction contribute to its final output under each training regime. Following the evaluation procedure in \cref{sec:supp-reconstruction-evaluation}, we compare 36 spatialBSF checkpoints: fixed-rank and mixed-rank dictionaries trained separately on Dataset-Selected and Dataset-Full at each of nine DWM--layer interfaces. Each checkpoint is evaluated before and after residual correction on the same 100 Dataset-Selected validation scenes in common units, using shared targets, centring, and denominators. On average, residual correction contributes more when spatialBSF is trained on Dataset-Full across both rank variants.

\paragraph{Reported quantities.}
We report initial and final $R^2$ and the percentage of the initial sum of squared errors (SSE) removed. \begin{equation}
 100\left(1-\frac{\mathrm{SSE}_{\mathrm{final}}}{\mathrm{SSE}_{\mathrm{initial}}}\right).
 \label{eq:supp-refinement-error-removed}
\end{equation}
We calculate the percentage of initial squared error removed per interface before averaging. To assess drive-composition sensitivity, leave-one-drive-out ranges recompute the difference between Dataset-Full-trained and Dataset-Selected-trained dictionaries in this error-reduction percentage after omitting each validation drive.

\subsubsection{Correction contributes more after training on Dataset-Full}
For both fixed-rank spatialBSF and mixed-rank spatialBSF, residual correction removes a larger mean percentage of initial reconstruction error after Dataset-Full training than after Dataset-Selected training across all three metrics (\cref{tab:supp-refinement-summary}). For fixed-rank spatialBSF, this advantage holds at every interface. By contrast, mixed-rank spatialBSF exhibits exceptions for activation and spatial error at DriveLaW L14, as well as for spatial error at L18; consequently, its mean percentages of initial activation and spatial error removed across the three DriveLaW layers are lower after Dataset-Full training than after Dataset-Selected training. For each of the two rank configurations and all three reconstruction metrics, the nine-interface mean difference in proportional error removal between Dataset-Full-trained and Dataset-Selected-trained dictionaries remains positive after every drive omission (\cref{tab:supp-refinement-drive-sensitivity}).

% Source: supplementary/tables/04-09-refinement-summary.tex
% Source: docs/RESIDUAL_CORRECTION_RESULTS.json; saved evidence only.
\begin{table}[!htbp]
\centering
\begin{minipage}{0.94\textwidth}

\centering\small\setlength{\tabcolsep}{3pt}
\caption{\textbf{Residual correction in fixed-rank and mixed-rank spatialBSF.} All checkpoints use the same 100 Dataset-Selected validation scenes; dataset names identify dictionary training data. Values weight 9 interfaces equally. Removed is the percentage of initial squared error removed, calculated per interface before averaging. Performance rankings compare the two rank configurations within training dataset, stage and metric; higher $R^2$ is better. Error removal is unranked.}
\label{tab:supp-refinement-summary}
\begin{adjustbox}{max width=\linewidth}\begin{tabular}{@{}llrrrrrr@{}}
\toprule
 & & \multicolumn{3}{c}{Dataset-Selected} & \multicolumn{3}{c}{Dataset-Full}\\
\cmidrule(lr){3-5}\cmidrule(lr){6-8}
Variant & Metric & Initial $R^2$ & Final $R^2$ & Removed (\%) & Initial $R^2$ & Final $R^2$ & Removed (\%)\\
\midrule
spatialBSF & Activation & \textbf{0.6465} & \textbf{0.6696} & 6.6 & \underline{-1.0926} & \textbf{0.7575} & 74.5\\
spatialBSF & Spatial & \textbf{0.5325} & \textbf{0.5621} & 6.5 & \underline{0.0739} & \textbf{0.6661} & 51.4\\
spatialBSF & Temporal & \textbf{0.4189} & \textbf{0.4569} & 6.7 & \underline{0.1571} & \textbf{0.5643} & 41.3\\
Mixed-rank & Activation & \underline{0.5080} & \underline{0.6597} & 30.1 & \textbf{-0.6837} & \underline{0.7495} & 74.1\\
Mixed-rank & Spatial & \underline{0.4301} & \underline{0.5479} & 21.8 & \textbf{0.3560} & \underline{0.6576} & 43.2\\
Mixed-rank & Temporal & \underline{0.3866} & \underline{0.4479} & 10.6 & \textbf{0.3514} & \underline{0.5555} & 30.7\\
\bottomrule
\end{tabular}\end{adjustbox}

\end{minipage}
\end{table}

% Source: supplementary/tables/04-10-refinement-drive-sensitivity.tex
% Source: docs/RESIDUAL_CORRECTION_RESULTS.json; saved evidence only.
\begin{table}[!htbp]
\centering
\begin{minipage}{0.94\textwidth}

\centering\small\setlength{\tabcolsep}{3pt}
\caption{\textbf{Change in spatialBSF correction benefit with training dataset.} Differences are Dataset-Full-trained minus Dataset-Selected-trained results at matched interfaces, evaluated on the same 100 scenes and reported separately for fixed-rank and mixed-rank spatialBSF. $\Delta R^2$ compares initial-to-final gains. Error-removal differences and their leave-1-drive-out ranges are in percentage points. The last column counts interfaces with smaller proportional benefits after Dataset-Full training. Ranges span 8 drive omissions; they are not confidence intervals.}
\label{tab:supp-refinement-drive-sensitivity}
\begin{adjustbox}{max width=\linewidth}\begin{tabular}{@{}llrrrr@{}}
\toprule
Variant & Metric & $\Delta R^2$ difference & Removal (pp) & Drive-omission range (pp) & Dataset-Full smaller\\
\midrule
spatialBSF & Activation & +1.8270 & +67.9 & +67.9 to +68.0 & 0/9\\
spatialBSF & Spatial & +0.5626 & +44.9 & +44.7 to +45.2 & 0/9\\
spatialBSF & Temporal & +0.3693 & +34.6 & +33.6 to +35.1 & 0/9\\
Mixed-rank & Activation & +1.2815 & +44.0 & +43.5 to +44.9 & 1/9\\
Mixed-rank & Spatial & +0.1838 & +21.3 & +21.1 to +21.6 & 2/9\\
Mixed-rank & Temporal & +0.1429 & +20.2 & +19.2 to +20.9 & 0/9\\
\bottomrule
\end{tabular}\end{adjustbox}

\end{minipage}
\end{table}

\paragraph{Dependence on the Residual Correction.}
For fixed-rank and mixed-rank spatialBSF, dictionaries trained on Dataset-Full have negative initial activation $R^2$ averaged across the nine interfaces, yet after residual correction they exceed their Dataset-Selected-trained counterparts on all three mean reconstruction metrics on this common cohort. This pattern shows that the initial stage acts as a starting point within a jointly trained system rather than a standalone predictor: because only final reconstruction is penalised and residual correction receives information directly from the original activation, the two stages can cooperate effectively without requiring high initial accuracy. Section~J tests separately trained single-stage models to evaluate standalone initial capacity; however, comparing these training endpoints alone does not isolate which specific protocol differences drive the stronger residual dependence.

% Source: supplementary/sections/reconstruction-results/05-residual-correction-control.tex
\supplementaryresultbreak
\subsection{Controlling reliance on residual correction}
\label{sec:supp-residual-correction-control}
We hypothesise that the strong dependence of fixed-rank spatialBSF trained on Dataset-Full on residual correction limits its semantic agreement. We therefore explore simple controls for rebalancing the initial reconstruction and residual correction. Encoder-only fine-tuning raises the mean strength and direction EARS indices to 0.5936 and 0.5957, respectively, exceeding the best main-comparison scores of 0.5889 for spatialBSF strength and 0.5930 for Grassmannian BSF direction. This improvement comes with a 2.45\% increase in final activation squared reconstruction error. These results demonstrate that the existing spatialBSF architecture can achieve higher semantic agreement through encoder adaptation. We retain the original checkpoints in the main comparison because this exploratory fine-tuning was applied only to spatialBSF; including the adapted checkpoints would give it additional optimisation unavailable to the other methods.

\subsubsection{Fine-tuning the first stage across 9 interfaces}
\paragraph{3 recipes.}
Each recipe starts from the original Dataset-Full checkpoint. $\mathcal L_1$ and $\mathcal L_2$ are first-stage and final activation MSE. Both divide by the fitting scene's mean squared normalised activation, floored at $10^{-12}$. $\mathcal L_T$ compares the first reconstruction with the original frozen checkpoint's final reconstruction using that denominator. \begin{align}
 \mathcal L_{\mathrm{joint}} &= \mathcal L_2+0.25\mathcal L_1, \nonumber\\
 \mathcal L_{\mathrm{encoder}} &= \mathcal L_2+\mathcal L_1, \nonumber\\
 \mathcal L_{\mathrm{distill}} &= \mathcal L_2+0.25\mathcal L_1+\mathcal L_T.
 \label{eq:supp-residual-refinement-losses}
\end{align}
\begin{itemize}
\item All-parameter fitting updates every trainable parameter using $\mathcal L_{\mathrm{joint}}$.
\item Encoder-only fitting uses $\mathcal L_{\mathrm{encoder}}$.
\item Distillation uses $\mathcal L_{\mathrm{distill}}$; its frozen teacher is unnecessary at inference.
\end{itemize}
Encoder-only fitting updates the affine encoder and shared score scale and bias; decoder and correction parameters stay fixed. Distillation updates the same parameters. It changes both the first-stage weight and teacher term relative to encoder-only fitting. Selections can change at both stages.

Each recipe uses 512 drive-balanced fitting scenes and 1,024 Adam updates at a learning rate of $3\times10^{-5}$, accumulating token-weighted gradients over one complete scene and clipping the global gradient norm at 1. All-parameter fitting also applies the existing parameter projection, while normalisation, architecture, dictionary size, rank and selection budgets remain unchanged. Reconstruction is evaluated on the 6,653 Dataset-Full validation scenes, whereas EARS readouts are fitted on the separate 232-frame fitting cohort and evaluated on its 66-frame validation cohort. All reported results are averaged equally across the nine interfaces using one training seed per recipe.

% Source: supplementary/tables/04-16-residual-refinement.tex
\begin{table}[!htbp]
\centering
\begin{minipage}{0.94\textwidth}

\centering
\small
\setlength{\tabcolsep}{6pt}
\caption{\textbf{Fine-tuning fixed-rank spatialBSF trained on Dataset-Full.} Error changes are equal means of interface-relative SSE changes from the 9 original checkpoints; negative is better. Reconstruction uses 6,653 Dataset-Full validation scenes; EARS uses its separate 66-frame validation cohort. Final errors cover activation, spatial and temporal reconstruction. EARS also weights the 9 interfaces equally; higher is better. Performance rankings compare recipes within each column, including the original where reported; dashes are unranked.}
\label{tab:supp-residual-refinement}
\begin{adjustbox}{max width=\linewidth}\begin{tabular}{lrrrrrr}
\toprule
& \multicolumn{4}{c}{Mean SSE change (\%)} & \multicolumn{2}{c}{EARS index} \\
Recipe & First activation & Final activation & Final spatial & Final temporal & Strength & Direction \\
\midrule
Original & --- & --- & --- & --- & \thirdbest{0.5889} & \thirdbest{0.5895} \\
Encoder-only fine-tuning & \thirdbest{-30.18} & \textbf{+2.45} & \textbf{+2.13} & \textbf{+1.73} & \textbf{0.5936} & \textbf{0.5957} \\
Encoder distillation & \underline{-30.73} & \underline{+3.20} & \underline{+2.53} & \underline{+2.00} & \underline{0.5904} & \underline{0.5938} \\
All-parameter fine-tuning & \textbf{-57.71} & \thirdbest{+9.02} & \thirdbest{+6.67} & \thirdbest{+5.76} & 0.5830 & 0.5878 \\
\bottomrule
\end{tabular}\end{adjustbox}

\end{minipage}
\end{table}

\begin{figure}[!htbp]
\centering
\begin{minipage}{0.90\textwidth}

 \centering
 \includegraphics[width=\textwidth,height=0.40\textheight,keepaspectratio]{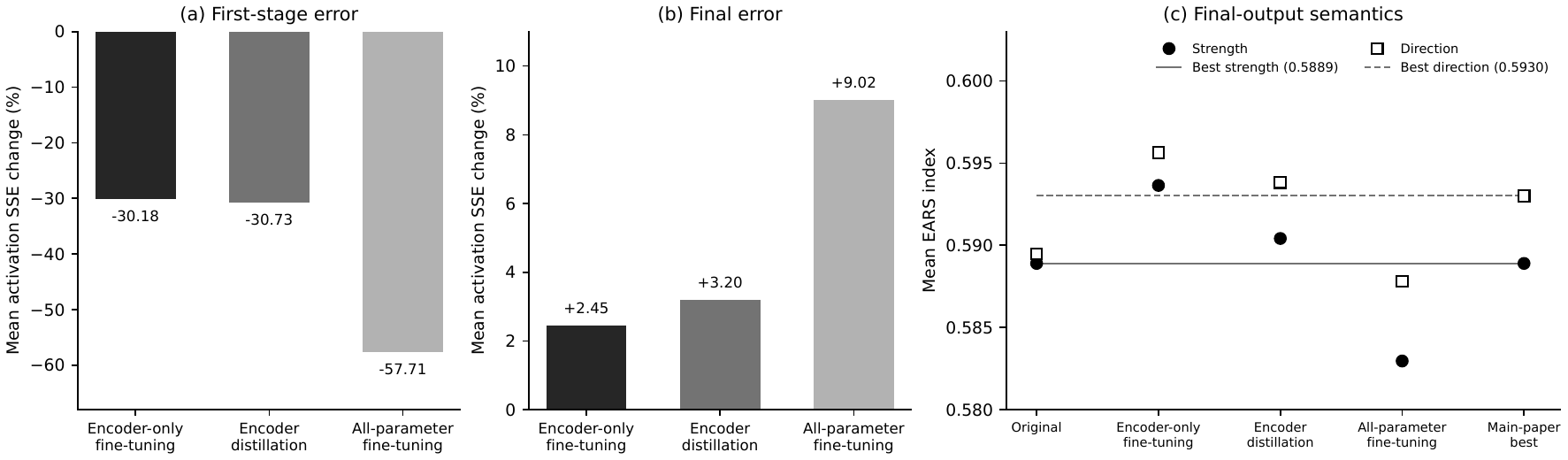}
 \caption{9-interface means after fine-tuning. (a,b) Interface-relative activation SSE changes; negative is better. (c) Final-output EARS scores; higher is better. Solid/dashed reference lines and the final column show the main-paper Dataset-Full best strength/direction scores (spatialBSF/Grassmannian BSF). Reconstruction and EARS use distinct cohorts.}
 \label{fig:supp-residual-refinement}

\end{minipage}
\end{figure}

\paragraph{Encoder adaptation improves EARS agreement.}
Encoder-only fine-tuning achieves the highest observed EARS indices among all compared recipes and original checkpoints, exceeding both the original fixed-rank spatialBSF strength EARS index and the original Grassmannian BSF direction EARS index, with a 2.45\% increase in final activation SSE relative to the unmodified fixed-rank spatialBSF checkpoints. Encoder distillation likewise surpasses both original EARS benchmarks, though at a larger reconstruction cost relative to its fixed-rank spatialBSF parent checkpoints. These results show that the existing architecture can support higher observed EARS agreement than the original checkpoints realise (\cref{tab:supp-residual-refinement,fig:supp-residual-refinement}).

All-parameter fitting improves the first reconstruction most, but increases final error by 9.02\% and lowers both EARS indices. The original checkpoints remain the main comparison because other methods receive no extra optimisation. These development-exposed gains vary by interface.

\supplementaryresultbreak
% Source: supplementary/sections/08-interpretability-results.tex
\FloatBarrier
\section{\terminologyedit{Semantic evaluation} results}
\label{sec:supp-semantic-results}

\begin{recentedits}
% Source: supplementary/sections/interpretability-results/02-full-checkpoints.tex
\subsection{Dataset-Full: checkpoints and readout evaluation}
\label{sec:supp-full-interpretability}
We evaluate 7 Dataset-Full methods at all 9 interfaces, giving 36 BSF and 27 SAE fits. \cref{sec:supp-training-protocol}--\cref{sec:supp-sae-baselines} define training and selection; semantic scores do not choose checkpoints. Evaluation uses native outputs, including spatialBSF's integrated correction.

% Source: supplementary/sections/interpretability-results/03-full-common-units.tex
\subsubsection{Normalisation and common channel units}
\label{sec:supp-full-common-units}
We convert Dataset-Full reconstructions into Dataset-Selected units to share projections and native controls. Fitting means $\vmu_f,\vmu_s$ and reconstructions are rows in $\mathbb R^{1\times d}$; $\sigma_f,\sigma_s$ are scalar RMS scales.
\begin{equation}
 \widehat{\vx}_s=\frac{\sigma_f\widehat{\vx}_f+\vmu_f-\vmu_s}{\sigma_s}.
\end{equation}
Relation and state readouts pool these converted reconstructions over entity regions before applying the shared projection. Individual block contributions receive only the scale $\sigma_f/\sigma_s$. Decoder bias and mean offsets enter collective features, not individual localisation maps. The encoder and source selections are unchanged.

% Source: supplementary/sections/interpretability-results/04-full-readouts.tex
\subsubsection{Readout fitting and task cohorts}
\label{sec:supp-full-readouts}
Dictionaries trained on Dataset-Selected and Dataset-Full are evaluated using the same semantic cohorts and fitting procedure described in \cref{sec:supp-interpretability}, with outputs from Dataset-Full-trained dictionaries first converted to common units as detailed above and each dictionary selecting its own localisation detectors.

% Source: supplementary/sections/interpretability-results/05-full-results.tex
Readout fitting and validation drives are disjoint; overlap with Dataset-Full dictionary training remains unverified.

% Source: supplementary/tables/06-14-family-layer-means.tex
\begin{table}[!htbp]
\centering
\begin{minipage}{0.94\textwidth}

\recenteditcolor\recenteditcaption
\centering\small\setlength{\tabcolsep}{4pt}
\caption{\textbf{Semantic scores by method family and layer for dictionaries trained on Dataset-Full.} Methods receive equal weight within each family. SAEs comprise signed SAE, Matryoshka TopK and RA-SAE; BSFs comprise Vanilla BSF, Grassmannian BSF, fixed-rank spatialBSF and mixed-rank spatialBSF. Panels (a,b) use strength localisation; (c) uses direction. Relation and state terms are shared across these readouts. Performance rankings compare layers within family, DWM and metric; higher is better.}
\label{tab:supp-family-layer-means}
\textbf{(a) DriveWAM: strength localisation, relations and states}\par\smallskip
\begin{adjustbox}{max width=\linewidth}\begin{tabular}{@{}lrrrrrr@{}}
\toprule
 & \multicolumn{3}{c}{SAE mean} & \multicolumn{3}{c}{BSF mean}\\
\cmidrule(lr){2-4}\cmidrule(l){5-7}
Metric & L7 & L15 & L23 & L7 & L15 & L23\\
\midrule
Car F1 & \thirdbest{0.2579} & \textbf{0.2868} & \underline{0.2708} & \underline{0.3590} & \textbf{0.3957} & \thirdbest{0.2981}\\
Large-vehicle F1 & \thirdbest{0.1701} & \textbf{0.2447} & \underline{0.2380} & \underline{0.2573} & \textbf{0.2747} & \thirdbest{0.2510}\\
Road F1 & \textbf{0.5455} & \underline{0.4908} & \thirdbest{0.3505} & \textbf{0.7479} & \underline{0.7016} & \thirdbest{0.6281}\\
Relation/composition $A_{\mathrm{rel}}$ & \underline{0.7094} & \textbf{0.7339} & \thirdbest{0.6816} & \underline{0.7148} & \textbf{0.7507} & \thirdbest{0.6699}\\
Red/green BA & \thirdbest{0.5444} & \textbf{0.8002} & \underline{0.5571} & \underline{0.6106} & \thirdbest{0.5849} & \textbf{0.6190}\\
Stationary/moving BA & \underline{0.7498} & \textbf{0.8329} & \thirdbest{0.7430} & \thirdbest{0.7306} & \textbf{0.8179} & \underline{0.7685}\\
\bottomrule
\end{tabular}\end{adjustbox}
\par\medskip
\textbf{(b) SimWAM: strength localisation, relations and states}\par\smallskip
\begin{adjustbox}{max width=\linewidth}\begin{tabular}{@{}lrrrrrr@{}}
\toprule
 & \multicolumn{3}{c}{SAE mean} & \multicolumn{3}{c}{BSF mean}\\
\cmidrule(lr){2-4}\cmidrule(l){5-7}
Metric & L7 & L15 & L23 & L7 & L15 & L23\\
\midrule
Car F1 & \thirdbest{0.2470} & \textbf{0.2707} & \underline{0.2509} & \underline{0.3796} & \textbf{0.3879} & \thirdbest{0.3207}\\
Large-vehicle F1 & \underline{0.2275} & \textbf{0.3089} & \thirdbest{0.2183} & \textbf{0.2742} & \underline{0.2361} & \thirdbest{0.1871}\\
Road F1 & \textbf{0.3668} & \underline{0.3482} & \thirdbest{0.3266} & \textbf{0.6325} & \underline{0.6236} & \thirdbest{0.5762}\\
Relation/composition $A_{\mathrm{rel}}$ & \underline{0.7426} & \textbf{0.7780} & \thirdbest{0.7134} & \underline{0.7562} & \textbf{0.8104} & \thirdbest{0.7186}\\
Red/green BA & \underline{0.6082} & \textbf{0.7993} & \thirdbest{0.5152} & \underline{0.6541} & \textbf{0.7290} & \thirdbest{0.6097}\\
Stationary/moving BA & \thirdbest{0.7052} & \textbf{0.8630} & \underline{0.7622} & \thirdbest{0.7091} & \textbf{0.8446} & \underline{0.7983}\\
\bottomrule
\end{tabular}\end{adjustbox}
\par\medskip
\textbf{(c) Direction localisation: L7 / L15 / L23}\par\smallskip
\begin{adjustbox}{max width=\linewidth}\begin{tabular}{@{}llrrr@{}}
\toprule
DWM & Family & Car F1 & Large-vehicle F1 & Road F1\\
\midrule
DriveWAM & SAE & \thirdbest{0.2553} / \textbf{0.2845} / \underline{0.2737} & \thirdbest{0.1682} / \textbf{0.2427} / \underline{0.2380} & \textbf{0.5453} / \underline{0.4908} / \thirdbest{0.3505}\\
DriveWAM & BSF & \underline{0.3280} / \textbf{0.3486} / \thirdbest{0.3011} & \thirdbest{0.2295} / \underline{0.2671} / \textbf{0.2775} & \textbf{0.7889} / \underline{0.7010} / \thirdbest{0.6389}\\
\midrule
SimWAM & SAE & \thirdbest{0.2472} / \textbf{0.2770} / \underline{0.2527} & \underline{0.2265} / \textbf{0.3085} / \thirdbest{0.2207} & \textbf{0.3668} / \underline{0.3482} / \thirdbest{0.3266}\\
SimWAM & BSF & \textbf{0.3890} / \underline{0.3524} / \thirdbest{0.3223} & \underline{0.2594} / \textbf{0.2839} / \thirdbest{0.2153} & \textbf{0.6624} / \underline{0.6434} / \thirdbest{0.5942}\\
\bottomrule
\end{tabular}\end{adjustbox}

\end{minipage}
\end{table}

\end{recentedits}
% Source: supplementary/sections/interpretability-results/06-road-spatial-transport.tex
\subsection{Road localisation and neighbour-placement controls}
\label{sec:supp-road-spatial-transport}

Rearranging neighbour transport reduces road-localisation strength F1 from 0.6493 to 0.5087 on average for 2-stage Dataset-Selected spatialBSF. Every tested permutation reduces strength F1 at every interface. Dataset-Full spatialBSF follows this pattern except DriveWAM L23, where 7 of 8 permutations reduce F1. Direction effects are less uniform and sometimes positive (\cref{tab:supp-road-placement-controls}).

Spatial context can contribute to localisation through where neighbouring signals arrive, as well as through their magnitude. We test the role of spatial arrangement by perturbing only the selected road block's neighbour transport while holding its local contribution and all other blocks fixed. Specifically, we apply eight degree-preserving permutations and rescale the transported field to preserve its norm—a manipulation that preserves neither total reconstruction energy nor road-weighted energy. In \cref{tab:supp-road-placement-controls}, we evaluate four dictionary configurations across the nine interfaces on the same 65 road-positive scenes: two-stage and one-stage fixed-rank spatialBSF trained on Dataset-Selected, alongside two-stage fixed-rank and mixed-rank spatialBSF trained on Dataset-Full. Readout choice, direction, threshold and source codes remain fixed throughout, and results are averaged over the eight permutations using the scoring rules from \cref{sec:supp-interpretability}.

\begin{table}[!htbp]
\centering
\begin{minipage}{0.94\textwidth}

\centering\small\setlength{\tabcolsep}{5pt}
\caption{\textbf{Road localisation after rearranging spatialBSF neighbour transport.} F1 uses the original 65-scene road cohort with source codes and readout settings fixed. Dataset names identify dictionary training data. Dataset-Selected rows compare two-stage and one-stage fixed-rank spatialBSF; Dataset-Full rows compare two-stage fixed-rank and mixed-rank spatialBSF. ``Permuted F1'' averages 8 norm-matched permutations; ``All 8 lower'' counts interfaces where every permutation lowers F1 relative to the learned arrangement.}
\label{tab:supp-road-placement-controls}
\begin{adjustbox}{max width=\linewidth}\begin{tabular}{@{}lllrrc@{}}
\toprule
Training & Configuration & Readout & Correct F1 & Permuted F1 & All 8 lower\\
\midrule
Dataset-Selected & spatialBSF & Strength & 0.6493 & 0.5087 & 9/9\\
Dataset-Selected & spatialBSF & Direction & 0.6696 & 0.5351 & 8/9\\
Dataset-Selected & 1 stage & Strength & 0.5857 & 0.4728 & 8/9\\
Dataset-Selected & 1 stage & Direction & 0.6074 & 0.4648 & 9/9\\
Dataset-Full & spatialBSF & Strength & 0.6748 & 0.5060 & 8/9\\
Dataset-Full & spatialBSF & Direction & 0.6860 & 0.6079 & 5/9\\
Dataset-Full & Mixed-rank spatialBSF & Strength & 0.6560 & 0.4816 & 9/9\\
Dataset-Full & Mixed-rank spatialBSF & Direction & 0.6870 & 0.5964 & 6/9\\
\bottomrule
\end{tabular}\end{adjustbox}

\end{minipage}
\end{table}

The losses show that the learned neighbour arrangement supports road localisation in these readouts. This frozen placement test complements the separately trained local-decoder comparison (\cref{tab:supp-ablation-portfolio-tasks}); it measures localisation, not driving performance.

\supplementaryresultbreak
% Source: supplementary/sections/09-mean-activity.tex
\FloatBarrier
\section{Mean-activity comparisons}
\label{sec:supp-lasso-average-budget}

In the main comparison, the budget allocates 64 selected dictionary-unit slots separately at every token. By contrast, the mean-activity methods here target an average activity of 64 dictionary units across tokens, allowing individual tokens to activate more or fewer units. Measured activity counts dictionary units with nonzero coefficients, and the achieved mean activity can differ from the target across methods and evaluation cohorts.

We adapt spatialBSF to this mean-activity setting and compare it with Lasso-BSF and Matryoshka BatchTopK.

\begin{recentedits}
% Source: supplementary/sections/mean-activity/01-protocol.tex
\subsection{Mean-activity selection at 9 interfaces}
\label{sec:supp-mean-activity-protocol}

We compare Lasso-BSF, mean-activity spatialBSF and Matryoshka BatchTopK at a target mean activity of 64~\citep{bussmann2025matryoshka}.
Individual tokens may use more or fewer units.
Each BSF unit is a rank-4 block; BatchTopK units are scalar features.
The shared target therefore matches neither scalar capacity nor achieved activity.

\paragraph{BSF selectors.}
Both Lasso-BSF and mean-activity spatialBSF train fresh 1,024-block dictionaries under the Dataset-Selected protocol and select whole blocks whose coefficient norms exceed learned thresholds without coefficient shrinkage:
\begin{equation}
 \vz^{(s)}_{pb}=\va^{(s)}_{pb}\,\mathbf{1}\!\left\{\|\va^{(s)}_{pb}\|_2>\theta^{(s)}_b\right\},
\end{equation}
where each selected block retains all four coefficients and a Lagrange controller regulates the mean active block count. While Lasso-BSF applies this selection rule in a single stage ($s=1$), mean-activity spatialBSF learns separate thresholds and positive multipliers for its initial and residual-correction stages ($s \in \{1, 2\}$), retaining the spatial decoder and residual-correction architecture of fixed-rank spatialBSF.

Training minimises final reconstruction MSE plus $\sum_s\lambda_s\bar K_s$, where $\bar K_s$ is mean stage-s source activity. The controller increases sparsity pressure above mean activity 64 and decreases it below 64. Thresholds use surrogate gradients; forward selection remains whole-block. Evaluation uses the learned thresholds without a per-token cap, so validation means can differ from 64.
Lasso-BSF constrains each decoder vector to unit norm, whereas spatialBSF uses orthonormal columns within each block.

\paragraph{BatchTopK selector.}
The 9 BatchTopK endpoints use 4,096 nonnegative scalar features with the same fitting split, update budget and learning-rate schedule.
Its nested reconstruction losses differ from the BSF objectives (\cref{sec:supp-sae-baselines}).
BatchTopK selects activations jointly over a training batch and applies a saved inference threshold calibrated after update 1,000.

% Source: supplementary/sections/mean-activity/02-current-results.tex
\subsection{Reconstruction and measured activity}
\label{sec:supp-mean-activity-results}
\label{sec:supp-mean-activity-reconstruction}
Mean-activity spatialBSF has the highest activation $R^2$ among mean-activity methods in every DWM. Achieved activity varies by method and cohort despite the common target. BSF tokens can activate almost the whole dictionary, and semantic-cohort means differ from reconstruction-cohort means. BatchTopK also exceeds 64 features on some tokens (\cref{tab:supp-lasso-average-reconstruction}).
\par\smallskip
\noindent{\small\begin{tabularx}{0.99\linewidth}{@{}Yrr@{}}
\toprule MI tool / cohort & Interface means & Token max.\\\midrule
Lasso-BSF / reconstruction & 59.52--68.01 & 939\\
Mean-activity spatialBSF / reconstruction & 61.77--68.69 & 975\\
Lasso-BSF / semantic & 49.02--85.98 & ---\\
Mean-activity spatialBSF / semantic & 51.11--91.77 & ---\\
BatchTopK / reconstruction & 61.40--71.30 & 536\\\bottomrule
\end{tabularx}}
\par\smallskip
BSFs count whole source blocks before transport, using spatialBSF's final stage.

% Source: supplementary/tables/07-03-lasso-average-budget-reconstruction.tex
\begin{table}[!htbp]
\centering
\begin{minipage}{0.94\textwidth}

\recenteditcolor
\recenteditcaption
\centering\small\setlength{\tabcolsep}{3pt}
\caption{\textbf{Mean-activity reconstruction with dictionaries trained on Dataset-Selected.} Values weight 3 layers per DWM equally, using the Dataset-Selected validation cohort and \cref{sec:supp-reconstruction-evaluation}'s metrics. Activity counts nonzero source units: whole blocks for Lasso-BSF and spatialBSF, or scalar features for Matryoshka BatchTopK. Achieved activity and coefficient budgets differ. Performance rankings compare reconstruction within DWM and metric; higher $R^2$ is better. Activity is unranked.}
\label{tab:supp-lasso-average-reconstruction}
\begin{adjustbox}{max width=\linewidth}\begin{tabular}{@{}llrrrr@{}}
\toprule
DWM & MI tool & $R^2_{\mathrm{act}}$ & $R^2_{\mathrm{sp}}$ & $R^2_{\mathrm{temp}}$ & \shortstack{Mean active\\dictionary units}\\
\midrule
DriveLaW & Lasso-BSF & \underline{0.7543} & \underline{0.6353} & \underline{0.4394} & 66.25\\
DriveLaW & Mean-activity spatialBSF & \textbf{0.8055} & \textbf{0.7129} & \textbf{0.5491} & 64.53\\
DriveLaW & Matryoshka BatchTopK & \thirdbest{0.7477} & \thirdbest{0.6269} & \thirdbest{0.4123} & 66.40\\
DriveWAM & Lasso-BSF & \underline{0.6153} & \underline{0.4951} & \underline{0.4129} & 63.12\\
DriveWAM & Mean-activity spatialBSF & \textbf{0.6524} & \textbf{0.5515} & \textbf{0.5013} & 65.19\\
DriveWAM & Matryoshka BatchTopK & \thirdbest{0.5338} & \thirdbest{0.3950} & \thirdbest{0.3006} & 67.14\\
SimWAM & Lasso-BSF & \underline{0.5594} & \underline{0.4748} & \underline{0.4149} & 62.57\\
SimWAM & Mean-activity spatialBSF & \textbf{0.5962} & \textbf{0.5234} & \textbf{0.4723} & 64.38\\
SimWAM & Matryoshka BatchTopK & \thirdbest{0.4593} & \thirdbest{0.3584} & \thirdbest{0.2952} & 65.55\\
\bottomrule
\end{tabular}\end{adjustbox}

\end{minipage}
\end{table}

% Source: supplementary/sections/mean-activity/04-eleven-task-results.tex
\subsection{11-task semantic agreement}
\label{sec:supp-mean-activity-semantics}

The 3 mean-activity methods have similar EARS indices despite different reconstruction fidelity. BatchTopK has the highest observed aggregate indices and large-vehicle F1; mean-activity spatialBSF has the highest direction road F1 (\cref{tab:supp-ears-eleven-task-mean-activity}). All 27 checkpoints follow D's cohorts and readout protocol.

% Source: supplementary/tables/07-08-ears-eleven-task-mean-activity.tex
\begin{table}[!htbp]
\centering
\begin{minipage}{0.94\textwidth}

\recenteditcolor\recenteditcaption
\centering\small\setlength{\tabcolsep}{4pt}
\caption{\textbf{EARS for mean-activity dictionaries trained on Dataset-Selected.} (a) Task means over 9 equally weighted interfaces; the index weights all 11 tasks equally. Evaluation uses the common semantic validation cohorts. Large vehicle combines truck and bus regions, excluding car. Each method targets mean activity of 64 units; achieved activity and scalar capacity differ. Performance rankings compare methods within each row and readout; higher is better.}
\label{tab:supp-ears-eleven-task-mean-activity}
\begin{adjustbox}{max width=\linewidth}\begin{tabular}{@{}lrrrrrr@{}}
\toprule
& \multicolumn{2}{c}{Lasso-BSF} & \multicolumn{2}{c}{Mean-activity spatialBSF} & \multicolumn{2}{c}{Matryoshka BatchTopK}\\
\cmidrule(lr){2-3}\cmidrule(lr){4-5}\cmidrule(l){6-7}
& Strength & Direction & Strength & Direction & Strength & Direction\\
\midrule
\multicolumn{7}{l}{\textbf{(a) Task means across all 9 interfaces}}\\
\addlinespace[2pt]
Car localisation F1 & \underline{0.2650} & \underline{0.2877} & \thirdbest{0.2480} & \thirdbest{0.2700} & \textbf{0.2988} & \textbf{0.2988}\\
Large-vehicle localisation F1 & \thirdbest{0.1775} & \underline{0.2140} & \underline{0.1796} & \thirdbest{0.1898} & \textbf{0.2218} & \textbf{0.2218}\\
Road localisation F1 & \textbf{0.5668} & \underline{0.5920} & \underline{0.5643} & \textbf{0.6021} & \thirdbest{0.5301} & \thirdbest{0.5301}\\
Traffic-light localisation F1 & \thirdbest{0.0194} & \thirdbest{0.0374} & \underline{0.0308} & \textbf{0.0396} & \textbf{0.0377} & \underline{0.0377}\\
Ahead of vehicle BA & \thirdbest{0.8104} & \thirdbest{0.8104} & \underline{0.8151} & \underline{0.8151} & \textbf{0.8399} & \textbf{0.8399}\\
Left of vehicle BA & \thirdbest{0.8910} & \thirdbest{0.8910} & \underline{0.9200} & \underline{0.9200} & \textbf{0.9224} & \textbf{0.9224}\\
Vehicle in lane BA & \underline{0.7045} & \underline{0.7045} & \textbf{0.7088} & \textbf{0.7088} & \thirdbest{0.6961} & \thirdbest{0.6961}\\
Same-lane vehicle BA & \thirdbest{0.7525} & \thirdbest{0.7525} & \textbf{0.8137} & \textbf{0.8137} & \underline{0.8013} & \underline{0.8013}\\
Car ahead in same lane F1 & \textbf{0.3753} & \textbf{0.3753} & \thirdbest{0.3589} & \thirdbest{0.3589} & \underline{0.3662} & \underline{0.3662}\\
Traffic light red BA & \textbf{0.6172} & \textbf{0.6172} & \thirdbest{0.5958} & \thirdbest{0.5958} & \underline{0.6155} & \underline{0.6155}\\
Car stationary BA & \underline{0.7747} & \underline{0.7747} & \thirdbest{0.7725} & \thirdbest{0.7725} & \textbf{0.7819} & \textbf{0.7819}\\
\midrule
$I_\chi$ (11 tasks) & \thirdbest{0.5413} & \thirdbest{0.5506} & \underline{0.5461} & \underline{0.5533} & \textbf{0.5556} & \textbf{0.5556}\\
\midrule
\bottomrule
\end{tabular}\end{adjustbox}

\end{minipage}
\end{table}

% Source: supplementary/sections/mean-activity/05-fixed-unit-comparison.tex
\subsection{Comparison with fixed per-token selection}
\label{sec:supp-mean-activity-fixed-comparison}

On aggregate EARS, all 3 mean-activity methods score below fixed-unit BSFs and Matryoshka TopK, while remaining above signed SAE and RA-SAE.
Reconstruction does not consistently track this semantic agreement. BatchTopK reconstructs less accurately than mean-activity BSFs but scores slightly higher on EARS. Compared with fixed Matryoshka TopK, BatchTopK improves activation reconstruction while lowering EARS.
These results compare the fitted methods under their respective selection assumptions. Their objectives and scalar capacities also differ, so they do not isolate the effect of varying the number of selected units per token.

\end{recentedits}

\supplementaryresultbreak
% Source: supplementary/sections/10-component-ablation.tex
\FloatBarrier
\section{spatialBSF component ablation}
\label{sec:supp-bsf-ablation}

Residual correction remains useful despite the Dataset-Full first-stage difficulties in \cref{sec:supp-residual-correction-diagnostic,sec:supp-residual-correction-control}. In separately trained Dataset-Selected ablations, it gives the most consistent gains across interfaces. Encoder and transport effects vary. We examine these component effects, then score calibration and block rank.

\begin{recentedits}
% Source: supplementary/sections/component-ablation/01-design-and-training.tex
\subsection{Controlled design and training}
\label{sec:supp-ablation-design}
Reference spatialBSF uses an independent encoder, neighbour transport and 2 stages with residual correction.
We compare 4 separately trained ablations at 9 interfaces.
The tied encoder instead projects onto the decoder basis with a positive learned block gain; bias and shared rotation remain.
The local-decoder variant removes neighbour transport but keeps a bounded local gain and consistent adjoint.
The 1-stage variant removes residual feedback and reselection together; local + 1 stage also removes transport.
The 4 independent-encoder configurations test transport and residual correction in a $2\times2$ design; the tied encoder tests encoder choice within the full architecture.

All 45 fits start fresh under the Dataset-Selected protocol.
Normalisers and scene order are shared; initial decoded dictionaries and proposals match up to numerical round-off in single-precision arithmetic.
Each gate selects 64 of 1,024 rank-4 blocks, allocating 256 coefficient slots.

% Source: supplementary/sections/component-ablation/02-evaluation-and-findings.tex
\subsection{Evaluation and findings}
\label{sec:supp-ablation-evaluation}
Adding residual correction yields the most consistent gains across separately trained architectures (\cref{tab:supp-ablation-contrasts}). Among dictionaries with neighbour transport, two-stage spatialBSF outperforms one-stage spatialBSF on every reconstruction metric and both EARS indices at all nine interfaces; with a local decoder, the corresponding two-stage versus one-stage comparison improves reconstruction at all nine interfaces and EARS at eight. Within two-stage spatialBSF, tying the initial encoder to the decoder basis usually improves activation fidelity, whereas learning the encoder independently gives higher aggregate EARS indices. Finally, comparing neighbour transport against a local decoder shows that transport improves reconstruction most consistently among one-stage architectures, whereas its gains are less uniform among two-stage architectures (\cref{tab:supp-ablation-contrasts}).

Evaluation follows B and \cref{sec:supp-interpretability}, using unablated spatialBSF as reference.
Drive contrasts and bootstrap intervals describe within-cohort variation.

\subsubsection{Clarification of absolute score gains}
\label{sec:supp-ablation-absolute-gains}
Since gains compare scores on their original scales, they are absolute differences: 2-stage minus 1-stage is 0.0731 for activation $R^2$, 0.0466 for strength EARS and 0.0411 for direction EARS.
Using unrounded 9-interface means, these are the main paper's 7.31, 4.66 and 4.11 percentage-point gains, not relative percentage increases (main Table~5; \cref{tab:supp-ablation-summary}).

\end{recentedits}

% Source: supplementary/sections/component-ablation/06-score-calibration.tex
\begin{samepage}
\subsection{Calibrated versus raw-energy block selection}
\label{sec:supp-score-calibration}

We next test inference-time block selection: calibration gives modest mean gains, with mixed semantic effects across interfaces.
We replace calibrated scores with raw energy at both selections in the 9 Dataset-Selected reference checkpoints.
This sets $\gamma_{br}$ = 1 and $v_{br}$ = 0; the shared energy normaliser does not alter ordering (\cref{eq:supp-prefix-score}).
All other trained parameters stay fixed; the changed initial reconstruction produces a newly computed correction.
We refit semantic readouts under D's protocol.
This tests inference-time calibration at fixed rank, not training without calibration.
\end{samepage}
\par\smallskip
\noindent{\small\begin{tabular}{@{}lrrr@{}}
\toprule Metric & Calibrated & Raw & Higher / 9\\\midrule
Activation $R^2$ & 0.6696 & 0.6614 & 7\\
Spatial $R^2$ & 0.5621 & 0.5501 & 7\\
Temporal $R^2$ & 0.4569 & 0.4411 & 6\\
EARS strength & 0.5785 & 0.5724 & 5\\
EARS direction & 0.5802 & 0.5765 & 4\\\bottomrule
\end{tabular}}
\par\smallskip
\noindent ``Higher / 9'' counts the interfaces (out of nine) where the reported evaluation metric is higher with calibrated block selection than with raw-energy block selection.

% Source: supplementary/sections/component-ablation/07-fixed-mixed-rank.tex
\subsection{Fixed and mixed rank}
\label{sec:supp-fixed-mixed-rank}

\subsubsection{Different ranks coexist across the tested noise levels}
Rank coexistence means selecting at least 2 distinct ranks at one source token; any pair among ranks 1, 2 and 4 qualifies.
All 45 mixed-rank spatialBSF checkpoints exhibit coexistence across 3 synthetic datasets, 3 seeds and 5 noise levels.
Without added noise, ranks coexist in 1,652 of 9,216 token evaluations (17.93\%).
Across noise levels, 13.35--18.14\% of token evaluations show coexistence, in all 9 checkpoints at each level.
This coexistence shows that mixed-rank spatialBSF learns to allocate different amounts of capacity within a token, using fewer coordinates for some selected blocks and retaining higher ranks for others. Such allocation permits more economical codes than assigning full rank to every selected block. The efficiency gain concerns selected coefficient capacity; the stored dictionary parameters are unchanged.
The table reports token-evaluation percentages; noise is the Gaussian noise-to-clean-signal energy ratio for each separate fit.
% Source: supplementary/tables/rank-coexistence.tex
\begin{center}
\small
\begin{tabular}{@{}rr@{}}
\toprule
Noise level & Tokens with rank coexistence (\%)\\
\midrule
0 & 17.93\\
0.1 & 18.14\\
0.25 & 16.89\\
0.5 & 16.96\\
1 & 13.35\\
\bottomrule
\end{tabular}
\end{center}

Each dataset contains 32 factors from 8 shape families, with 3 factors per token.
Training uses 2,400 updates, 128 blocks and exactly 3 selections per token; each checkpoint contributes 16 test graphs of 64 tokens.
These ranks describe coefficient allocation before spatial transport, not recovery of planted feature dimensions.
The shapes' linear spans have 1, 2 or 3 dimensions, distinct from the allowed ranks 1, 2 or 4 and intrinsic manifold dimension.
All 3 datasets were used during development and reused across seeds.

% Source: supplementary/tables/08-01-bsf-ablation-summary.tex
\begin{table}[!htbp]
\centering
\begin{minipage}{0.94\textwidth}

\recenteditcolor\recenteditcaption
\centering\small\setlength{\tabcolsep}{3pt}
\caption{\textbf{EARS for five spatialBSF configurations trained on Dataset-Selected.} Values weight 9 interfaces equally. Component means and the 11-task index follow \cref{sec:supp-interpretability-aggregation}. Performance rankings compare configurations within each column; higher is better.}
\label{tab:supp-ablation-summary}
\begin{adjustbox}{max width=\linewidth}\begin{tabular}{@{}lrrrrrr@{}}
\toprule
Configuration & $A_{\mathrm{rel}}$ & $F_{\mathrm{str}}$ & $F_{\mathrm{dir}}$ & $A_{\mathrm{state}}$ & $I_{\mathrm{str}}$ & $I_{\mathrm{dir}}$\\
\midrule
spatialBSF & \underline{0.7291} & \textbf{0.3128} & \underline{0.3175} & \underline{0.7335} & \textbf{0.5785} & \textbf{0.5802}\\
Tied encoder & \thirdbest{0.7083} & \underline{0.3055} & \textbf{0.3299} & \thirdbest{0.7136} & \thirdbest{0.5628} & \thirdbest{0.5717}\\
Local decoder & \textbf{0.7307} & \thirdbest{0.2910} & \thirdbest{0.3000} & \textbf{0.7395} & \underline{0.5724} & \underline{0.5757}\\
1 stage & 0.6999 & 0.2398 & 0.2598 & 0.6961 & 0.5319 & 0.5392\\
Local + 1 stage & 0.7046 & 0.2177 & 0.2249 & 0.7052 & 0.5277 & 0.5303\\
\bottomrule
\end{tabular}\end{adjustbox}

\end{minipage}
\end{table}

% Source: supplementary/tables/08-02-bsf-ablation-contrasts.tex
\begin{table}[!htbp]
\centering
\begin{minipage}{0.94\textwidth}

\recenteditcolor\recenteditcaption
\centering\small\setlength{\tabcolsep}{3pt}
\caption{\textbf{Interfaces where a spatialBSF component improves the score.} Counts compare separately trained Dataset-Selected configurations at 9 matched interfaces before rounding. Semantic columns use 11-task EARS on the common cohorts. The independent-versus-decoder-tied encoder comparison retains neighbour transport and 2 stages. The other contrasts compare neighbour transport with a local decoder, and 2 stages with 1, in a $2\times2$ design.}
\label{tab:supp-ablation-contrasts}
\begin{adjustbox}{max width=\linewidth}\begin{tabular}{@{}lrrrrr@{}}
\toprule
Retained component / context & $R^2_{\mathrm{act}}$ & $R^2_{\mathrm{sp}}$ & $R^2_{\mathrm{temp}}$ & $I_{\mathrm{str}}$ & $I_{\mathrm{dir}}$\\
\midrule
Independent encoder & 1/9 & 1/9 & 2/9 & 7/9 & 6/9\\
Transport with 2 stages & 7/9 & 4/9 & 4/9 & 5/9 & 4/9\\
Residual stage with transport & 9/9 & 9/9 & 9/9 & 9/9 & 9/9\\
Transport with 1 stage & 9/9 & 9/9 & 8/9 & 6/9 & 6/9\\
Residual stage with local decoder & 9/9 & 9/9 & 9/9 & 8/9 & 8/9\\
\bottomrule
\end{tabular}\end{adjustbox}

\end{minipage}
\end{table}

% Source: supplementary/tables/08-03-bsf-ablation-portfolio-tasks.tex
\begin{table}[!htbp]
\centering
\begin{minipage}{0.94\textwidth}

\recenteditcolor\recenteditcaption
\centering\scriptsize\setlength{\tabcolsep}{2pt}
\caption{\textbf{Task scores for spatialBSF configurations trained on Dataset-Selected.} Values weight 9 interfaces equally. Relations and states use balanced accuracy (BA); conjunction and localisation use F1. Localisation cells give strength/direction scores on the common positive-reference cohorts in \cref{sec:supp-interpretability-population}. Large denotes the truck/bus union; Red and Activity denote light state and car activity. Performance rankings compare configurations within task/readout; higher is better.}
\label{tab:supp-ablation-portfolio-tasks}
\resizebox{\linewidth}{!}{%
\begin{adjustbox}{max width=\linewidth}\begin{tabular}{@{}lrrrrrrrrrrrrrrr@{}}
\toprule
Configuration & Ahead & Left & Member & Shared & C & Car str & Car dir & Large str & Large dir & Road str & Road dir & Light str & Light dir & Red BA & Activity BA\\
\midrule
spatialBSF & \textbf{0.8430} & \textbf{0.9469} & \thirdbest{0.7146} & \thirdbest{0.7767} & \underline{0.3643} & \textbf{0.3199} & \thirdbest{0.3201} & \thirdbest{0.2293} & \underline{0.2277} & \textbf{0.6493} & \underline{0.6696} & \textbf{0.0528} & \textbf{0.0527} & \underline{0.6643} & \underline{0.8027}\\
Tied encoder & \thirdbest{0.8060} & \thirdbest{0.8983} & 0.7103 & \thirdbest{0.7767} & \thirdbest{0.3502} & \underline{0.3109} & \textbf{0.3385} & \textbf{0.2367} & \textbf{0.2449} & \underline{0.6482} & \textbf{0.6884} & 0.0263 & \underline{0.0478} & 0.6328 & \thirdbest{0.7944}\\
Local decoder & \underline{0.8115} & \textbf{0.9469} & \underline{0.7148} & \textbf{0.7991} & \textbf{0.3810} & \thirdbest{0.3104} & \underline{0.3205} & \underline{0.2309} & \thirdbest{0.2230} & 0.5821 & \thirdbest{0.6151} & \underline{0.0406} & \thirdbest{0.0414} & \textbf{0.6710} & \textbf{0.8080}\\
1 stage & 0.8001 & 0.8929 & 0.7133 & 0.7561 & 0.3370 & 0.2206 & 0.2529 & 0.1276 & 0.1577 & \thirdbest{0.5857} & 0.6074 & 0.0254 & 0.0211 & 0.6181 & 0.7741\\
Local + 1 stage & 0.7762 & \underline{0.9343} & \textbf{0.7156} & \underline{0.7799} & 0.3172 & 0.2256 & 0.2400 & 0.1308 & 0.1366 & 0.4871 & 0.4938 & \thirdbest{0.0274} & 0.0291 & \thirdbest{0.6400} & 0.7704\\
\bottomrule
\end{tabular}\end{adjustbox}
}

\end{minipage}
\end{table}

\end{supplementaryresults}
\supplementaryresultbreak

\supplementaryresultbreak
% Source: supplementary/sections/12-application-demo.tex
\FloatBarrier
\section{Driving world model intervention case studies}
\label{sec:supp-application-demo}

We examine 16 simulated crash cases: eight from SimWAM and eight from DriveLaW-Act.
We label the cases S1--S8 for SimWAM and D1--D8 for DriveLaW-Act.
They cover 13 distinct driving scenes, three of which produce a crash in both models.
In block labels, L denotes the transformer layer and B the block index within that layer's dictionary; L15/B460 therefore identifies block 460 at layer 15.
EARS identifies candidate blocks through car and road localisation; we then test activation edits and measure their effects on driving.
The study includes the two interventions in Sec.~6 of the main paper and additional interventions evaluated with the same driving criteria.
The results show changes in arrival timing, path placement, or both, but contact removal does not always preserve driving quality.
\Cref{tab:supp-demo-all-case-clearance} compares all cases; the following figures explain individual outcomes.
% Source: supplementary/sections/application-demo/01-selection-and-recipes.tex
\subsection{Selecting intervention candidates with EARS-Graph}
\label{sec:supp-demo-selection}

\paragraph{Selecting crash cases.}
We screened saved planner trajectories from 68,538 fitting and 6,653 validation scenes per model; original-test trajectories were excluded.
The 16 selected cases are simulated vehicle collisions attributable to the ego planner under NAVSIM's fault rules.
All come from the fitting population.
Each meets the following criteria.
\par\smallskip
\noindent{\small\setlength{\tabcolsep}{3pt}\renewcommand{\arraystretch}{1.12}
\begin{tabularx}{0.99\linewidth}{@{}p{0.24\columnwidth}Y@{}}
\toprule Criterion & Requirement\\\midrule
Contact & The ego footprint overlaps a recorded actor along the saved 4-second plan.\\
Robustness & Overlap survives shrinking both boxes by 20 cm on every edge.\\
Persistence & Robust overlap occurs at 2 consecutive screening samples, spaced 0.5 s apart.\\
Confirmation & NAVSIM's independent PDM simulation~\citep{dauner2024navsim} classifies the collision as ego-at-fault.\\
Actor identity & The confirmed collision involves the same screened vehicle.\\
Clean reference & This vehicle overlaps neither ego at the start nor the recorded human-driven trajectory.\\\bottomrule
\end{tabularx}}
\par\smallskip
NAVSIM assigns fault to front collisions and collisions with stopped actors, and to lateral collisions when ego occupies multiple lanes or a non-drivable area.
These are failures of predicted plans against recorded actor motion, not crashes in the human driving record.
We impose no impact-speed or injury-severity threshold.
A DriveWAM vehicle graze failed the box-shrinkage criterion and was excluded.
\paragraph{Selecting blocks to intervene on.}
EARS ranks Dataset-Full spatialBSF blocks by how well their direction readouts localise cars or road surfaces, using cross-validation across fitting drives.
We shortlist blocks from SimWAM layers 7, 15 and 23 and DriveLaW-Act layers 14, 18 and 24.
The initial top 16 blocks per concept and layer yield 96 distinct SimWAM candidates; subsequent searches expand selected lists to 64.
For the interventions in Sec.~6 of the main paper, L15/B460 ranks first for car localisation (F1 0.3821), and L7/B952 ranks third for road localisation (F1 0.6918).
EARS supplies these candidates; the intervention tests in \cref{sec:supp-demo-edit-scale} determine how to change them.

In the case captions, ``car-associated'' and ``road-associated'' refer to these localisation results.
Car-based selection preceded the truck/bus task, so truck/bus associations are used only to interpret the results.
The crash-scene graphs in \cref{sec:supp-demo-lane-clearance} were also added after the interventions.

\subsection{Intervention construction and edit magnitude}
\label{sec:supp-demo-edit-scale}

Following previous work~\citep{swann_multi-dimensional_2026}, we change selected block coefficients and add their decoded change to the original activation field.
This preserves the part of the original activation that the dictionary does not reconstruct, so the intervention tests the selected change without also replacing the rest of the field.
For a block $b$,
\begin{align}
 \Delta\mC &= \sigma_{\mathrm{fit}}\mH_b\Delta\mZ_b\mD_b^{\mathsf T},\notag\\
 \mC_{\mathrm{edit}} &= \operatorname{BF16}(\mC+\Delta\mC),\notag\\
 \rho &= \frac{\|\mC_{\mathrm{edit}}-\mC\|_F}{\|\mC\|_F}.
 \label{eq:supp-demo-edit-size}
\end{align}
Here $\mC$ is the original field and $\Delta\mZ_b$ is the coefficient change; $\mD_b$ decodes block coordinates and $\mH_b$ transports contributions between neighbours.
The fitting scale $\sigma_{\mathrm{fit}}$ restores the model's activation units.
The relative edit size $\rho$ measures the delivered change, including the model's bfloat16 (BF16) rounding.

\paragraph{Choosing an edit.}
Starting from the EARS shortlist, we test block multipliers, within-block directions, spatial regions and block combinations.
We select edits by contact removal, driving quality and progress, using the checks in \cref{sec:supp-demo-checks}.
This procedure is shared by both models and by the main-paper examples and subsequent cases.
The initial search used three development scenes; the later searches extend it to all 16 cases.

\paragraph{Model interfaces and combinations.}
We apply the decoded change before query/key/value projection.
SimWAM produces a plan through 20 action-denoising steps.
For DriveLaW-Act, we edit the action-block input at 256 retained positions in the latest observed latent slot; this context is reused through all 10 action-denoising calls.
A multiplier $m$ adds $(m-1)$ times the block's original contribution.
For joint edits, we weight and sum the decoded changes; across layers, edit size is the root-sum-square of the individual edit-to-field ratios.
Component-omission tests leave one component unedited while keeping the other component's weighted change fixed.

\paragraph{The two interventions in Sec.~6 of the main paper.}
For the box-truck scene (S3), we add a perturbation direction within the car-associated block L15/B460, restricted spatially to a manually marked truck region. This intervention direction is fitted on a separate scene using a local action Jacobian to favour slowing and penalise lateral change, subject to a nonnegative first-order change in the frozen car-direction readout averaged over the marked region; the fitted intervention direction is then applied to S3 without refitting. For the UPS scene (S8), we ablate the decoded contribution of the road-associated block L7/B952 across the entire spatial field by setting $\Delta\mZ_{952}=-\mZ_{952}$. These two interventions yield relative edit sizes of $\rho=10.34$ and $\rho=2.54$, respectively—ratios that can exceed unity because individual decoded block contributions partially cancel in the unperturbed field.

% Source: supplementary/sections/application-demo/02-outcomes-and-limits.tex
\subsection{Evaluating intervention outcomes}
\label{sec:supp-demo-checks}

An edit can prevent a collision while causing other driving problems.
We therefore evaluate both collision removal and driving quality, then measure how the edit changes the vehicle's path and timing.

\paragraph{Compare original and edited runs.}
Each pair uses the same observation, vehicle state, navigation command and action noise.
NAVSIM tracks each four-second plan without replanning; other actors follow their recorded trajectories and do not react to ego.
We keep the selected edit fixed and repeat the comparison twice with different action noise, giving three paired runs per intervention.
Figures show the run used to select the edit.

\paragraph{Check collision removal and driving quality.}
We report two outcomes: no at-fault contact, and passing all driving checks.
The latter also requires road and direction compliance, time-to-collision and comfort checks, positive progress, and at most 0.1 m of backward travel along the initial forward axis.
This last check is distinct from reversing relative to the vehicle's changing heading during a turn.

\begin{figure}[!htbp]
\centering
\begin{minipage}{0.90\textwidth}

\centering\includegraphics[width=\textwidth,height=0.40\textheight,keepaspectratio]{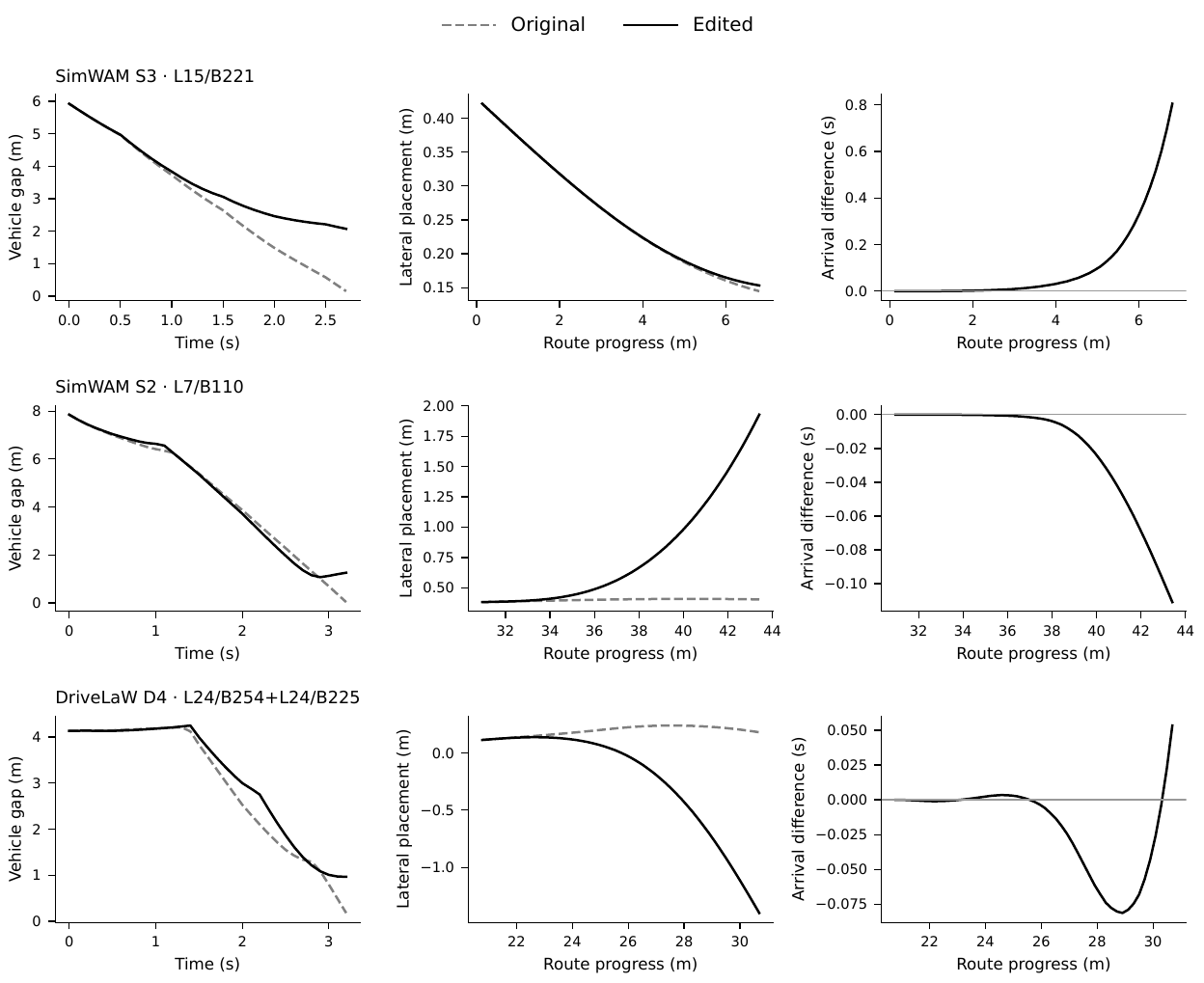}
\caption{\textbf{Interventions can change timing, path placement, or both.}
Rows show SimWAM S3 (L15/B221), SimWAM S2 (L7/B110), and DriveLaW D4 (L24/B254+B225).
Dashed grey is original; solid black is edited. The right column shows edited minus original arrival time at the same route position.
S3 arrives 0.81 s later with less than 0.01 m lateral difference.
S2 shifts laterally by up to 1.52 m and arrives 0.11 s earlier; D4 shifts by up to 1.58 m with a 0.05 s delay.
S3 here uses B221; Sec.~6 of the main paper uses B460 on the same observation.}
\label{fig:supp-demo-clearance-mechanisms}

\end{minipage}
\end{figure}

\paragraph{Measure path and timing changes.}
Vehicle separation is the shortest distance between the ego footprint and another vehicle's footprint at the same time, using their tracked headings and positions every 0.1 s.
We record footprint intersections separately because zero distance can mean touching or overlap.
To distinguish a path change from a speed change, we compare lateral position and heading at the same position along the intended route.
We then compare when the original and edited runs reach that position: a positive arrival-time difference means the edited vehicle arrives later.
These comparisons stop before the first physical contact in either run.
Route matching is unavailable for S1 and D6.
\Cref{fig:supp-demo-clearance-mechanisms} illustrates the measurements.

\paragraph{Results across the 16 cases.}
In the selection run, 15 interventions remove at-fault contact and 10 pass all driving checks.
Across all three runs, 11 consistently remove contact and six consistently pass all checks.
\Cref{tab:supp-demo-all-case-clearance} gives the case-level results.
These are selected fitting-scene examples; the repeated runs assess consistency on those scenes.

% Source: supplementary/tables/12-46a-crash-clearance-comparison.tex
\begin{table}[!htbp]
\centering
\begin{minipage}{0.94\textwidth}

\centering\small\setlength{\tabcolsep}{3pt}
\begin{adjustbox}{max width=\linewidth}\begin{tabular}{@{}llccrrrrcc@{}}
\toprule
Case / edit & Blocks & $c_{\rm lane}$ & $c_{\rm route}$ & Gap (m) & $|\Delta l|$ (m) & $|\Delta\psi|$ ($^\circ$) & $\Delta t$ (s) & No at-fault & All checks\\
\midrule
\multicolumn{10}{@{}l}{\textbf{SimWAM}}\\
S1 & L15/B460+L15/B953 & 0.976 & 0.834 & 0.07$\to$2.89 & NA & NA & NA & 3/3 & 3/3\\
S2 & L7/B110 & NA & 0.891 & 0.03$\to$1.07 & 1.52 & 20.8 & -0.11 & 3/3 & 3/3\\
S3 & L15/B221 & 0.994 & 0.805 & 0.15$\to$2.07 & 0.01 & 0.3 & +0.81 & 3/3 & 3/3\\
S3 (Sec. 6) & L15/B460 & 0.994 & 0.805 & 0.15$\to$1.25 & 0.03 & 0.6 & +0.48 & 3/3 & 3/3\\
S4 & L7/B977 & NA & 1.000 & 0.45$\to$2.48 & 0.83 & 10.9 & +0.46 & 3/3 & 3/3\\
S5 & L15/B278+L15/B383 & NA & 0.956 & 0.07$\to$2.69 & 0.62 & 13.8 & +0.60 & 3/3 & 2/3\\
S6 & L15/B664+L15/B808 & NA & 0.982 & 0.47$\to$1.85 & 0.89 & 12.1 & +0.29 & 2/3 & 1/3\\
S7 & L7/B40+L23/B877 & 1.000 & 0.942 & 0.11$\to$0.03 & 0.89 & 7.7 & -0.23 & 2/3 & 1/3\\
S8 & L23/B468 & 1.000 & 0.933 & 0.37$\to$0.08 & 0.42 & 4.8 & -0.39 & 2/3 & 1/3\\
S8 (Sec. 6) & L7/B952 & 1.000 & 0.933 & 0.37$\to$1.79 & 0.31 & 6.8 & +0.38 & 3/3 & 3/3\\
\midrule
\multicolumn{10}{@{}l}{\textbf{DriveLaW-Act}}\\
D1 & L18/B431 & NA & 0.956 & 0.13$\to$1.80 & 1.35 & 25.4 & +0.14 & 3/3 & 0/3\\
D2 & L24/B225 & NA & 0.982 & 0.43$\to$0.90 & 1.36 & 18.1 & -0.01 & 3/3 & 0/3\\
D3 & L24/B225+L18/B310 & NA & 0.856 & 0.04$\to$0.98 & 2.05 & 23.6 & -0.02 & 3/3 & 0/3\\
D4 & L24/B254+L24/B225 & NA & 0.836 & 0.18$\to$0.97 & 1.58 & 20.1 & +0.05 & 3/3 & 3/3\\
D5 & L18/B310 & NA & 0.766 & 0.10$\to$0.02 & 0.82 & 17.2 & -0.25 & 1/3 & 0/3\\
D6 & L18/B431 & 0.976 & 0.834 & 0.18$\to$0.40 & NA & NA & NA & 3/3 & 0/3\\
D7 & L18/B865 & NA & 0.588 & 1.27$\to$0.18 & 0.00 & 0.1 & -0.31 & 0/3 & 0/3\\
D8 & L18/B431+L18/B310 & NA & 1.000 & 0.35$\to$1.14 & 0.83 & 7.8 & -0.03 & 3/3 & 3/3\\
D8 (B310 alone) & L18/B310 & NA & 1.000 & 0.35$\to$1.10 & 0.76 & 6.9 & +0.02 & 2/3 & 1/3\\
\bottomrule\end{tabular}\end{adjustbox}
\caption{\textbf{Interventions across all 16 crash cases.}
The last two columns count edited runs with no at-fault contact and runs passing all driving checks; 3/3 means all three runs pass.
Gap is original $\to$ edited minimum vehicle separation before the first physical contact in either run.
$|\Delta l|$ and $|\Delta\psi|$ are maximum lateral and heading differences at the same route position; $\Delta t$ is edited minus original arrival time at the end of the shared route interval.
Positive $\Delta t$ means later arrival. Trajectory measurements use the selection run.
The image scores $c_{\rm lane}$ and $c_{\rm route}$ describe the shared observation and stay fixed across edits; higher values mean less vehicle-box overlap (\cref{sec:supp-demo-lane-clearance}).
NA means unavailable lane geometry or route matching.
The S3/S8 subrows reproduce the interventions in Sec.~6 of the main paper; D8 compares the joint edit with B310 alone.
Qualification requires all checks in \cref{sec:supp-demo-checks}.}
\label{tab:supp-demo-all-case-clearance}

\end{minipage}
\end{table}

\supplementaryresultbreak
% Source: supplementary/sections/application-demo/03-lane-clearance-association.tex
\subsection{EARS-Graph references and lane clearance}
\label{sec:supp-demo-lane-clearance}

Clearance misestimation is one plausible explanation for collision-prone plans.
We therefore test whether the natural activity of the selected intervention blocks is associated with vehicle obstruction of the ego lane.

\begin{figure}[!htbp]
\centering
\begin{minipage}{0.90\textwidth}

 \centering
 \includegraphics[width=\textwidth,height=0.24\textheight,keepaspectratio]{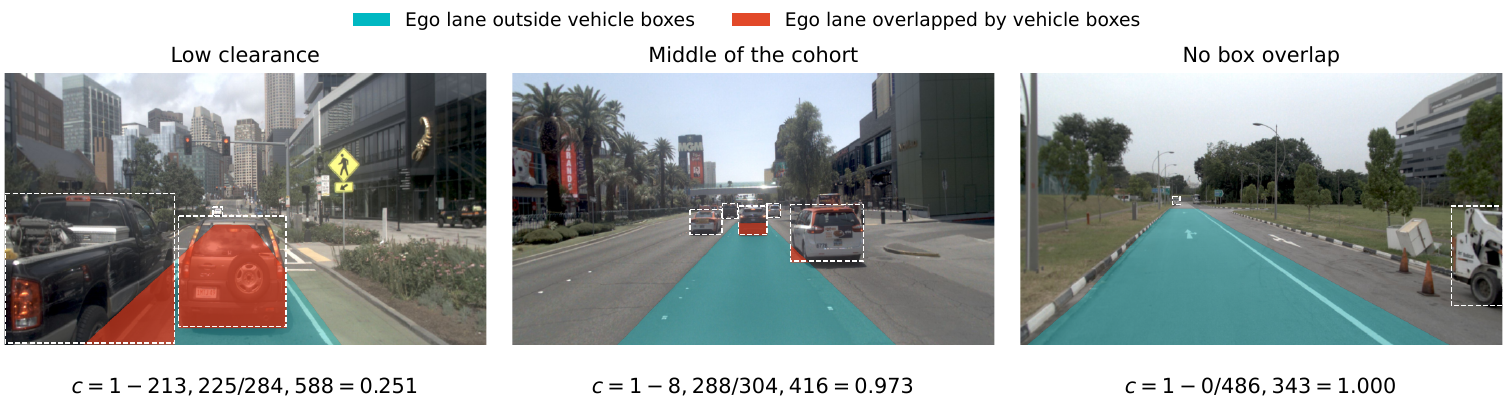}
 \caption{\textbf{Visual lane clearance from EARS-Graph scene annotations.}
 Cyan marks ego-lane pixels outside vehicle boxes; red marks their overlap, and dashed white outlines show the boxes.
 The denominator in \cref{eq:supp-demo-clearance} includes both cyan and red pixels.
 The 3 fitting scenes span low, middle and high reference values on distinct drives, independently of block activity.
}
 \label{fig:supp-lane-clearance-ears}
\par\medskip\hrule\medskip
\centering\includegraphics[width=\textwidth,height=0.24\textheight,keepaspectratio]{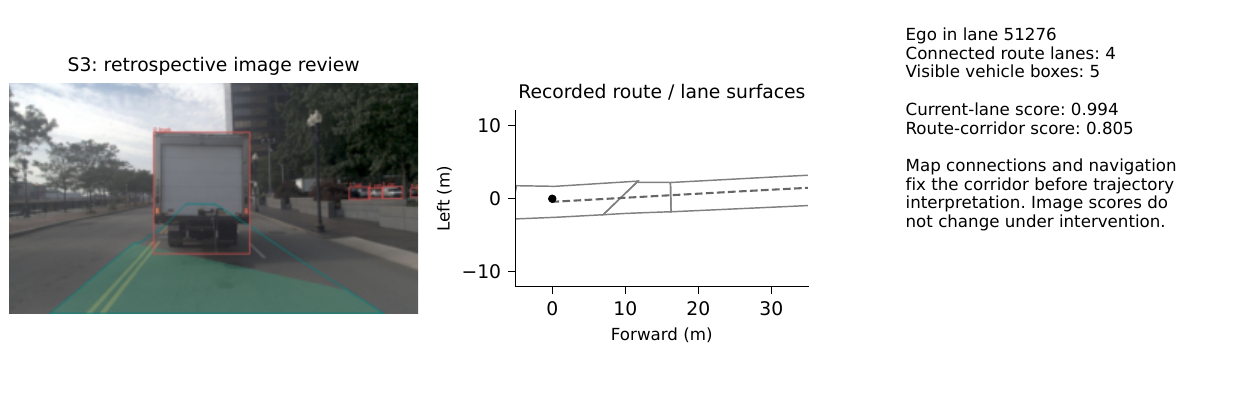}
\caption{\textbf{EARS scene reference for crash case S3.}
Left: vehicle boxes, current lane and route boundary in the rectified camera.
Middle: connected route-lane surfaces and centreline in ego coordinates; right: the supported graph relations and two image scores.
The current-lane and route-corridor scores describe the same observation; the latter follows the connected navigation route.}
\label{fig:supp-demo-clearance-crash-reference}

\end{minipage}
\end{figure}

\paragraph{Measuring clearance with EARS.}
EARS graphs provide the lane geometry and vehicle boxes used to measure image-based clearance (\cref{sec:supp-crs-schema}).
We construct one graph for each of the 13 crash observations and share it across models using the same image.
Let $\mathcal L_i$ be the projected current ego-lane surface and $\mathcal V_i$ the union of visible car, truck and bus boxes.
We define
\begin{equation}
 c_i=1-\frac{|\mathcal L_i\cap\mathcal V_i|}{|\mathcal L_i|}.
 \label{eq:supp-demo-clearance}
\end{equation}
The score is the fraction of the projected lane not covered by vehicle boxes (\cref{fig:supp-lane-clearance-ears}).
Lane surfaces are clipped to the image, and overlapping boxes contribute only once.
We leave the score unavailable when ego-lane membership is ambiguous or the projected surface is empty.

For turns, the current ego lane does not necessarily cover the intended outgoing road.
We therefore extend the same calculation to the connected lane surfaces along the next 50 m of the recorded navigation route, yielding an exploratory route-corridor score (\cref{fig:supp-demo-clearance-crash-reference}).
Current-lane clearance is available for 4 of the 13 crash observations, and route-corridor clearance for all 13.
Both describe the observed scene and remain fixed across interventions; trajectory-level vehicle separation is evaluated separately in \cref{sec:supp-demo-checks}.

To test whether block activity tracks visual ego-lane clearance, we analyse a separate cohort of 122 fitting scenes from 35 drives, excluding the crash-case observations and repeated intervention runs. For each scene, we define a block's activity as the mean of its five largest source-contribution norms across spatial positions, after conversion to common units; inactive positions contribute zeros. We calculate Spearman correlations with equal weight assigned to each drive. We also adjust the ranked activity and clearance values for drive identity and the pixel areas of lanes, roads and vehicle boxes, then correlate the residuals. This adjustment tests whether the association persists after accounting for measured differences in recording conditions and scene composition. Descriptive 95\% confidence intervals use 2,000 whole-drive bootstrap resamples for all reported blocks, including SimWAM L15/B460 and L7/B952.

\begin{figure}[!htbp]
\centering
\begin{minipage}{0.90\textwidth}

\centering\includegraphics[width=\textwidth,height=0.40\textheight,keepaspectratio]{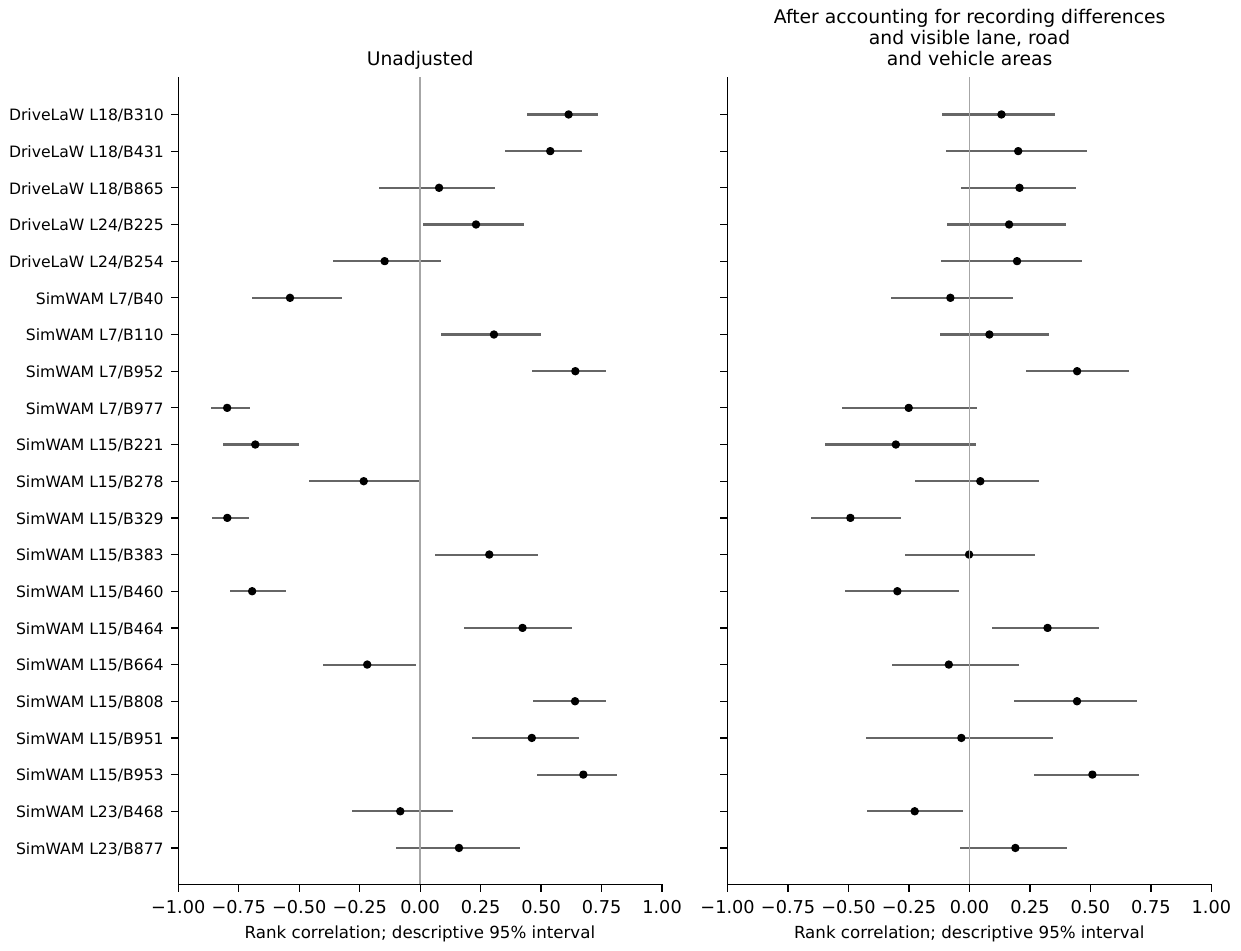}
\caption{\textbf{Natural clearance associations for all 21 crash-intervention blocks.}
Both panels use the same 122 fitting scenes from 35 drives.
The right panel accounts for recording differences and lane, road and vehicle-box pixel areas.
Its descriptive 95\% intervals exclude zero for 7 of 16 SimWAM blocks and none of the 5 DriveLaW-Act blocks.}
\label{fig:supp-demo-clearance-all-blocks}

\end{minipage}
\end{figure}

\paragraph{Clearance associations in the selected blocks.}
Within the selected intervention blocks, both models show clearance associations, with stronger evidence after adjustment in SimWAM (\cref{fig:supp-demo-clearance-all-blocks}).
Unadjusted 95\% intervals exclude zero for 13 of 16 SimWAM blocks and 3 of 5 DriveLaW-Act blocks; after adjustment, these counts fall to 7 of 16 and 0 of 5, respectively.
For example, DriveLaW-Act L18/B310 has an unadjusted correlation of 0.614, but its adjusted estimate is 0.132 with an interval of $[-0.110,0.350]$.

The two main-paper SimWAM blocks retain associations of opposite sign (\cref{tab:supp-demo-lane-clearance}).
Activity in road-associated L7/B952 increases with clearance: its correlation decreases from 0.642 to 0.445 after adjustment, with both intervals excluding zero.
Activity in car-associated L15/B460 decreases with clearance, with an adjusted correlation of $-0.298$.
These directions are consistent with their EARS road and car profiles and support a clearance-related interpretation of their activity.
\begin{table}[!htbp]
\centering
\begin{minipage}{0.94\textwidth}

 \centering
 \small
 \setlength{\tabcolsep}{3pt}
 \caption{\textbf{SimWAM block activity and visual ego-lane clearance.} Equal-drive Spearman correlations and drive-bootstrap 95\% intervals use 122 fitting scenes from 35 drives. Adjustment accounts for recording identity and lane, road and vehicle-box pixel areas.}
 \label{tab:supp-demo-lane-clearance}
 \begin{adjustbox}{max width=\linewidth}\begin{tabular}{@{}lcc@{}}
 \toprule
 Block & Unadjusted & Adjusted\\
 \midrule
 L7/B952 & \shortstack{$0.642$\\$[0.466,0.766]$} & \shortstack{$0.445$\\$[0.239,0.653]$}\\[3pt]
 L15/B460 & \shortstack{$-0.694$\\$[-0.780,-0.561]$} & \shortstack{$-0.298$\\$[-0.510,-0.046]$}\\
 \bottomrule
 \end{tabular}\end{adjustbox}

\end{minipage}
\end{table}

\supplementaryresultbreak
% Source: supplementary/sections/application-demo/04-eight-scene-block-analysis.tex
\subsection{Block effects across 8 SimWAM scenes}
\label{sec:supp-demo-block-semantics}

Clearance associations measure natural block activity across 122 scenes from 35 drives, using the Spearman correlations and drive/image-area adjustments detailed in \cref{sec:supp-demo-lane-clearance}. Because image clearance is fixed when editing the same observation, we evaluate driving effects using placement and arrival time at matched route positions, as well as vehicle separation prior to either run's first contact. Measurements refer to the displayed run, while driving outcomes span three saved seeds (\cref{sec:supp-demo-checks}). For joint edits, we isolate a component's contribution by holding the other edit constant. Additional seed-42 single-block searches identify alternative contact-avoidance strategies using different doses; these are kept separate from both the fixed-dose component tests and the 289-condition clearance analysis.

Our separate trajectory and clearance analysis spans 289 saved conditions across all 16 SimWAM and DriveLaW-Act cases. Each condition defines the DWM, scene, action-sampling seed, and intervention recipe (including unedited baselines). This dataset covers the main-paper interventions, illustrated alternatives, confirmation runs, matched controls, and component-omission tests.

\begin{figure}[!htbp]
\centering
\begin{minipage}{0.90\textwidth}

 \centering
 \includegraphics[width=\textwidth,height=0.40\textheight,keepaspectratio]{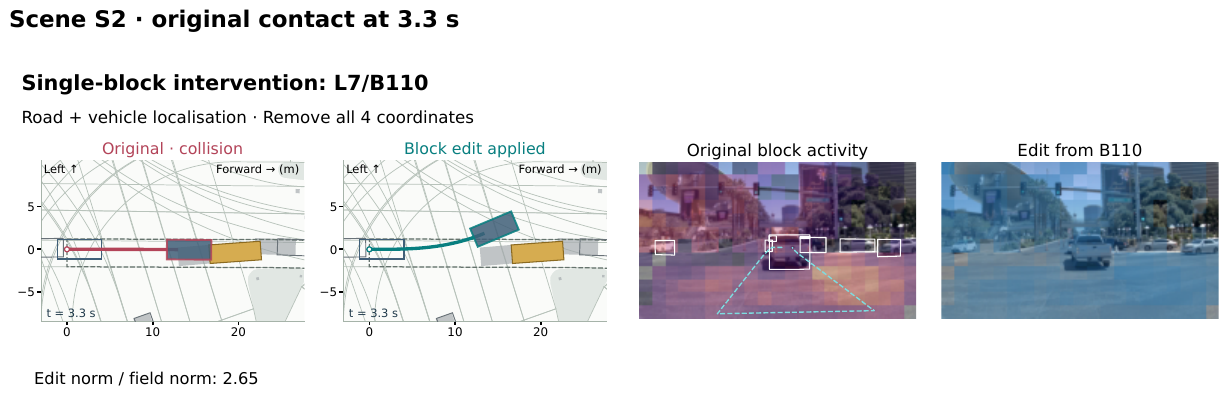}
 \caption{\textbf{S2: removing B110 changes path placement.}
L7/B110 (road-associated) also localises vehicle regions.
Removing its complete contribution shifts the path laterally and passes driving checks in all 3 runs.}
 \label{fig:supp-demo-case-b}

\end{minipage}
\end{figure}

\paragraph{Responding to vehicles ahead: L15/B460 and L15/B329.}
B460 and B329 are vehicle-associated and more active in scenes with lower lane clearance, including after adjustment (\cref{fig:supp-demo-clearance-all-blocks}). In S3, adding a selected direction within B460 delays the approach to the truck with little sideways movement, avoiding contact and passing every driving check in 3/3 runs (main-paper Fig.~\ref{fig:simwam-application}(a)). To test whether this improvement depended on the selected direction, we compared it with three random directions within B460; all three controls still collide. Restoring the original activations also restores the collision. Together with B460's vehicle association, the delayed approach suggests that vehicle-related information had too little influence on the original approach timing. This effect does not require editing the truck region: the selected direction also works outside it. \paragraph{Choosing a path on the road: L7/B110 and L7/B952.}
B110 and B952 are road-associated and more active in clearer scenes; the adjusted relationship remains clear for B952 but is uncertain for B110. Removing B110's contribution shifts the S2 path sideways by up to 1.52 m, giving the pickup more room (\cref{fig:supp-demo-clearance-mechanisms,fig:supp-demo-case-b}). Removing B952 likewise changes passing position and delays arrival around the S8 truck (main-paper Fig.~\ref{fig:simwam-application}(b)). Both edits avoid contact and pass every driving check in 3/3 runs. To test whether B952's benefit depends on how and where it is edited, we compared the selected removal with random rotations and spatially restricted removals. One of three random rotations avoids contact, whereas removal confined to either the actor region or its complement fails. As with B460, restoring the original activations restores the collision. The path changes suggest that the original road-related contributions favoured paths that left too little room for nearby vehicles.

\FloatBarrier

\clearpage
{\small
\bibliographystyle{ieeenat_fullname}
\bibliography{references}
}
\end{document}